\documentclass{article}

\usepackage[preprint]{neurips_2026}

\usepackage[utf8]{inputenc}
\usepackage[T1]{fontenc}
\usepackage{CJKutf8}
\usepackage{amsmath}
\usepackage{amssymb}
\usepackage{amsfonts}
\usepackage{booktabs}
\usepackage{graphicx}
\usepackage{subcaption}
\usepackage{multirow}
\usepackage{nicefrac}
\usepackage{microtype}
\usepackage{xcolor}
\usepackage{float}
\usepackage{placeins}
\usepackage{url}
\usepackage{hyperref}
\hypersetup{hidelinks}

\graphicspath{{figures/}}

\title{BRIDGE: Background Routing and Isolated Discrete Gating for Coarse-Mask Local Editing}

\author{%
  Peilin Xiong \qquad Honghui Yuan \qquad Junwen Chen \qquad Keiji Yanai \\
Department of Informatics, The University of Electro-Communications, Tokyo, Japan \\
\texttt{xiong-p@mm.inf.uec.ac.jp} \qquad \texttt{yuan-h@mm.inf.uec.ac.jp} \\
\texttt{chen-j@mm.inf.uec.ac.jp} \qquad \texttt{yanai@cs.uec.ac.jp}
}

\begin{document}

\maketitle

\begingroup
\setlength{\intextsep}{0pt}
\captionsetup{skip=4pt,font=small}
\begin{figure}[H]
    \centering
    \setlength{\tabcolsep}{1pt}
    \renewcommand{\arraystretch}{0.5}
    \begin{tabular}{c c c c}
        \rotatebox{90}{\scriptsize \textbf{Remove}} &
        \includegraphics[width=0.24\linewidth]{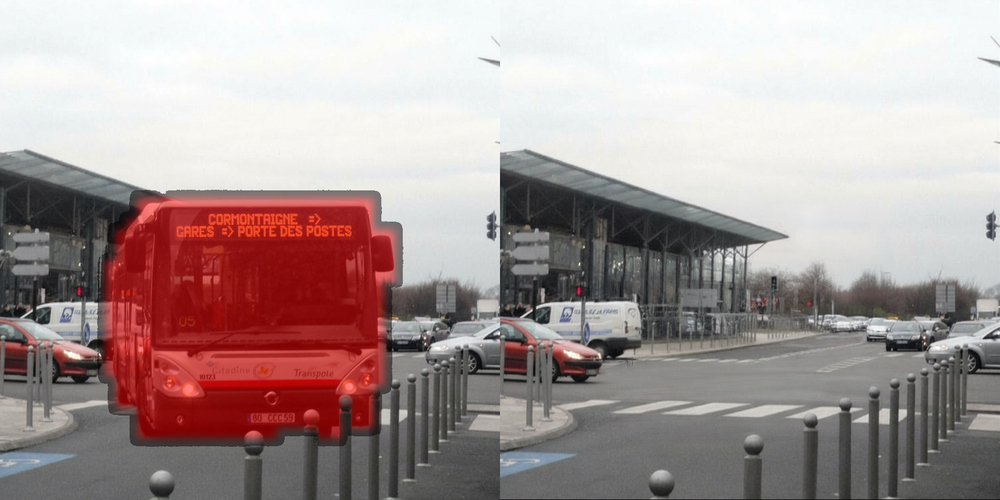} &
        \includegraphics[width=0.24\linewidth]{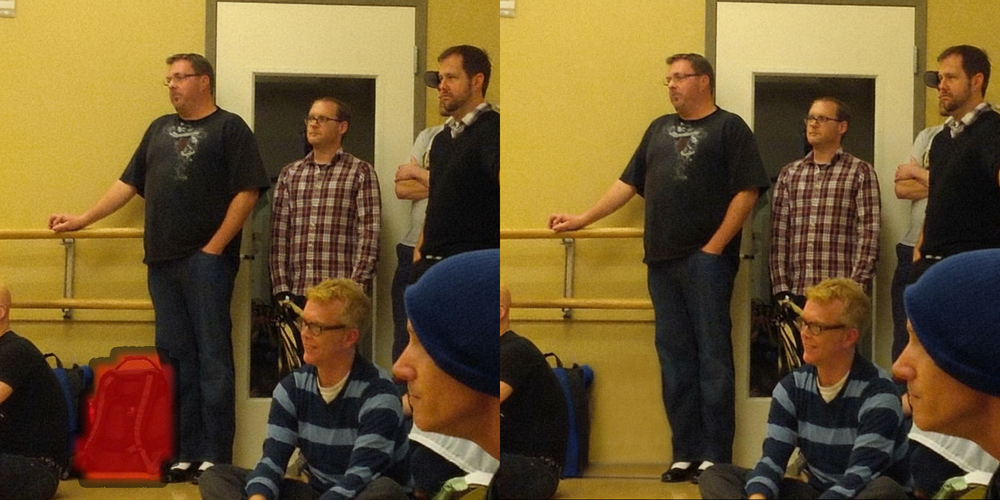} &
        \includegraphics[width=0.24\linewidth]{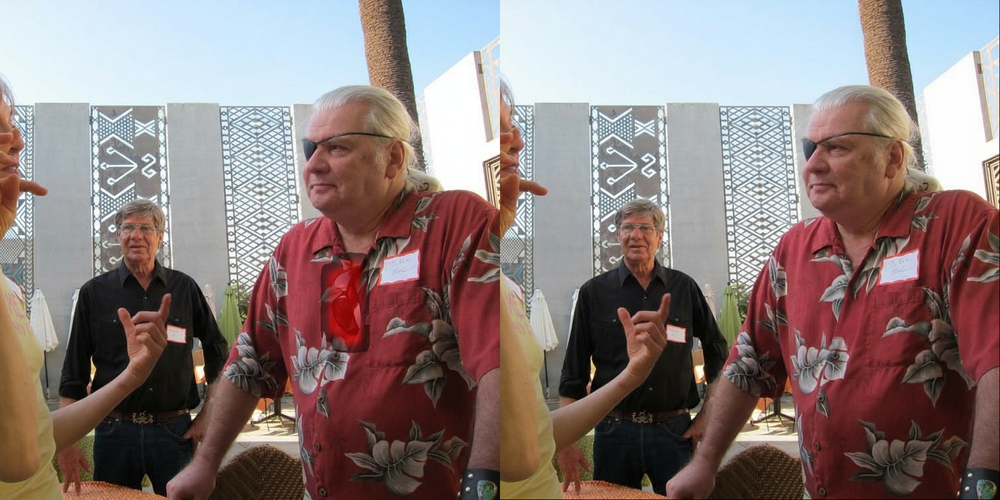} \\
        & \scriptsize \textit{``Remove the bus...''} & \scriptsize \textit{``Remove the backpack...''} & \scriptsize \textit{``Remove sunglasses...''} \\
        \rotatebox{90}{\scriptsize \textbf{Add}} &
        \includegraphics[width=0.24\linewidth]{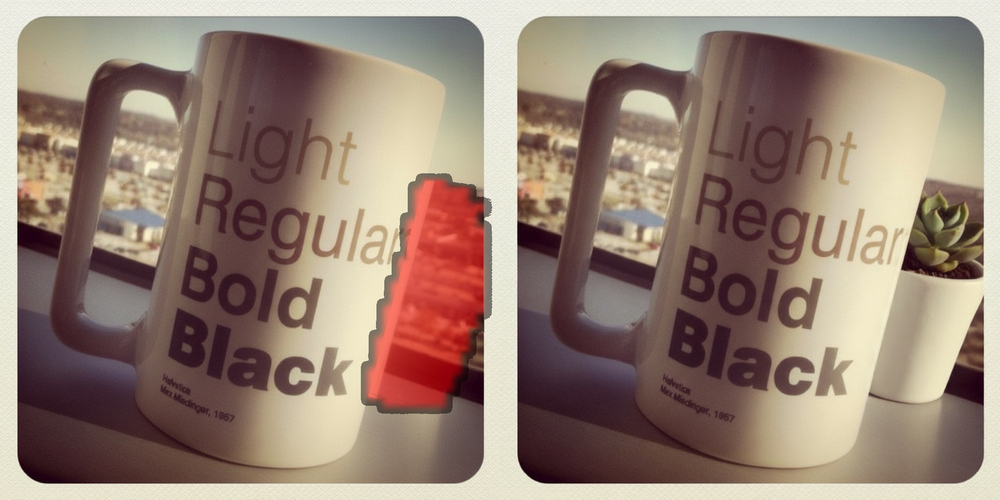} &
        \includegraphics[width=0.24\linewidth]{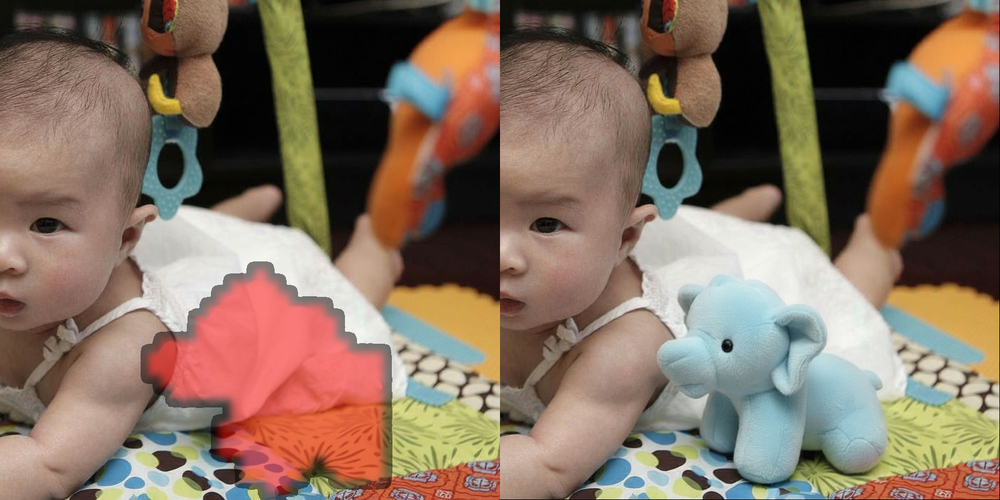} &
        \includegraphics[width=0.24\linewidth]{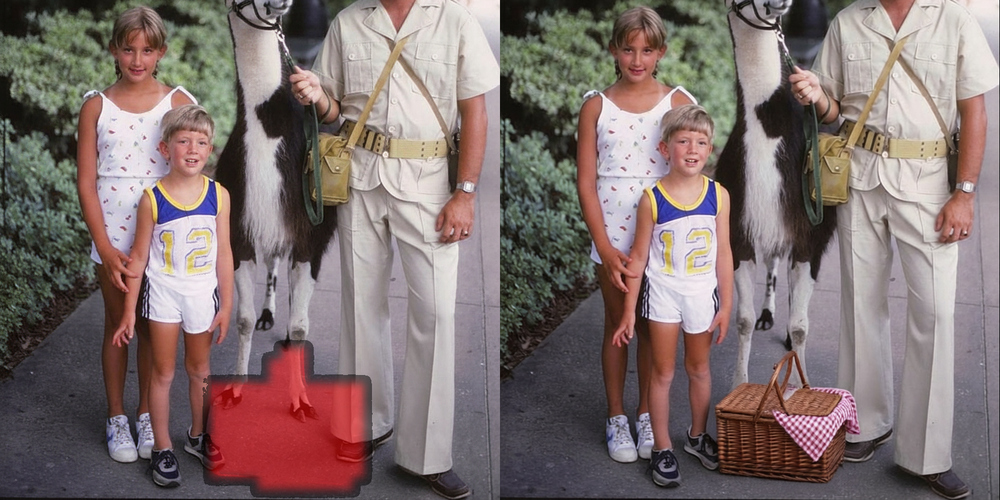} \\
        & \scriptsize \textit{``Add a succulent...''} & \scriptsize \textit{``Add an elephant plush...''} & \scriptsize \textit{``Add a picnic basket...''} \\
        \rotatebox{90}{\scriptsize \textbf{Change}} &
        \includegraphics[width=0.24\linewidth]{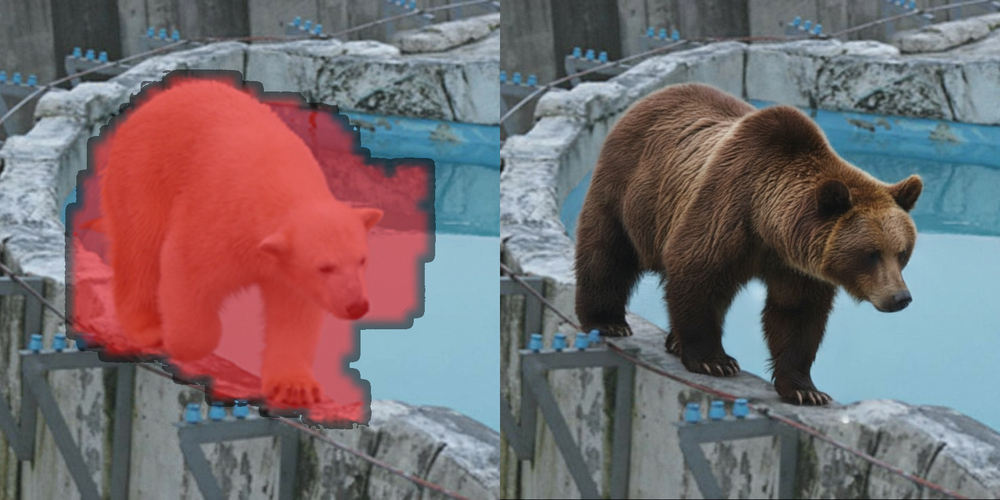} &
        \includegraphics[width=0.24\linewidth]{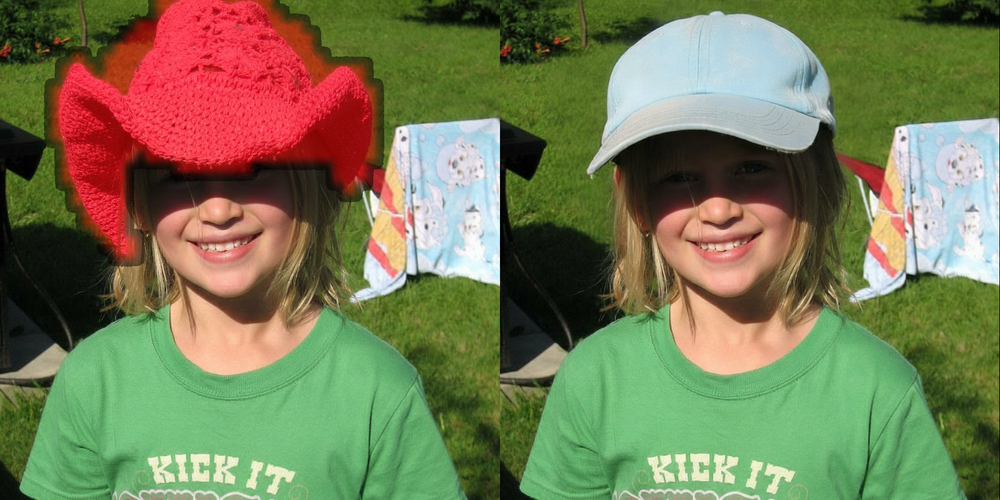} &
        \includegraphics[width=0.24\linewidth]{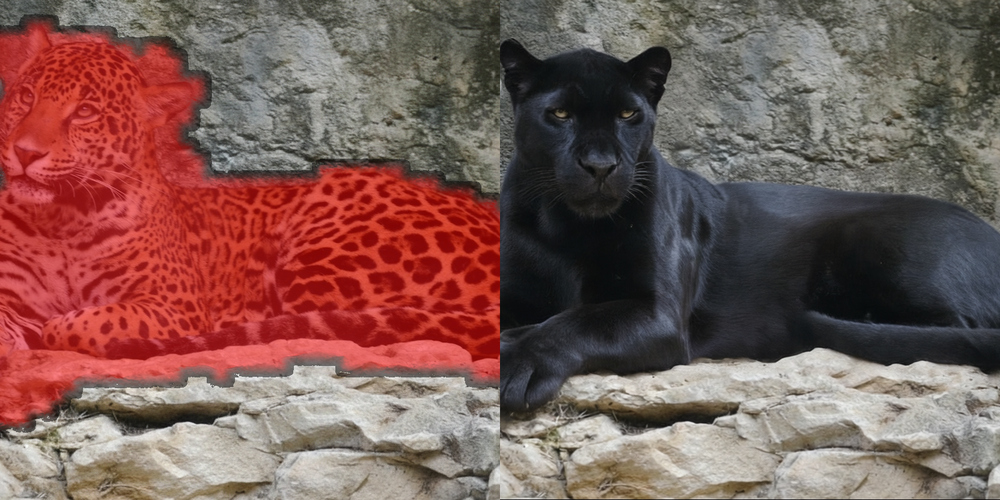} \\
        & \scriptsize \textit{``Polar bear $\rightarrow$ Brown bear''} & \scriptsize \textit{``Pink hat $\rightarrow$ Baseball cap''} & \scriptsize \textit{``Jaguar $\rightarrow$ Panther''} \\
    \end{tabular}
    \caption{\textbf{BRIDGE produces high-quality coarse-mask local edits while preserving seamless background fusion.} All examples in this figure are generated by our final BRIDGE model. Across remove, add, and change edits, the generated results remain well integrated with the surrounding scene while allowing the edited subject to depart from the rough mask shape when needed. Together, these examples illustrate the central goal of coarse-mask local editing: maintain background consistency without inheriting mask-shape bias.}
    \label{fig:teaser}
\end{figure}

\endgroup
\begin{abstract}
Coarse-mask local editing must preserve the background while allowing new
content to depart from the mask contour. Existing mask-conditioned editors
often entangle these goals and treat the coarse mask as a shape prior. We
introduce \textbf{BRIDGE}, a dual-path architecture with a full-canvas Main
Path, a local Subject Path, and a learned binary gate that routes subject-token
positional embeddings between subject-centric and background-anchored
coordinates. On Qwen-Image-Edit-2511, BRIDGE keeps the mask outside the DiT
backbone and trains both targets with independent noise. On BRIDGE-Bench, it
achieves 0.503 local SigLIP2-T, 0.902 DINO, and 0.175 DreamSim, outperforming
ACE++ and FLUX.1-Fill in local precision. We then instantiate BRIDGE on FLUX.2
[klein] 9B Base with full-transformer training and subject-reference
conditioning. Dense BBox supervision preserves Sub-image quality; optional
mask-cropped inference removes out-of-mask Sub tokens, tightening user support
without training a sparse model that overfits the mask silhouette. On supported
ICE-Bench tasks, FLUX.2 BRIDGE averages 0.5941 over four local edits and 0.6254
over subject-guided edits, leading the compared open-source models on the
latter. Across Qwen and FLUX.2, BRIDGE separates support localization from
subject geometry and improves local generation while preserving the
surrounding scene.
Code is available at \url{https://github.com/XIONGPEILIN/BRIDGE}.
\end{abstract}

\section{Introduction}
\label{sec:intro}

\begin{figure}[t]
  \centering
  \includegraphics[width=0.99\linewidth]{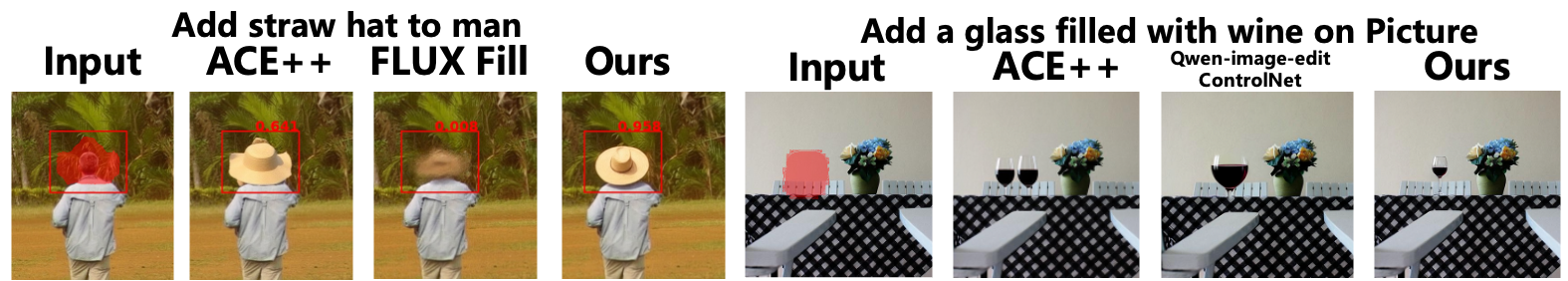}
  \caption{\textbf{Mask-shape bias in coarse-mask local editing.} The left four examples are from the test split of our constructed dataset, and the right four examples are from the public ICE-Bench benchmark. On the left, ACE++ produces a jagged hat boundary that follows the rough mask, FLUX.1-Fill fails to complete the edit, and our method \textbf{BRIDGE} generates a natural round hat. On the right, \textbf{BRIDGE} generates a single cup as requested, whereas ACE++ produces two cups and Q-Control~\citep{diffsynth2025qwencontrol} enlarges the cup unrealistically, showing that mask-conditioned baselines often overfit the wide mask instead of following the instruction.}
  \label{fig:mask_bias}
\end{figure}

In real local image editing, users rarely provide object-accurate masks. They
draw rough scribbles or boxes to indicate where an edit should occur. The mask
must therefore localize the operation without prescribing the geometry of the
result, while the editor must preserve the surrounding scene. We call the
failure to satisfy this separation \emph{mask-shape bias}: a model treats a
localization hint as the target object contour. This produces a Two-Zone
Constraint: the background should remain stable, whereas the editable region
should follow the instruction and may depart from the mask boundary.
Figs.~\ref{fig:teaser} and~\ref{fig:mask_bias} show the resulting tension in
remove, add, and change edits.

Existing methods obtain locality mainly through explicit mask conditioning or
inference-time attention control. Inpainting and control-branch editors such as
FLUX~\citep{labs2025flux}, ACE++~\citep{han2025aceplus},
BrushNet~\citep{ju2024brushnet}, and PowerPaint-style task
prompting~\citep{zhuang2024task} encode the mask as latent tokens, auxiliary
features, or visual context. This provides strong spatial support, but also
allows an imprecise input contour to influence object shape. Training-free
methods such as Prompt-to-Prompt~\citep{hertz2022prompt}, Null-text
Inversion~\citep{mokady2023nulltext}, MasaCtrl~\citep{cao2023masactrl}, and
Plug-and-Play diffusion features~\citep{tumanyan2023plug} manipulate attention
without retraining, but large geometric changes depend on schedules and
attention heuristics. Neither route explicitly separates the support used for
localization from the coordinate system used to generate a new subject.

A diagnostic Qwen-Image experiment~\citep{qwenimage} provides the key design
clue. Assigning the same positional embedding (PE) to different visual tokens
causes them to reuse similar image context, while a cross-region attention mask
suppresses this coupling. PE assignment and attention connectivity therefore
jointly control where a token borrows context and how strongly that context is
used. Based on this observation, we propose \textbf{BRIDGE} (Background Routing
and Isolated Discrete Gating for Coarse-Mask Local Editing). BRIDGE processes a
full-canvas Main Path and a local Subject Path in one DiT. The mask defines the
Subject Path support and external latent-blending region; a learned binary gate
routes each subject token between subject-centric and background-anchored PEs.
Boundary tokens can borrow background coordinates for fusion, while interior
tokens retain the geometric freedom needed to create new structure.

Our first instantiation adapts Qwen-Image-Edit-2511~\citep{qwenimage} with
rank-512 LoRA~\citep{hu2022lora} and 13.31M GateBlock parameters across 60
layers. The Main and Subject targets receive independently sampled noise, and
the final edited image is decoded from the Main Path. On BRIDGE-Bench, this
model reaches 0.503 local SigLIP2-T, 0.902 local DINO, and 0.175 local
DreamSim. It also transfers zero-shot to MagicBrush and ICE-Bench, providing
evidence that the learned routing is useful beyond the constructed benchmark.

We then ask whether the same separation transfers to a different backbone and
to subject-reference editing. The FLUX.2 [klein] 9B Base~\citep{bfl2026flux2kleinbase9b} instantiation trains
the full transformer with layer-wise gates and conditions on a generated
subject reference. Its Sub branch uses an aligned crop. Training only
mask-selected Sub tokens produced less coherent Sub images, color drift for
small supports, and strong mask-shape overfitting. We therefore retain dense
BBox supervision for Sub-image quality and optionally crop Sub tokens by the
user mask only at inference. Tokens outside the mask are removed together with
their PE candidates, tightening spatial support without teaching a sparse
training model to fill the mask silhouette. The Main image is decoded
independently, so the cropped Sub preview is a diagnostic rather than the final
output.

The two backbones are evaluated under separate protocols. The Qwen model is
compared on BRIDGE-Bench, MagicBrush~\citep{zhang2023magicbrush}, and
ICE-Bench~\citep{pan2025icebench}. The FLUX.2 model is evaluated on supported
ICE-Bench mask tasks using rewritten self-contained prompts and subject
references when provided. It averages 0.5941 over four supported local-edit
tasks and 0.6254 over subject-guided editing, leading the compared open-source
models on the latter group. These results connect the original Qwen routing
mechanism with the later FLUX.2 extension while keeping their training and
evaluation settings distinct.

Our contributions are threefold:
\begin{itemize}
    \item We formulate mask-shape bias as a failure to separate spatial support from output geometry and introduce BRIDGE, which combines Main/Subject paths with learned token-level PE routing.
    \item We instantiate BRIDGE on both Qwen-Image-Edit-2511 and FLUX.2, and introduce dense BBox training with optional mask-cropped Sub inference for the FLUX.2 subject-reference setting.
    \item We evaluate the Qwen and FLUX.2 models under separate benchmark protocols, showing improved local precision, cross-benchmark transfer, and competitive open-source ICE-Bench results.
\end{itemize}

\section{Related Work}
\label{sec:related}

\subsection{Instruction-Guided Image Editing}
Generative image editing has advanced rapidly with diffusion models~\citep{ho2020denoising,rombach2022high,peebles2023scalable}. Early text-guided methods such as InstructPix2Pix~\citep{brooks2023instructpix2pix}, DiffEdit~\citep{couairon2023diffedit}, and Prompt-to-Prompt~\citep{hertz2022prompt} enabled global style changes and local attribute edits from text descriptions. More recent models such as ACE~\citep{han2024ace} and ACE++~\citep{han2025aceplus} introduce multimodal conditioning for more context-aware editing. These models improve instruction following, but they do not directly address the coarse-mask regime where the mask is only a spatial hint and the target object geometry must deviate from the marked contour.

\subsection{Attention and Layout Control in Diffusion}
Training-based layout control learns spatial conditioning from annotated layouts. LayoutDiffusion~\citep{zheng2023layoutdiffusion} trains a layout-to-image diffusion model with object-aware cross-attention that aligns bounding boxes and generated objects. BRIDGE instead targets coarse-mask local editing: its full-canvas and bbox-local paths are jointly trained, and a per-layer gate learns how subject-token coordinates interact with source-image context.

Training-free methods manipulate attention maps or positional embeddings in pretrained diffusion models. Prompt-to-Prompt~\citep{hertz2022prompt} and Plug-and-Play~\citep{tumanyan2023plug} use cross-attention or self-attention features for spatial and semantic control. BoxDiff~\citep{xie2023boxdiff} and LoCo~\citep{zhao2025loco} impose localized constraints for layout adherence, while MasaCtrl~\citep{cao2023masactrl}, PosBridge~\citep{xiong2025posbridge}, and RoPECraft~\citep{gokmen2025ropecraft} manipulate attention keys or positional embeddings to transfer structure, identity, or motion. BRIDGE learns a token-level PE routing policy from the current visual features instead of applying a fixed routing rule.

\subsection{Inpainting and Generative Filling}
Inpainting models are a strong baseline for local editing. LaMa~\citep{suvorov2022resolution} established high-resolution completion, while diffusion-based inpainters such as FLUX.1-Fill~\citep{labs2025flux} and BrushNet~\citep{ju2024brushnet} improve photorealistic filling under mask constraints. PowerPaint-style task prompting~\citep{zhuang2024task} further improves instruction alignment by conditioning on textual edits. Recent works like OminiControl~\citep{tan2025ominicontrol} introduce universal spatial control via in-context learning by concatenating image and mask inputs. In these families, the mask is typically compressed together with image content through a VAE or related encoder and then provided to the DiT as latent tokens, feature maps, or in-context visual conditions. Treating a coarse user mask as a geometric boundary can couple the final object shape to an irregular support region. BRIDGE assigns the mask to external support construction and blending, while BridgePath layout and PE routing govern geometry formation inside the DiT.

\section{Method}
\label{sec:method}

BRIDGE addresses coarse-mask local editing by separating two decisions that inpainting models often entangle: where the edit is allowed to occur, and how the new content should use the surrounding context. The mask defines the bbox-aligned Subject Path support and the external latent-blending region. BridgePath layout and learned positional-embedding (PE) routing control generation inside the DiT. We first formalize this separation through the Two-Zone Constraint, then introduce BridgePath generation, the Discrete Geometric Gate, the training objective, and the data-construction pipeline.

\subsection{Problem formulation and diagnostic insight}
\label{subsec:problem_insight}

Given a source image $I$, an instruction $y$, and a coarse user mask $M$, local editing can be described by two zones. The background zone $\Omega_b = 1-M$ should preserve appearance, texture, and layout. The editing zone $\Omega_e=M$ should satisfy the instruction while allowing the generated object to take a natural shape that may differ from the mask contour. In implementation, the coarse mask is converted to an axis-aligned patch-grid bounding box for Subject Path support, and the original mask remains available for optional blending. The mask localizes the user's intent while allowing the generated geometry to differ from its contour.

\begin{figure}[t]
  \centering
  \includegraphics[width=0.6\linewidth]{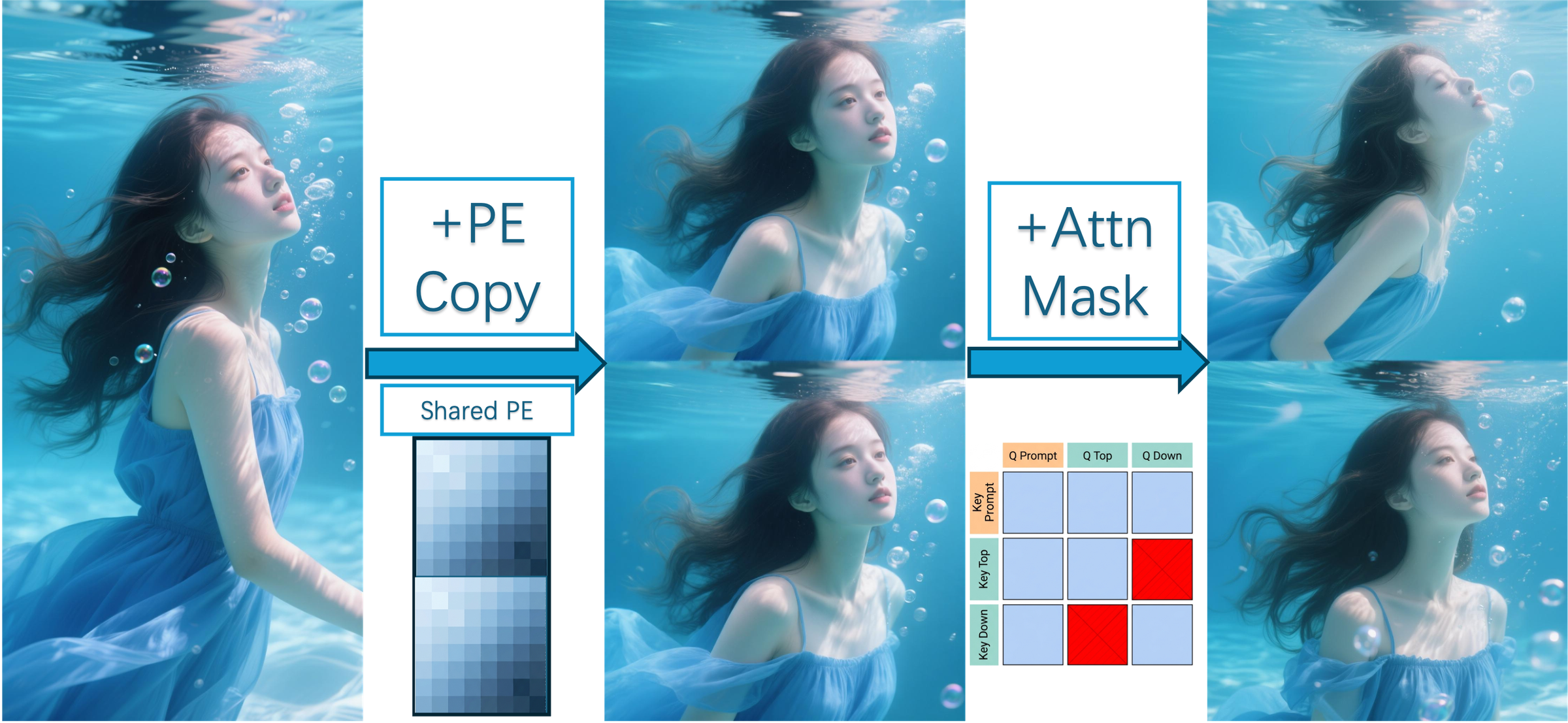}
  \caption{\textbf{Diagnostic Qwen-Image~\citep{qwenimage} experiment.} Positional embeddings determine which image context a token reuses. Sharing a PE across different visual tokens yields duplicated or highly similar content, while a cross-region attention mask suppresses this aliasing. This motivates BRIDGE's token-level PE switch for controlling where subject tokens borrow context from.}
  \label{fig:qwen_prior}
\end{figure}

We use a diagnostic experiment on Qwen-Image~\citep{qwenimage} to motivate PE routing. We assign the same PE to distinct visual tokens and observe that they tend to generate duplicated or highly similar content, as illustrated in Fig.~\ref{fig:qwen_prior}. This indicates that PE can act as a context selector: changing a token's coordinates changes which visual context the pre-trained generator is inclined to reuse. At the same time, applying a cross-region attention mask suppresses this duplication, showing that PE assignment and attention connectivity jointly determine both the source and the strength of contextual reuse. BRIDGE builds on this observation by training LoRA adapters and a token-level PE switch, so Subject Path tokens can decide where to borrow background context from and how strongly to use that context during generation.

\subsection{BridgePath generation and PE routing}
\label{subsec:bridgepath_routing}

\begin{figure}[t]
  \centering
  \includegraphics[width=\linewidth]{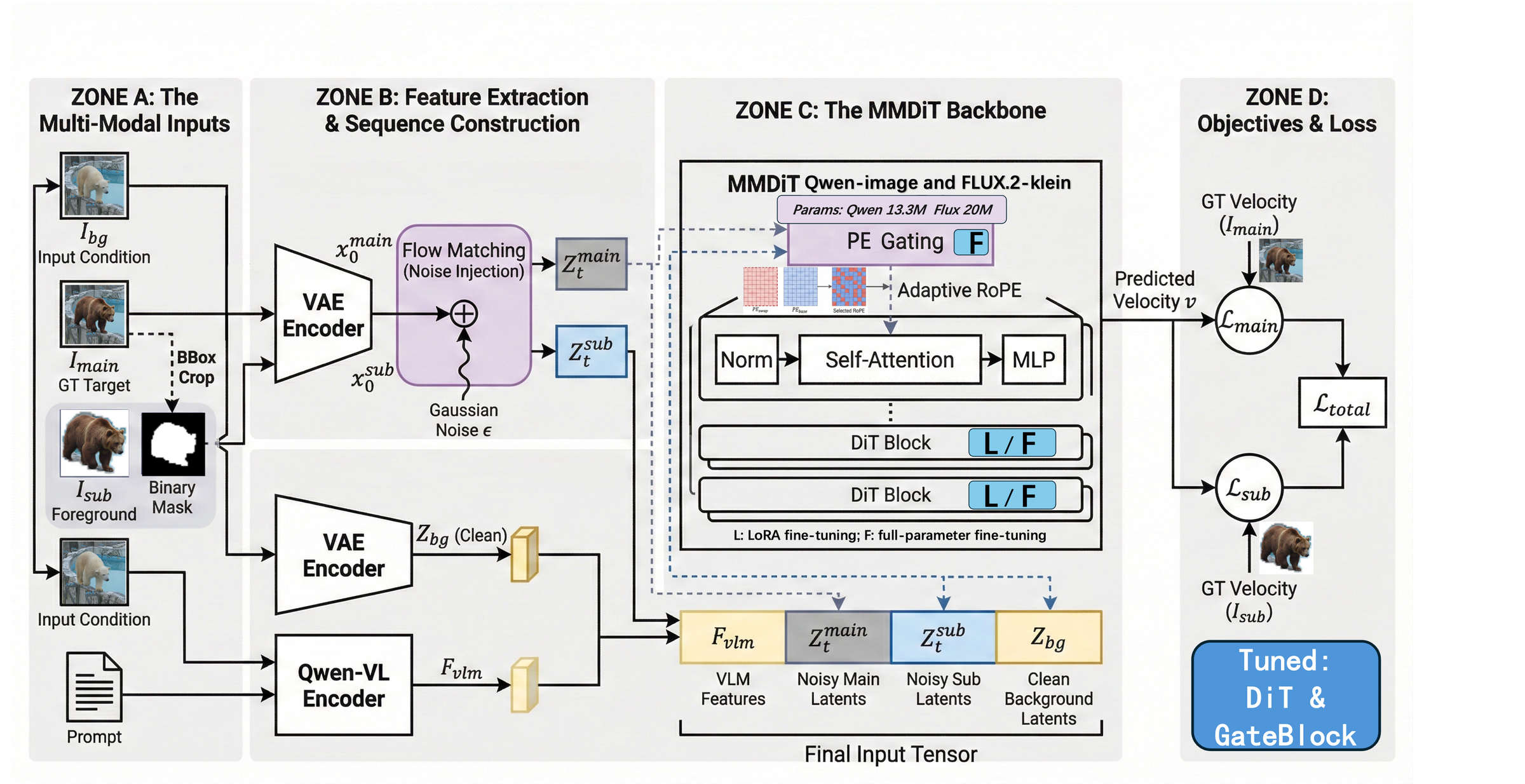}
  \vspace{4pt}
  \includegraphics[width=\linewidth]{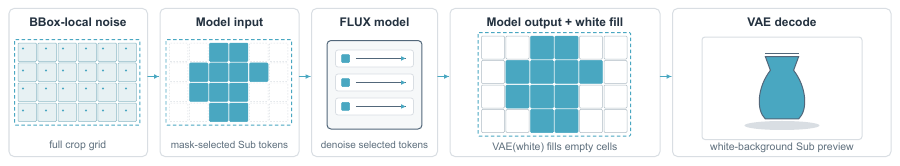}
  \caption{\textbf{BRIDGE overview and optional mask-cropped Sub inference.}
  \textbf{Top:} The Main Path preserves full-canvas context, while the Subject
  Path generates the editable region and the Discrete Geometric Gate routes
  subject-token positional embeddings. \textbf{Bottom:} During optional
  mask-cropped inference, Gaussian noise is initialized on the bbox-local Sub
  grid and only mask-selected tokens are retained and denoised. The sparse
  output is restored to its original crop-grid positions, and missing cells
  receive the VAE latents of an all-white crop before decoding. This operation
  belongs to the BRIDGE inference path and is independent of the specific
  backbone. The white-background Sub preview is diagnostic; the final edited
  image is decoded independently from the Main branch.}
  \label{fig:model}
\end{figure}

BRIDGE uses a BridgePath architecture processed by one multimodal DiT backbone~\citep{esser2024scaling}. The full-canvas Main Path generates the final edited image, while the bbox-local Subject Path generates the editable content. At inference, the two paths start from independently sampled Gaussian noise. Their tokens are concatenated with source-image condition tokens in the visual sequence and jointly processed with instruction tokens in each DiT block. Attention and PE routing determine how subject tokens use or separate from source-image context. Fig.~\ref{fig:model} summarizes the architecture; Appendix~\ref{sec:implementation_details} gives the training-field and supervision mapping.

The lower panel of Fig.~\ref{fig:model} shows BRIDGE's optional mask-cropped
Sub inference path. Appendix~\ref{sec:flux2_bbox_sparse_rationale} details the
training observations and inference design evaluated in the cross-backbone
extension.

The Subject Path uses background-anchored coordinates when scene alignment is useful and subject-centric coordinates when new structure is required. The learned gate makes this choice separately for each subject token and DiT layer.

\begin{figure}[!t]
  \centering
  \includegraphics[width=0.94\linewidth]{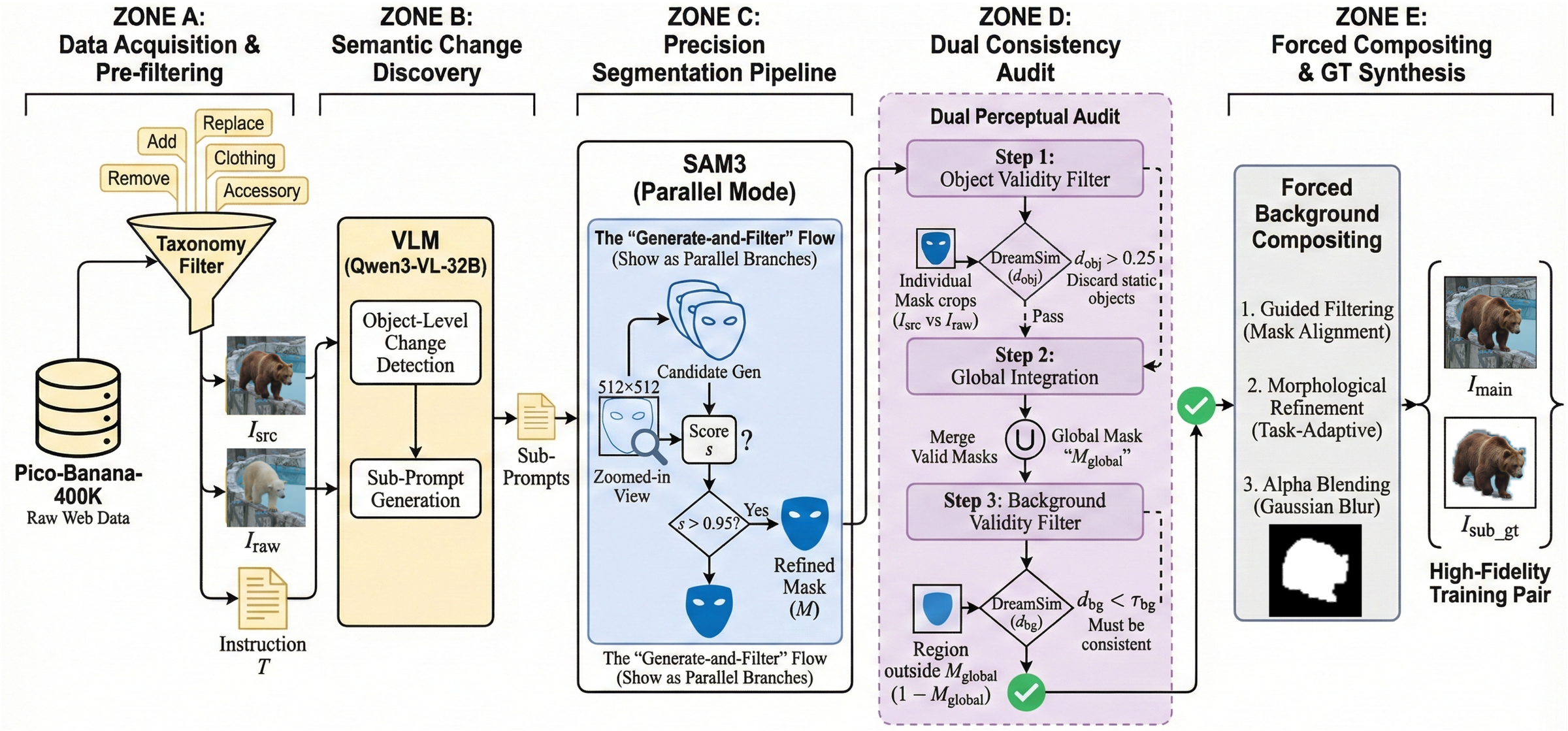}
  \caption{\textbf{Automated data pipeline for BRIDGE-Bench.} A VLM discovers local edit concepts, SAM proposes candidate masks, DreamSim filters for sufficient foreground change and limited background drift, and forced compositing restores the untouched scene. The resulting records provide the full edited target, source-image condition, subject target, and support mask used by the two-path training pipeline.}
  \label{fig:data_pipeline}
\end{figure}

\subsection{Discrete geometric gating}
\label{subsec:discrete_gate}

Complete isolation can yield floating or poorly integrated objects, while always using background coordinates can reintroduce mask-shape bias. We therefore introduce a Discrete Geometric Gate that predicts a binary routing variable for subject tokens in each DiT block.

For layer $l$, let $H^l = [Z_{\mathrm{main}}^l; Z_{\mathrm{sub}}^l; Z_{\mathrm{cond}}^l] \in \mathbb{R}^{N_l \times D}$ denote the visual sequence containing Main, Subject, and source-image condition tokens, with $D=3072$. Each layer has an independent GateBlock $h_\phi^l$. The GateBlock projects the visual sequence to a 64-dimensional hidden space, applies a single-layer Transformer encoder, and predicts a scalar logit for each subject-token index $i \in \mathcal{S}_l$, where $\mathcal{S}_l$ is the token range of the bbox-local Subject Path. The routing probabilities are
\begin{equation}
    p_i^l = \sigma\!\left(h_\phi^l(H^l)_i\right), \quad i \in \mathcal{S}_l,
\end{equation}
The threshold stays fixed at 0.5, with entropy and sparsity regularization weights set to zero. Its final linear head is zero-initialized. The hard forward decision and straight-through value are
\begin{equation}
    G_{\mathrm{hard},i}^l
    = \mathbf{1}\!\left[h_\phi^l(H^l)_i \geq 0\right]
    = \mathbf{1}\!\left[p_i^l \geq 0.5\right],
\end{equation}
\begin{equation}
    G_{\mathrm{ST},i}^l
    = G_{\mathrm{hard},i}^l + p_i^l - \operatorname{stopgrad}(p_i^l).
\end{equation}
The forward value remains binary, while the sigmoid probability supplies the surrogate gradient~\citep{bengio2013estimating}. The effective PE for subject tokens is
\begin{equation}
    PE_{\mathrm{eff},i}^l = G_{\mathrm{ST},i}^l \, PE_{\mathrm{base},i} + (1-G_{\mathrm{ST},i}^l) \, PE_{\mathrm{swap},i}, \quad i \in \mathcal{S}_l,
\end{equation}
where $PE_{\mathrm{base}}$ denotes the original Subject Path coordinates and $PE_{\mathrm{swap}}$ denotes background-anchored coordinates copied from the corresponding Main Path support. Thus, a forward value of 1 favors structural freedom, while 0 favors scene fusion. Zero logits give $p=0.5$, and the inclusive threshold initially selects $PE_{\mathrm{base}}$. Across a 60-layer DiT, the GateBlocks add about 13.31M parameters, compared with roughly 1.13B parameters for a ControlNet-style block copy under the same base-model dimensions. Additional implementation details are provided in Appendix~\ref{sec:implementation_details}.

\subsection{Training objective}
\label{subsec:training_objective}

BRIDGE is trained with the flow-matching objective used by modern DiT generators~\citep{esser2024scaling}. Let $Z_0 = z_0^{\mathrm{main}} \oplus z_0^{\mathrm{sub}}$ be the concatenated clean targets: $z_0^{\mathrm{main}}$ encodes the full background-preserved edited image, and $z_0^{\mathrm{sub}}$ encodes the foreground-on-white target crop within the Subject Path support. We form $Z_1=z_1^{\mathrm{main}}\oplus z_1^{\mathrm{sub}}$ by sampling $z_1^{\mathrm{main}}$ and $z_1^{\mathrm{sub}}$ independently from matching Gaussian distributions. For $t\sim\mathcal{U}(0,1)$,
\begin{equation}
    Z_t = tZ_1 + (1-t)Z_0.
\end{equation}
The network predicts the velocity field over the concatenated Main/Subject sequence, conditioned on the instruction embedding $T$, timestep $t$, and source-image condition $z_{\mathrm{edit}}$:
\begin{equation}
    \mathcal{L}_{fm} =
    \mathbb{E}\left[
    \left\|
    v_\theta(Z_t,T,t,z_{edit}) - (Z_1-Z_0)
    \right\|_2^2
    \right].
\end{equation}
We optimize the LoRA adapters~\citep{hu2022lora} and GateBlocks end-to-end as a joint generation problem over the Main and Subject paths. The user mask supplies bbox construction and inference-time blending support. Appendix~\ref{sec:implementation_details} specifies the complete tensor and supervision flow.

\subsection{Data construction pipeline}
\label{subsec:data_pipeline}

Training requires examples where the foreground changes while the background remains stable. We build an internally processed dataset from Pico-Banana-400K~\citep{Gan2025PicoBanana400K}: a VLM discovers local edit concepts, SAM proposes candidate masks~\citep{carion2025sam}, DreamSim filters for sufficient object change and limited background drift~\citep{fu2023dreamsim}, and forced compositing preserves the untouched scene. Fig.~\ref{fig:data_pipeline} summarizes this construction flow. The processed pool defines the training split and the BRIDGE-Bench test dataset; full thresholds, target construction, and evaluation preprocessing are provided in Appendix~\ref{sec:implementation_details}.

\section{Experiments}
\label{sec:experiments}

We evaluate local precision, global quality, transfer, ablations, and cost.

\subsection{Qwen-BRIDGE Results}

\paragraph{Setup.}
Our processed Pico-Banana-400K~\citep{Gan2025PicoBanana400K} pool contains
42,425 training pairs, and the BRIDGE-Bench test dataset contains 1,444 pairs.
Both splits follow the same filtering and compositing pipeline. We also evaluate
zero-shot transfer on MagicBrush~\citep{zhang2023magicbrush} and
ICE-Bench~\citep{pan2025icebench}. External baselines use released checkpoints
and settings; internal ablations share the Qwen-Image-Edit-2511~\citep{qwenimage}
backbone, LoRA configuration, data, optimizer, hardware, and budget. We report
FLUX.1-Fill~\citep{labs2025flux}, ACE/ACE++~\citep{han2024ace,han2025aceplus},
UltraEdit~\citep{zhao2024ultraedit}, Q-Control~\citep{diffsynth2025qwencontrol},
DINOv3~\citep{simeoni2025dinov3}, SigLIP2~\citep{tschannen2025siglip2},
DreamSim~\citep{fu2023dreamsim}, and CLIP~\citep{radford2021clip}. Local
BRIDGE-Bench metrics use the edit bounding box; full protocols and training
details are in the Appendix.

\paragraph{BRIDGE-Bench.}

On the BRIDGE-Bench test dataset, BRIDGE leads all reported local metrics in Table~\ref{tab:combined_results},
including SigLIP2-T (0.503), DINO (0.902), and DreamSim (0.175).

\begin{table}[t]
    \centering
    \caption{BRIDGE-Bench test-dataset results. Local metrics use the edit bounding box; lower DreamSim is better and all other metrics are higher-is-better.}
    \label{tab:combined_results}
    \resizebox{\linewidth}{!}{
        \setlength{\tabcolsep}{2pt}
        \begin{tabular}{l|cccc|cccc}
            \toprule
            & \multicolumn{4}{c|}{\textbf{Local Precision}} & \multicolumn{4}{c}{\textbf{Global Quality (BRIDGE-Bench)}} \\
            \textbf{Method} & \textbf{Sig-T}$\uparrow$ & \textbf{DINO}$\uparrow$ & \textbf{DSim}$\downarrow$ & \textbf{Sig-I}$\uparrow$ & \textbf{DINO}$\uparrow$ & \textbf{DSim}$\downarrow$ & \textbf{Sig-I}$\uparrow$ & \textbf{Sig-T}$\uparrow$ \\
            \midrule
            ACE++~\citep{han2025aceplus} & 0.390 & 0.852 & 0.303 & 0.856 & 0.982 & 0.081 & 0.954 & 0.705 \\
            FLUX.1-Fill~\citep{labs2025flux} & 0.262 & 0.807 & 0.321 & 0.835 & 0.978 & 0.076 & 0.947 & 0.632 \\
            \textbf{BRIDGE} & \textbf{0.503} & \textbf{0.902} & \textbf{0.175} & \textbf{0.907} & \textbf{0.988} & \textbf{0.054} & \textbf{0.965} & \textbf{0.776} \\
            \bottomrule
        \end{tabular}
    }
\end{table}

\begin{table}[t]
    \centering
    \caption{MagicBrush zero-shot results on full images using current-turn local descriptions (current protocol; all $\uparrow$). SigLIP2-T is the model-native sigmoid score defined in Eq.~\ref{eq:siglip_native}.}
    \label{tab:magicbrush_results}
    \resizebox{\linewidth}{!}{
        \setlength{\tabcolsep}{2.1pt}
        \begin{tabular}{l|ccccc|ccccc}
        \toprule
        & \multicolumn{5}{c|}{\textbf{Final Turn}} & \multicolumn{5}{c}{\textbf{All Turn}} \\
        \textbf{Method} & \textbf{CLIP-I} & \textbf{CLIP-T Local} & \textbf{DINOv3} & \textbf{SigLIP2-I} & \textbf{SigLIP2-T Local} & \textbf{CLIP-I} & \textbf{CLIP-T Local} & \textbf{DINOv3} & \textbf{SigLIP2-I} & \textbf{SigLIP2-T Local} \\
        \midrule
        ACE++ & 0.8909 & 0.2819 & 0.9464 & 0.8036 & 0.3612 & 0.9362 & 0.2824 & 0.9706 & 0.8871 & 0.4177 \\
        FLUX.1-Fill & 0.8979 & 0.2701 & 0.9503 & 0.8058 & 0.2691 & \textbf{0.9466} & 0.2738 & 0.9742 & 0.8969 & 0.3399 \\
        Q-Control & 0.8984 & 0.2934 & 0.9565 & \textbf{0.8446} & 0.4797 & 0.9425 & 0.2908 & 0.9765 & 0.9067 & 0.5081 \\
        \textbf{BRIDGE} & \textbf{0.9070} & \textbf{0.2959} & \textbf{0.9587} & 0.8438 & \textbf{0.5584} & 0.9463 & \textbf{0.2925} & \textbf{0.9775} & \textbf{0.9070} & \textbf{0.5593} \\
        \bottomrule
        \end{tabular}
    }
\end{table}

\begin{table}[!b]
    \centering
    \caption{ICE-Bench results grouped by input/prompt protocol; comparisons are made within each block. The left block uses the orange data set, where we replace \texttt{<SOURCE>} with ``Picture 1'' and describe the masked region as the black area. The right block uses the original ICE-Bench inputs and prompts, which may include \texttt{<Mask>} and \texttt{<SOURCE>}.}
    \label{tab:icebench_results}
    \small
    \resizebox{\linewidth}{!}{
    \begin{tabular}{lcccc|lcccc}
        \toprule
        \multicolumn{5}{c|}{\textbf{Orange data set (modified inputs/prompts)}} & \multicolumn{5}{c}{\textbf{Original ICE-Bench inputs/prompts}} \\
        Model & AES$\uparrow$ & IMG$\uparrow$ & PF$\uparrow$ & SRC$\uparrow$ & Model & AES$\uparrow$ & IMG$\uparrow$ & PF$\uparrow$ & SRC$\uparrow$ \\
        \midrule
        ACE++  & 0.480 & \textbf{0.575} & 0.575 & 0.925 & ACE & 0.480 & 0.487 & \textbf{0.643} & 0.916 \\
        Q-Control & 0.472 & 0.569 & \textbf{0.588} & 0.937 & ACE++ & \textbf{0.481} & \textbf{0.577} & 0.616 & 0.930 \\
        \textbf{BRIDGE} & \textbf{0.494} & 0.565 & 0.582 & \textbf{0.942} & UltraEdit & 0.455 & 0.458 & 0.442 & \textbf{0.951} \\
        \bottomrule
    \end{tabular}
    }
\end{table}

\begin{table}[t]
    \centering
    \caption{BRIDGE-Bench test-dataset ablation. The mask-free row disables external blending ($\alpha=0$); trained variants share the backbone, LoRA, data, optimizer, and budget.}
    \label{tab:ablation_study}
    \resizebox{\linewidth}{!}{
        \setlength{\tabcolsep}{3pt}
        \begin{tabular}{lcccccccc}
            \toprule
            & \multicolumn{4}{c}{\textbf{Global Quality}} & \multicolumn{4}{c}{\textbf{Local Precision}} \\
            \cmidrule(lr){2-5} \cmidrule(lr){6-9}
            Config & DINO $\uparrow$ & DreamSim $\downarrow$ & SigLIP2-I $\uparrow$ & SigLIP2-T $\uparrow$ & SigLIP2-T $\uparrow$ & DINO $\uparrow$ & DreamSim $\downarrow$ & SigLIP2-I $\uparrow$ \\
            \midrule
            Mask-free Baseline ($\alpha=0$) & 0.977 & 0.087 & 0.949 & 0.754 & 0.453 & 0.878 & 0.227 & 0.890 \\
            Baseline & 0.986 & 0.065 & 0.957 & 0.773 & 0.464 & 0.870 & 0.239 & 0.881 \\
            w/o Discrete Gate & 0.979 & 0.089 & 0.946 & \textbf{0.784} & 0.463 & 0.875 & 0.237 & 0.885 \\
            \textbf{BRIDGE} & \textbf{0.988} & \textbf{0.054} & \textbf{0.965} & 0.776 & \textbf{0.503} & \textbf{0.902} & \textbf{0.175} & \textbf{0.907} \\
            \bottomrule
        \end{tabular}
    }
\end{table}

\begin{figure}[t]
  \centering
  \makebox[0.06\linewidth][l]{\scriptsize\bfseries Main}%
  \includegraphics[width=0.94\linewidth,trim=0 46bp 0 0,clip]{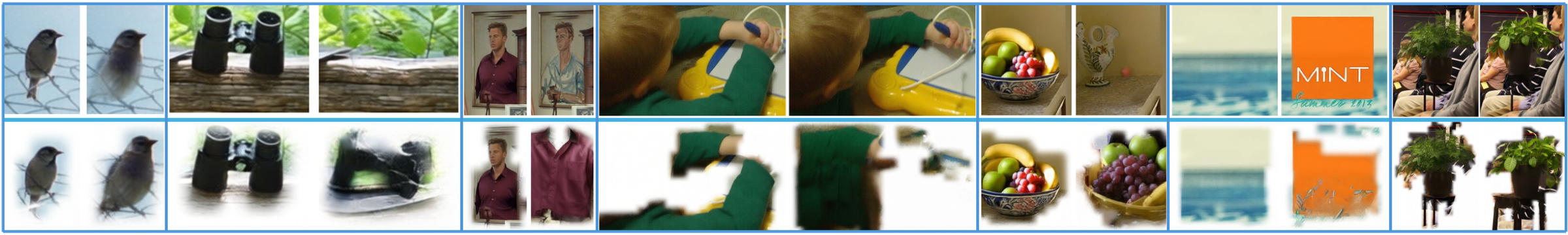}
  \vspace{1pt}
  \makebox[0.06\linewidth][l]{\scriptsize\bfseries Sub}%
  \includegraphics[width=0.94\linewidth,trim=0 0 0 46bp,clip]{abl.png}
  \caption{\textbf{Visual identity of the routing ablation.} The top and bottom rows show Main and Sub outputs, respectively. Within each example pair, Full BRIDGE is on the left and BridgePath with fixed PE is on the right. Adaptive routing yields sharper Sub content in these examples, consistent with Table~\ref{tab:ablation_study}.}
  \label{fig:subject_path_ablation}
\end{figure}

\begin{figure}[!t]
\centering
\includegraphics[width=\linewidth,height=0.74\textheight,keepaspectratio]{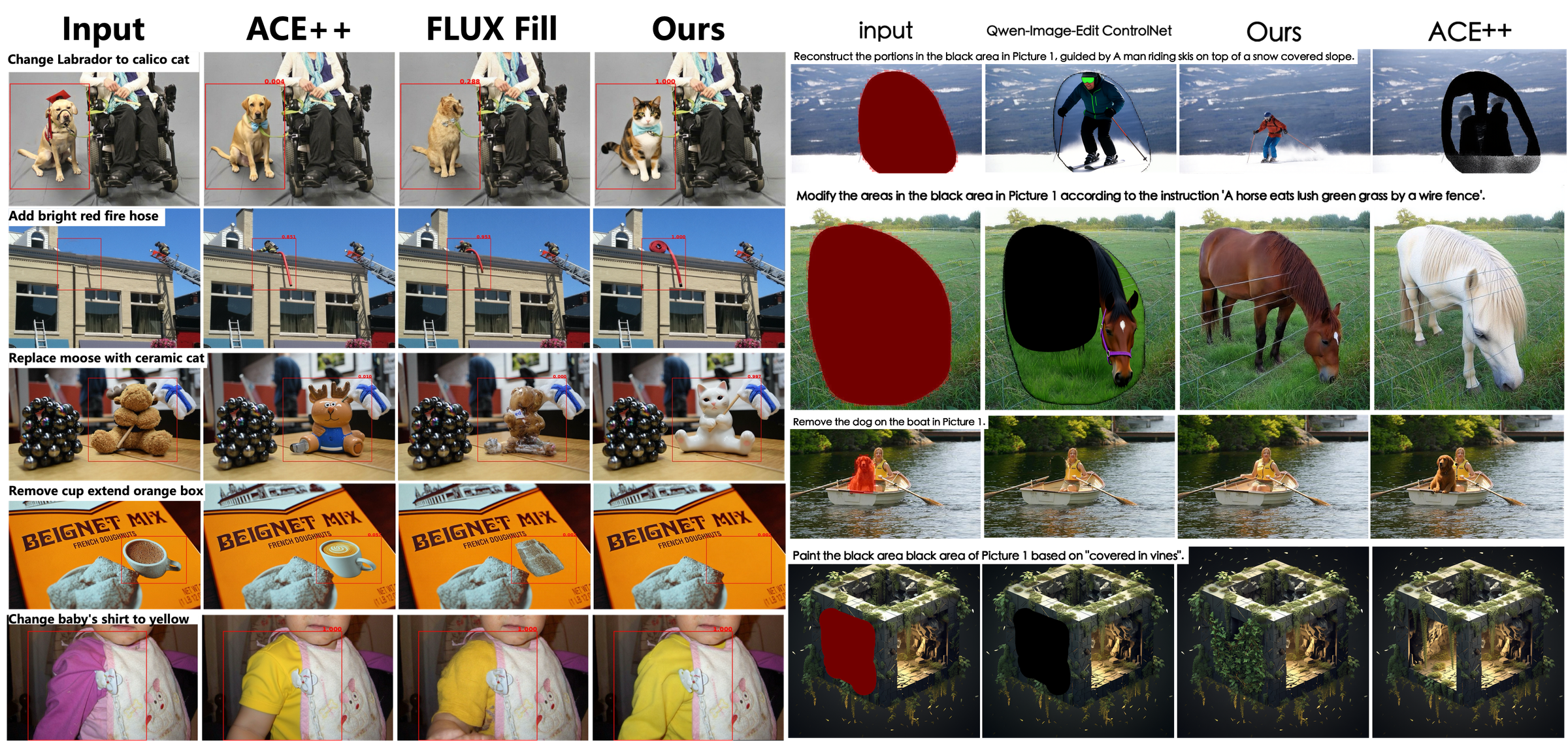}
\caption{\textbf{BRIDGE improves prompt-consistent local generation under coarse masks.} Left: our constructed test split. Right: ICE-Bench examples. Compared with Q-Control and ACE++, BRIDGE more often follows prompts, generates plausible local geometry, and blends edits cleanly into the scene.}
\label{fig:qualitative_comparison}
\vspace{-8pt}
\end{figure}

\FloatBarrier

\subsection{FLUX.2 Cross-Backbone Results}
\label{sec:flux2_extension}

We instantiate BRIDGE on FLUX.2 [klein] 9B Base~\citep{bfl2026flux2kleinbase9b} with full-transformer training,
layer-wise gates, and subject-reference conditioning. Main predicts the full
canvas and Sub uses an aligned crop. BBox supervision retains Sub quality;
optional inference-time token cropping tightens mask support (Fig.~\ref{fig:model}).
Appendix~\ref{sec:flux2_token_control_evidence} reports the irregular-mask stress
test and indexes the full STE animations in the supplement.

For ICE-Bench~\citep{pan2025icebench}, GPT-6 Luna rewrites placeholder-based
instructions into self-contained prompts. We evaluate the step-1,400 Dense/BBox
checkpoint with CFG 4 and 50 steps; Appendix~\ref{sec:flux2_grouped_comparison_provenance}
reports the protocol and per-task scores. Task-by-task visual results from the
same Dense/BBox run are shown in Appendix~\ref{sec:flux2_qualitative_showcase}.

\begin{table}[H]
    \centering
    \caption{\textbf{ICE-Bench mask-group comparison.} Original-model scores
    are reconstructed from ICE-Bench~\citep{pan2025icebench}; later scores are
    from Emu3.5~\citep{cui2025emu35}. $\dagger$ Ours was not trained for
    Inpainting (Task 17), Outpainting (Task 18), or Face Swap (Task 31); these
    are diagnostic. For Task 17, the model receives the original source image
    and a separate \texttt{SourceMask}; Luna rewrites the prompt from the source
    and mask. Details are in the Appendix.}
    \label{tab:flux2_grouped_comparison}
    \textbf{Open-source models}\\[1pt]
    \textbf{Original ICE-Bench models}\\[-1pt]
    {\setlength{\tabcolsep}{4pt}%
    \begin{tabular}{lccccc}
        \toprule
        \textbf{Model} & \textbf{19--22} & \textbf{17} & \textbf{18} & \textbf{29--30} & \textbf{31} \\
        \midrule
        ACE++~\citep{han2025aceplus} & 0.5830 & 0.6430 & \textbf{0.6230} & 0.6066 & 0.5396 \\
        ACE~\citep{han2024ace} & 0.5695 & 0.5954 & 0.5528 & 0.5722 & 0.4907 \\
        OmniGen~\citep{xiao2025omnigen} & 0.5364 & 0.5454 & 0.5733 & 0.5759 & 0.5324 \\
        \textbf{Ours (FLUX.2 BBox-Sub)$\boldsymbol{\dagger}$} & \textbf{0.5941} & \textbf{0.6506} & 0.5801 & \textbf{0.6254} & \textbf{0.5700} \\
        \bottomrule
    \end{tabular}}
    \vspace{4pt}

    \textbf{Later open-source models}\\[-1pt]
    {\setlength{\tabcolsep}{8pt}%
    \begin{tabular}{lccc}
        \toprule
        \textbf{Model} & \textbf{17--22} & \textbf{29--30} & \textbf{31} \\
        \midrule
        Qwen-Edit-2509~\citep{qwenimage} & 0.587 & 0.595 & 0.539 \\
        Emu3.5~\citep{cui2025emu35} & \textbf{0.613} & 0.624 & 0.528 \\
        \textbf{Ours (FLUX.2 BBox-Sub)$\boldsymbol{\dagger}$} & 0.6012 & \textbf{0.6254} & \textbf{0.5700} \\
        \bottomrule
    \end{tabular}}
    \vspace{4pt}

    \textbf{Closed models: architectures and training recipes undisclosed}\\[-1pt]
    {\setlength{\tabcolsep}{8pt}%
    \begin{tabular}{lccc}
        \toprule
        \textbf{Closed model} & \textbf{17--22} & \textbf{29--30} & \textbf{31} \\
        \midrule
        Gemini 2.5 Flash Image Preview~\citep{cui2025emu35} & 0.615 & 0.617 & 0.565 \\
        Gemini 2.5 Flash Image~\citep{cui2025emu35} & \textbf{0.617} & \textbf{0.627} & \textbf{0.595} \\
        \bottomrule
    \end{tabular}}
\end{table}

Among compared open-source models, FLUX.2 BRIDGE leads subject-guided editing
(Tasks 29--30, 0.6254) and the diagnostic Face Swap evaluation (0.5700).
Its four supported local-edit tasks average 0.5941. Later papers report only the
combined Tasks 17--22 score; Gemini references are separated as closed models.
With the prompt fixed, BRIDGE remains reliable across Local Subject Addition
mask variants; in the Add-a-glass case, it consistently inserts a plausible
glass, while ACE++ is inconsistent and Q-Control fails
(Appendix~\ref{sec:flux_mask_intervention}).


\FloatBarrier
\section{Conclusion}
\label{sec:conclusion}

BRIDGE separates spatial support from subject geometry with full-canvas Main and local Subject paths plus learned token-level PE routing. Main preserves the scene, Subject generates local content, and optional mask-cropped inference tightens irregular support. Qwen results improve local precision and transfer to MagicBrush and ICE-Bench, while the FLUX.2 instantiation extends the design to full-transformer, subject-reference training.

The model currently assumes a mask. Without one, full-canvas Subject support makes both paths generate complete canvases and reduces quality relative to the base model. Disabling Subject does not restore mask-free generation: in our no-Sub diagnostic, Main showed no visible prompt-driven change. Future work will unify masked editing and mask-free generation in one model.

\clearpage
\subsection*{Reproducibility statement}

Sections~\ref{sec:method}--\ref{sec:experiments} and the appendix document the
model, training objective, data construction, fixed training configuration,
evaluation protocols, ablations, and supporting results.

\bibliographystyle{plainnat}
\bibliography{references}

\appendix
\section{Implementation Details}
\label{sec:implementation_details}

\paragraph{Training Details.}
The reported Qwen BRIDGE model uses the pre-trained \textbf{Qwen-Image-Edit-2511} transformer~\citep{qwenimage}. Rank-\textbf{512} LoRA~\citep{hu2022lora} adapts selected attention, image/text feed-forward, modulation, and input/output projection layers, while the \textbf{GateBlocks} are trained from scratch. The final linear head of each GateBlock is zero-initialized. We use the \textbf{Prodigy}~\citep{mishchenko2024prodigy} optimizer with a Schedule-Free learning-rate policy~\citep{Defazio2024Road} (initial LR=1.0). Training uses \textbf{5 $\times$ NVIDIA RTX PRO 6000 Blackwell (96GB)} GPUs for \textbf{30K micro-steps}, corresponding to about \textbf{1.875K optimizer updates}. Gradient accumulation is 16 per process, giving an effective total batch size of $5 \times 16=80$ and about 3.54 passes over the 42,425-pair training set ($1{,}875 \times 80 / 42{,}425 \approx 3.54$). The final fine-tuning checkpoint contains \textbf{4,725,211,136} LoRA parameters and \textbf{13,313,340} GateBlock parameters. Classifier-Free Guidance (CFG) training~\citep{ho2022classifier} drops the conditioning text with probability 10\%. Before deployment, the LoRA weights are fused into the backbone. Runtime overhead then comes from BridgePath, primarily the Subject Path.

\paragraph{Training Data Flow and Path Supervision.}
Each processed record separates the source-image condition from the two generation targets. The \texttt{image} field is the full background-preserved edited target for the Main Path. The \texttt{edit\_image} field supplies the source image to the Qwen visual-conditioning stream. Forced compositing uses source-guided boundary refinement, erosion for removal, dilation for addition and replacement, $33\!\times\!33$ Gaussian smoothing, and alpha blending. The refined soft support extracts the edited foreground onto white; its crop is rounded outward to a 16-pixel grid and encoded by the frozen VAE as \texttt{sub\_input\_latents} for Subject Path supervision. Table~\ref{tab:training_data_flow} summarizes the complete mapping.

\begin{table}[H]
    \centering
    \caption{Training-data and tensor flow for the reported Qwen BRIDGE model. Main and Subject targets receive independently sampled Gaussian noise at the same sampled timestep.}
    \label{tab:training_data_flow}
    \scriptsize
    \setlength{\tabcolsep}{2pt}
    \begin{tabular}{p{0.18\linewidth}p{0.28\linewidth}p{0.25\linewidth}p{0.21\linewidth}}
        \toprule
        \textbf{Field/tensor} & \textbf{Semantic and spatial content} & \textbf{Training role} & \textbf{Inference role} \\
        \midrule
        \texttt{prompt} / $T$ & Edit instruction; text sequence & Joint DiT conditioning & Joint DiT conditioning \\
        \texttt{edit\_image} / $z_{\mathrm{edit}}$ & Source image; full image & Clean visual-condition tokens & Source-image condition \\
        \texttt{image} / $z_0^{\mathrm{main}}$ & Background-preserved edited target; full canvas & Main target with noise $z_1^{\mathrm{main}}$ & Replaced by Main noise \\
        \texttt{ref\_gt} / $z_0^{\mathrm{sub}}$ & Edited foreground on white; 16-pixel-aligned support crop & Subject target with noise $z_1^{\mathrm{sub}}$ & Replaced by Subject noise \\
        \texttt{back\_mask} & Edit support; full image and aligned bbox & Defines Subject support & Defines support and blending \\
        Main output & Final edited image; full canvas & Main flow-matching prediction & Returned and decoded \\
        Subject output & Local subject representation; bbox-local canvas & Subject flow-matching prediction & Decoded for diagnostics \\
        \bottomrule
    \end{tabular}
\end{table}

The Main and Subject targets are noised separately: $z_1^{\mathrm{main}}\sim\mathcal{N}(0,I)$ and $z_1^{\mathrm{sub}}\sim\mathcal{N}(0,I)$ are independent draws. At each DiT block, instruction tokens are jointly processed with a visual sequence ordered as Main tokens, Subject tokens, and source-image condition tokens. The GateBlock observes this visual sequence and produces routing values used for Subject-token PE selection. At inference, the pipeline returns the decoded Main Path as the edited image; the Subject Path is also decoded for analysis. External background blending is applied to the Main latent outside the selected support after each denoising update.

\paragraph{Controlled Ablation Budget.}
All trained variants in Table~\ref{tab:ablation_study} use the same Qwen-Image-Edit-2511 backbone, rank-512 LoRA configuration, BRIDGE training set, optimizer and hyperparameters, 5 $\times$ 96GB hardware configuration, and 30K-micro-step budget. The Baseline uses a single generation path. The ``w/o Discrete Gate'' variant adds BridgePath with fixed positional assignments, and BRIDGE adds adaptive per-token PE routing.

\paragraph{Supplementary Ablation Study.}
Table~\ref{tab:supplementary_data_control} presents the complete supplementary
ablation study on BRIDGE-Bench. It adds the earlier single-path data control to
the four inference, architecture, and routing configurations reported in the
main-paper ablation.

\begin{table}[H]
    \centering
    \caption{Complete supplementary ablation study on BRIDGE-Bench. The table combines the earlier single-path data control with the mask-free inference, single-path, fixed-routing, and adaptive-routing configurations. Lower DreamSim is better; all other metrics are higher-is-better.}
    \label{tab:supplementary_data_control}
    \resizebox{\linewidth}{!}{%
    \setlength{\tabcolsep}{3pt}
    \begin{tabular}{lcccccccc}
        \toprule
        & \multicolumn{4}{c}{\textbf{Global Quality}} & \multicolumn{4}{c}{\textbf{Local Precision}} \\
        \cmidrule(lr){2-5} \cmidrule(lr){6-9}
        \textbf{Configuration} & DINO $\uparrow$ & DreamSim $\downarrow$ & SigLIP2-I $\uparrow$ & SigLIP2-T $\uparrow$ & SigLIP2-T $\uparrow$ & DINO $\uparrow$ & DreamSim $\downarrow$ & SigLIP2-I $\uparrow$ \\
        \midrule
        Earlier single-path data control & 0.986 & 0.057 & 0.959 & 0.712 & 0.402 & 0.864 & 0.263 & 0.873 \\
        Mask-free Baseline ($\alpha=0$) & 0.977 & 0.087 & 0.949 & 0.754 & 0.453 & 0.878 & 0.227 & 0.890 \\
        Baseline & 0.986 & 0.065 & 0.957 & 0.773 & 0.464 & 0.870 & 0.239 & 0.881 \\
        w/o Discrete Gate & 0.979 & 0.089 & 0.946 & \textbf{0.784} & 0.463 & 0.875 & 0.237 & 0.885 \\
        \textbf{BRIDGE} & \textbf{0.988} & \textbf{0.054} & \textbf{0.965} & 0.776 & \textbf{0.503} & \textbf{0.902} & \textbf{0.175} & \textbf{0.907} \\
        \bottomrule
    \end{tabular}}
\end{table}

The PE diagnostic uses Qwen-Image with 40 inference steps, seed 0, and CFG 4.
The reported Qwen BRIDGE results use Qwen-Image-Edit-2511 with the training
configuration above, CFG 2, and latent-blending strength $\alpha=0.1$ unless a
benchmark-specific protocol states otherwise. MagicBrush and ICE-Bench follow
their respective protocols below.

Table~\ref{tab:parameter_breakdown} gives the parameter breakdown of the final
fine-tuning checkpoint.

\begin{table}[H]
    \centering
    \caption{Parameter breakdown of the final BRIDGE fine-tuning checkpoint. The four LoRA rows sum to the reported LoRA total.}
    \label{tab:parameter_breakdown}
    \small
    \begin{tabular}{lr}
        \toprule
        \textbf{Component} & \textbf{Parameters} \\
        \midrule
        Attention-projection LoRA & 1,509,949,440 \\
        Image/text feed-forward LoRA & 1,887,436,800 \\
        Modulation-projection LoRA & 1,321,205,760 \\
        Input/output-projection LoRA & 6,619,136 \\
        \textbf{LoRA total} & \textbf{4,725,211,136} \\
        GateBlocks & 13,313,340 \\
        \textbf{All trained parameters} & \textbf{4,738,524,476} \\
        \bottomrule
    \end{tabular}
\end{table}

\paragraph{STE parameter accounting across backbones.}
The straight-through routing module has a backbone-dependent parameterization.
The counts below include only learned STE/GateBlock parameters; the threshold
buffer $\tau$ is not learnable and is therefore excluded from the STE total.
If checkpoint tensor storage is counted instead, the single scalar buffer adds
one element. For a model with $L$ transformer layers, the learned parameter
count is
\[
P_{\mathrm{Qwen}}(L)=221{,}889L,\qquad
P_{\mathrm{FLUX.2\ Klein}}(L)=624{,}001L.
\]
Thus, including the scalar $\tau$ buffer in checkpoint tensor storage gives
13,313,341 elements for Qwen-Image and 19,968,033 for either FLUX.2 Klein
variant.

Table~\ref{tab:ste_parameter_accounting} summarizes the backbone-specific
learned parameter counts.

\begin{table}[H]
    \centering
    \caption{Backbone-specific parameter accounting for the straight-through
    routing module. Sparse/Mask and Dense/BBox FLUX.2 variants use the same
    GateBlock shape and therefore have the same parameter count.}
    \label{tab:ste_parameter_accounting}
    \small
    \setlength{\tabcolsep}{4pt}
    \begin{tabular}{lrrrrr}
        \toprule
        \textbf{Backbone} & \textbf{Layers} & \textbf{Input dim.} & \textbf{STE dim.} & \textbf{Per-layer} & \textbf{STE total} \\
        \midrule
        Qwen-Image & 60 & 3,072 & 64 & 221,889 & 13,313,340 \\
        FLUX.2 Klein Sparse/Mask & 32 & 4,096 & 128 & 624,001 & 19,968,032 \\
        FLUX.2 Klein Dense/BBox & 32 & 4,096 & 128 & 624,001 & 19,968,032 \\
        \bottomrule
    \end{tabular}
\end{table}

\paragraph{GateBlock Architecture.}
At layer $l$, the gate observes $H^l=[Z_{\mathrm{main}}^l;Z_{\mathrm{sub}}^l;Z_{\mathrm{cond}}^l]\in\mathbb{R}^{N_l\times3072}$, the visual sequence containing Main, Subject, and source-image condition features. Each of the 60 DiT layers has its own GateBlock with independent parameters. A GateBlock projects from 3072 to 64 channels, applies a single-layer Transformer encoder, and predicts scalar logits whose Subject-token entries control PE routing. Subject tokens occupy the bbox-local range on the Subject Path canvas. The hard forward decision uses the inclusive 0.5 threshold, and the sigmoid probability supplies the straight-through gradient~\citep{bengio2013estimating}.

\paragraph{Gate Initialization and Visual-Ablation Identity.}
Zero-initialized gate heads produce $p=0.5$. With the inclusive threshold, the initial hard forward route selects $PE_{\mathrm{base}}$; the straight-through estimator changes gradients only. Figure~\ref{fig:subject_path_ablation} labels the Main/Subject output strips and maps the left/right image in every pair to Full BRIDGE/fixed-PE BridgePath.

\paragraph{Inference Configuration.}
The evaluation settings are as follows:
\begin{itemize}
    \item \textbf{Inference Parameters:} Our default evaluation uses a classifier-free guidance (CFG)~\citep{ho2022classifier} scale of 2.0. MagicBrush follows a separate protocol: the main-paper BRIDGE result uses CFG$=4$ with mask-based blending, and Table~\ref{tab:magicbrush_extra_variants} reports additional model and inference settings, including the Qwen-Image-edit ablation baseline with CFG$=4$. To mitigate the potential saturation and over-exposure artifacts derived from high guidance scale, we apply the CFG Rescaling trick~\citep{lin2024common}. Specifically, we rescale the guided noise prediction to match the norm of the conditional noise prediction: $\epsilon_{pred} \leftarrow \epsilon_{pred} \cdot \frac{\|\epsilon_{pos}\|}{\|\epsilon_{pred}\|}$. Here, $\epsilon_{pos}$ denotes the conditional noise prediction. Baseline models (e.g., FLUX~\citep{labs2025flux}) are evaluated using their respective official default inference settings.
    \item \textbf{Background Preservation:} The user mask determines the support for latent blending~\citep{avrahami2023blendedlatent}. Unless otherwise noted, experiments use bounding-box support $M_{bbox}$; the main MagicBrush result uses mask-based support, and Table~\ref{tab:magicbrush_extra_variants} includes a bbox-support variant. The default blending strength is $\alpha = 0.1$, where larger $\alpha$ retains more source-image background. For bbox support, each denoising step blends the Main latent outside the bounding box ($1-M_{bbox}$) with source-image latents noised to the same timestep:
    \begin{equation}
        z_{t-1}^{blend} = z_{t-1} \cdot M_{bbox} + \big((1 - \alpha) \cdot z_{t-1} + \alpha \cdot z_{t-1}^{orig}\big) \cdot (1 - M_{bbox})
    \end{equation}
    where $z_{t-1}^{orig}$ denotes the source-image latents corrupted to timestep $t-1$. This operation preserves background context while allowing generation throughout the bbox support.
\end{itemize}

\paragraph{Figure~\ref{fig:qwen_prior} Diagnostic Setting.}
Figure~\ref{fig:qwen_prior} uses the official DiffSynth-Studio Qwen-Image Quick Start setting: model \texttt{Qwen/Qwen-Image}, prompt ``\begin{CJK*}{UTF8}{gbsn}精致肖像，水下少女，蓝裙飘逸，发丝轻扬，光影透澈，气泡环绕，面容恬静，细节精致，梦幻唯美。\end{CJK*}'', seed 0, 40 inference steps, and the pipeline-default CFG of 4.0. The panels vary only PE assignment and cross-region attention. The main BRIDGE setting uses CFG$=2$, while MagicBrush uses CFG$=4$.

\begin{figure}[H]
  \centering
  \includegraphics[width=0.98\linewidth]{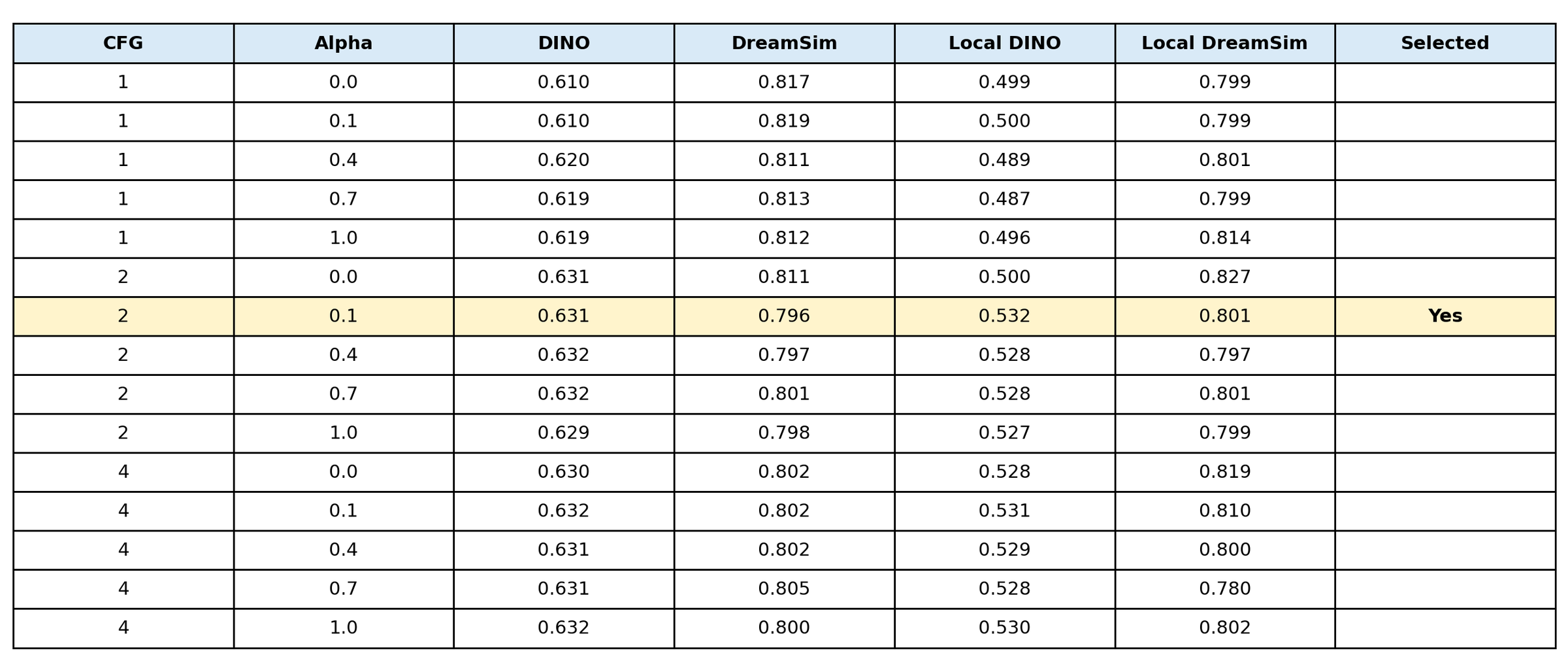}
  \caption{\textbf{Grid-search summary for the main setting.} We compare 15 combinations of CFG and latent-blending strength $\alpha$. The highlighted row marks the selected configuration used in the main experiments.}
  \label{fig:grid_search_main_table}
\end{figure}

\begin{figure}[H]
  \centering
  \includegraphics[width=0.98\linewidth]{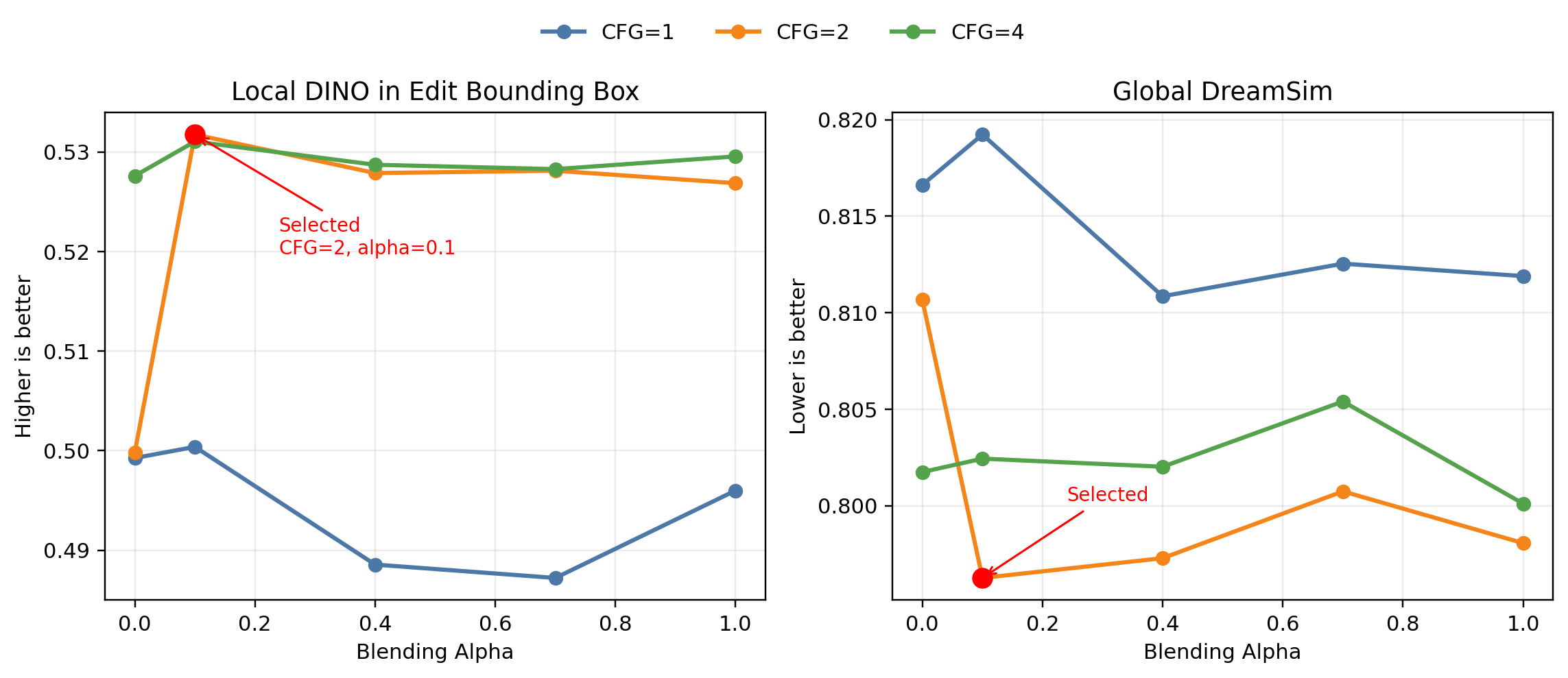}
  \caption{\textbf{Selection rationale for CFG and blending alpha in the main setting.} The left panel plots local DINO within the edit bounding box, and the right panel plots global DreamSim. The selected setting, CFG$=2$ and $\alpha=0.1$, is highlighted in red because it provides the strongest joint trade-off between local edit fidelity and global perceptual consistency among the tested combinations.}
  \label{fig:grid_search_main_plot}
\end{figure}

\paragraph{Main-Setting Grid Search for CFG and Blending Alpha.}
We additionally performed a grid search over CFG $\in \{1,2,4\}$ and blending $\alpha \in \{0.0,0.1,0.4,0.7,1.0\}$ for the main setting. The table image in Fig.~\ref{fig:grid_search_main_table} reports the aggregated metrics for all 15 configurations, and the line plot in Fig.~\ref{fig:grid_search_main_plot} highlights the two criteria we used most directly for selection: local DINO within the edit bounding box and global DreamSim. Among the tested settings, CFG$=2$ with $\alpha=0.1$ simultaneously achieves the best local DINO (0.5317) and the best global DreamSim (0.7963), which is why we use this pair for the main experiments. MagicBrush uses the separate inference setting stated above.

\paragraph{Historical MagicBrush Global-Description Results.}
Table~\ref{tab:magicbrush_global_results} reports the earlier protocol that pairs outputs with global descriptions. Main-paper Table~\ref{tab:magicbrush_results} uses current-turn local descriptions.

\begin{table}[H]
    \centering
    \caption{Historical zero-shot results on the MagicBrush test set using global descriptions. Final Turn and All Turn are reported separately, and Avg. Rank averages the ranks over the 10 displayed metrics.}
    \label{tab:magicbrush_global_results}
    \resizebox{\linewidth}{!}{
        \setlength{\tabcolsep}{2.1pt}
        \begin{tabular}{l|ccccc|ccccc|c}
            \toprule
            & \multicolumn{5}{c|}{\textbf{Final Turn}} & \multicolumn{5}{c|}{\textbf{All Turn}} & \textbf{Avg. Rank}$\downarrow$ \\
            \textbf{Method} & \textbf{L1}$\downarrow$ & \textbf{L2}$\downarrow$ & \textbf{CLIP-I}$\uparrow$ & \textbf{DINO}$\uparrow$ & \textbf{CLIP-T}$\uparrow$ & \textbf{L1}$\downarrow$ & \textbf{L2}$\downarrow$ & \textbf{CLIP-I}$\uparrow$ & \textbf{DINO}$\uparrow$ & \textbf{CLIP-T}$\uparrow$ & Avg. Rank \\
            \midrule
            ACE++ & 0.081 & \underline{\underline{0.032}} & 0.891 & 0.817 & \underline{\underline{0.307}} & 0.047 & 0.017 & 0.936 & 0.894 & \underline{\underline{0.306}} & 3.7 \\
            FLUX.1-Fill & \textbf{0.075} & \textbf{0.025} & \underline{\underline{0.898}} & \underline{\underline{0.839}} & 0.297 & \underline{0.044} & \textbf{0.013} & \textbf{0.947} & \textbf{0.919} & 0.300 & \underline{2.1} \\
            Q-Control & \underline{\underline{0.078}} & \underline{0.030} & \underline{0.898} & \textbf{0.840} & \underline{0.322} & \underline{\underline{0.044}} & \underline{0.016} & \underline{\underline{0.943}} & \underline{0.912} & \underline{0.315} & \underline{\underline{2.2}} \\
            \textbf{BRIDGE} & \underline{0.076} & \underline{\underline{0.032}} & \textbf{0.907} & \underline{0.839} & \textbf{0.324} & \textbf{0.043} & \underline{\underline{0.017}} & \underline{0.946} & \underline{\underline{0.910}} & \textbf{0.315} & \textbf{2.0} \\
            \bottomrule
        \end{tabular}
    }
\end{table}

\paragraph{MagicBrush Current-Protocol Evaluation.}
Main-paper Table~\ref{tab:magicbrush_results} evaluates full output images paired with current-turn local descriptions. Table~\ref{tab:magicbrush_local_bbox} evaluates a common tight mask-defined crop for every method and target. The crop evaluation uses this shared local image domain and mask-based blending.

\begin{table}[H]
    \centering
    \caption{Additional MagicBrush evaluation on the common tight edit-region crop (all $\uparrow$). Main-paper Table~\ref{tab:magicbrush_results} evaluates full images; this table evaluates the shared local crop.}
    \label{tab:magicbrush_local_bbox}
    \resizebox{0.94\linewidth}{!}{
    \begin{tabular}{llcccc}
        \toprule
        \textbf{Split} & \textbf{Method} & \textbf{CLIP-I} & \textbf{CLIP-T Local} & \textbf{SigLIP2-I} & \textbf{SigLIP2-T Local} \\
        \midrule
        Final & ACE++ & 0.7978 & 0.2607 & 0.7559 & 0.1864 \\
        & FLUX.1-Fill & 0.8054 & 0.2580 & 0.7743 & 0.1397 \\
        & Q-Control & 0.8150 & 0.2684 & 0.7932 & 0.2935 \\
        & \textbf{BRIDGE} & \textbf{0.8215} & \textbf{0.2727} & \textbf{0.8011} & \textbf{0.3377} \\
        \midrule
        All & ACE++ & 0.8113 & 0.2590 & 0.7797 & 0.1713 \\
        & FLUX.1-Fill & 0.8228 & 0.2591 & 0.7999 & 0.1522 \\
        & Q-Control & 0.8317 & 0.2676 & 0.8127 & 0.2985 \\
        & \textbf{BRIDGE} & \textbf{0.8383} & \textbf{0.2715} & \textbf{0.8208} & \textbf{0.3240} \\
        \bottomrule
    \end{tabular}}
\end{table}

\paragraph{Additional MagicBrush Inference Variants.}
Table~\ref{tab:magicbrush_extra_variants} summarizes supplementary MagicBrush evaluations by model and inference setting. The first row is the Qwen-Image-Edit ablation baseline evaluated with CFG$=4$ and mask blending. The other two rows are BRIDGE variants comparing CFG$=2$ with mask blending and CFG$=4$ with bbox blending.

\begin{table}[H]
    \centering
    \caption{Supplementary MagicBrush model and inference settings.}
    \label{tab:magicbrush_extra_variants}
    \resizebox{\linewidth}{!}{
        \setlength{\tabcolsep}{3pt}
        \begin{tabular}{lcc|ccccc|ccccc}
            \toprule
            \textbf{Model / Setting} & \textbf{CFG} & \textbf{Blend} & \multicolumn{5}{c|}{\textbf{Final Turn}} & \multicolumn{5}{c}{\textbf{All Turn}} \\
            & & & \textbf{L1}$\downarrow$ & \textbf{L2}$\downarrow$ & \textbf{CLIP-I}$\uparrow$ & \textbf{DINO}$\uparrow$ & \textbf{CLIP-T}$\uparrow$ & \textbf{L1}$\downarrow$ & \textbf{L2}$\downarrow$ & \textbf{CLIP-I}$\uparrow$ & \textbf{DINO}$\uparrow$ & \textbf{CLIP-T}$\uparrow$ \\
            \midrule
            Ablation baseline (Qwen-Image-edit) & 4 & mask & 0.0808 & 0.0324 & \textbf{0.9020} & 0.8188 & 0.3230 & 0.0468 & 0.0178 & \textbf{0.9418} & 0.8953 & 0.3140 \\
            BRIDGE (ours) & 2 & mask & 0.0815 & 0.0329 & 0.9006 & 0.8177 & 0.3224 & 0.0472 & 0.0181 & 0.9412 & 0.8949 & 0.3136 \\
            BRIDGE (ours) & 4 & bbox & 0.0838 & 0.0355 & 0.8997 & 0.8174 & \textbf{0.3246} & 0.0484 & 0.0195 & 0.9385 & 0.8910 & \textbf{0.3155} \\
            \bottomrule
        \end{tabular}
    }
\end{table}

\section{Data Curation and BRIDGE-Bench Construction}
\label{sec:appendix_data_curation}

\paragraph{Raw Corpus and Processed Split.}
Our raw source corpus is derived from Pico-Banana-400K~\citep{Gan2025PicoBanana400K}, a collection of approximately 400,000 text-guided image-editing examples based on real-world OpenImages photographs. Its dual-model pipeline uses a multimodal model to generate edits and a reasoning model for quality auditing, spanning a taxonomy of 35 edit types. We select five categories that require structural modification: Category Replacement, Object Addition, Object Removal, Clothing Edit, and Accessory Modification. We create the paper's train/test split after the filtering and compositing steps described below.

\paragraph{Precision Segmentation and Filtering.}
Raw generative outputs typically lack precise ground-truth segmentation masks. To address this, we re-engineered the Segment Anything Model 3 (SAM3)~\citep{carion2025sam} into an efficient parallel generate-and-filter pipeline. Driven by fine-grained sub-prompts generated by a Vision-Language Model (Qwen3-VL-32B~\citep{bai2025qwen3}), SAM3 generates candidate masks. The VLM then performs a ``Zoomed-in Analysis'' on cropped mask regions to assign semantic confidence scores, retaining only highly aligned mask proposals with a strict confidence threshold of $s \ge 0.95$.

\paragraph{Dual Audit and Background Consistency Verification.}
To reduce background drift and improve background consistency, we implement a dual audit:
\begin{itemize}
    \item \textbf{Individual Mask Filtering (Object Change):} We compute the DreamSim~\citep{fu2023dreamsim} distance between the source and target cropped regions. We enforce a minimum threshold of $d_{obj} > 0.25$. Regions failing to meet this threshold represent negligible perceptual changes and are discarded as false positives.
    \item \textbf{Global Background Audit:} We compute the union of all valid masks to form a global foreground mask. The background DreamSim distance (calculated on the inverted mask region) must be below a cut-off threshold ($d_{bg} < 0.6$), which discards samples with large global shifts outside the intended edit region.
\end{itemize}
Finally, for all retained samples, we apply a ``Forced Compositing'' strategy using guided filters~\citep{he2013guided} and morphological operations (e.g., erosion for removal, dilation for addition) paired with Gaussian alpha blending. This process weakens explicit shape boundaries, better simulates messy human user scribbles, and makes the retained background nearly identical at the pixel level ($L_1 \approx 0$).

\paragraph{Curating the BRIDGE-Bench Evaluation Set.}
We construct BRIDGE-Bench from the cleaned pool using a stricter background
threshold (e.g., DreamSim $D_{bg}<0.4$) and task-class balancing; the remaining
pairs form the training split. We audit the evaluation pairs using seam quality
along the expanded mask boundary and fine-grained background perceptual
distance.

\begin{figure}[H]
  \centering
  \includegraphics[width=0.95\linewidth]{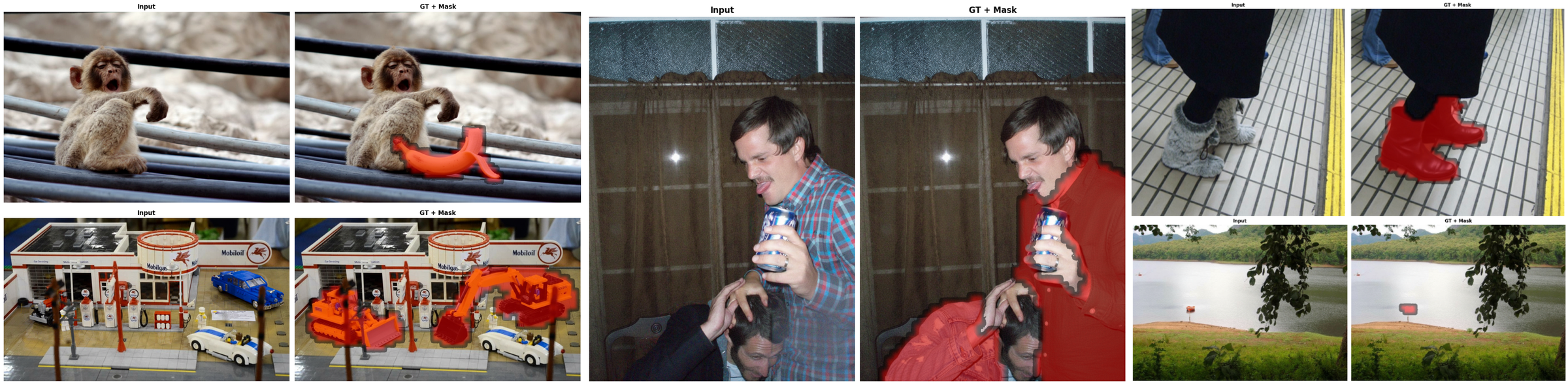}
  \caption{\textbf{Selected processed examples from our internal dataset.} These examples illustrate the instruction-guided local edits retained after semantic filtering, background auditing, and forced background compositing.}
  \label{fig:appendix_dataset_examples}
\end{figure}

Figure~\ref{fig:appendix_dataset_examples} shows selected processed examples from our internal dataset after the full filtering and compositing pipeline.

\paragraph{Matched Common-ID Evaluation.}
Table~\ref{tab:bridge_bench_common_ids} reports a matched evaluation using the
sample identifiers available for all three methods. Each metric is aggregated
over the same identifier set within its evaluator. The numerical differences
from main-paper Table~\ref{tab:combined_results} arise from this common-ID
restriction.

\begin{table}[H]
    \centering
    \caption{Matched common-ID evaluation on BRIDGE-Bench. The metric order matches main-paper Table~\ref{tab:combined_results}.}
    \label{tab:bridge_bench_common_ids}
    \resizebox{\linewidth}{!}{
    \begin{tabular}{lcccccccc}
        \toprule
        & \multicolumn{4}{c}{\textbf{Local Precision}} & \multicolumn{4}{c}{\textbf{Global Quality}} \\
        \cmidrule(lr){2-5}\cmidrule(lr){6-9}
        \textbf{Method} & \textbf{Sig-T}$\uparrow$ & \textbf{DINO}$\uparrow$ & \textbf{DSim}$\downarrow$ & \textbf{Sig-I}$\uparrow$ & \textbf{DINO}$\uparrow$ & \textbf{DSim}$\downarrow$ & \textbf{Sig-I}$\uparrow$ & \textbf{Sig-T}$\uparrow$ \\
        \midrule
        ACE++ & 0.392 & 0.852 & 0.303 & 0.856 & 0.982 & 0.082 & 0.954 & 0.705 \\
        FLUX.1-Fill & 0.262 & 0.807 & 0.321 & 0.835 & 0.978 & 0.076 & 0.947 & 0.631 \\
        \textbf{BRIDGE} & \textbf{0.507} & \textbf{0.903} & \textbf{0.173} & \textbf{0.908} & \textbf{0.988} & \textbf{0.053} & \textbf{0.965} & \textbf{0.778} \\
        \bottomrule
    \end{tabular}}
\end{table}

For naming clarity, the reported SigLIP2-T values are model-native sigmoid scores,
\begin{equation}
\label{eq:siglip_native}
\operatorname{sigmoid}\!\left(\cos(\widehat{e}_{I},\widehat{e}_{T})\,\exp(s)+b\right),
\end{equation}
with the model's learned log-scale parameter $s$ and bias $b$; they are not raw embedding cosines. We use the abbreviation ``SigLIP2-T'' for this score throughout the paper.

\section{Additional ICE-Bench Results}
\label{sec:appendix_icebench}

\paragraph{Data Preprocessing and Prompt Standardization.}
We evaluate 1,171 ICE-Bench samples across five local-editing tasks with the following modified-input and standardized-prompt protocol:
\begin{itemize}
    \item \textbf{Prompt template:} Each instruction begins with ``Picture 1 is the image to modify.'', and \texttt{<SOURCE>} is replaced with ``Picture 1''.
    \item \textbf{Region terminology:} Inpainting and Removal instructions replace ``mask'' references with ``black area''; Addition and Text Render instructions remove those references.
    \item \textbf{Removal inputs:} For \textit{Local Subject Removal} and \textit{Local Text Removal}, pixels inside the source mask are set to black before inference, matching the removal-specialized input format.
\end{itemize}

\paragraph{Score Definition.}
For ICE-Bench local editing tasks (Tasks 17--22)~\citep{pan2025icebench}, we map raw metrics into four normalized dimensions, where $\text{musiq\_koniq}$ follows MUSIQ~\citep{ke2021musiq}:
\begin{equation}
    S_{\text{AES}} = \frac{\text{aes\_v2.5}}{10}, \quad
    S_{\text{IMG}} = \frac{\text{musiq\_koniq}}{100},
\end{equation}
\begin{equation}
    S_{\text{PF}} = \frac{2 \cdot \text{CLIP-cap} + \text{VLLM-QA}}{2}, \quad
    S_{\text{SRC}} = \frac{\text{CLIP-src} + (1 - \text{L1-src})}{2}.
\end{equation}
The final score for each task is computed as:
\begin{equation}
    \text{TaskScore} = 0.3 S_{\text{AES}} + 0.3 S_{\text{IMG}} + 0.3 S_{\text{PF}} + 0.1 S_{\text{SRC}}.
\end{equation}
We report results on five tasks (Tasks 17, 19--22); Task 18 (Outpainting) is excluded in our evaluation.
The reported overall score is the equal-weight macro average over these tasks rather than a pooled sample average.

\paragraph{Results Analysis.}
As shown in Table~\ref{tab:icebench_local_dim_avg}, our method achieves the highest average score in \textbf{Aesthetic Quality (AES)} and competitive performance in \textbf{Source Consistency (SRC)}.
A high AES score (e.g., in Task 17 and Task 19) indicates that our generated results possess superior visual appeal and composition, which is critical for user preference in real-world editing scenarios.
High SRC scores (e.g., 0.942 average) demonstrate that BRIDGE effectively preserves the non-edited background regions, a key requirement for local editing tasks.
In Task 17 (Inpainting) and Task 19 (Local Subject Addition), our method obtains strong aesthetic and source consistency metrics (see Tables~\ref{tab:icebench_t17_metrics} and \ref{tab:icebench_t19_metrics}), suggesting that BridgePath geometric guidance helps balance local editing and background preservation.

\begin{table}[H]
    \centering
    \caption{Averaged dimension scores on ICE-Bench local editing (Tasks 17, 19--22). Baseline metrics are taken from the original ICE-Bench paper~\citep{pan2025icebench}.}
    \label{tab:icebench_local_dim_avg}
    \small
    \setlength{\tabcolsep}{3pt}
    \begin{tabular}{lcccc}
        \toprule
        \textbf{Method} & \textbf{AES} & \textbf{IMG} & \textbf{PF} & \textbf{SRC} \\
        \midrule
        ACE++~\citep{han2025aceplus} & 0.481 & \textbf{0.577} & 0.616 & 0.930 \\
        ACE~\citep{han2024ace} & 0.480 & 0.487 & \textbf{0.643} & 0.916 \\
        UltraEdit~\citep{zhao2024ultraedit} & 0.455 & 0.458 & 0.442 & \textbf{0.951} \\
        \textbf{BRIDGE} & \textbf{0.494} & 0.565 & 0.582 & 0.942 \\
        \bottomrule
    \end{tabular}
\end{table}

\paragraph{Per-task Raw Metrics.}
Tables~\ref{tab:icebench_t17_metrics}, \ref{tab:icebench_t19_metrics}, \ref{tab:icebench_t20_metrics}, \ref{tab:icebench_t21_metrics}, and~\ref{tab:icebench_t22_metrics} report the raw metrics (before normalization) for each evaluated task. Baseline rows are taken from ICE-Bench~\citep{pan2025icebench}.

\begin{table}[H]
    \centering
    \caption{ICE-Bench Task 17 (Inpainting) detailed metrics. Baseline metrics are taken from the original ICE-Bench paper~\citep{pan2025icebench}.}
    \label{tab:icebench_t17_metrics}
    \resizebox{\linewidth}{!}{
        \setlength{\tabcolsep}{2.5pt}
        \begin{tabular}{lcccccc}
            \toprule
            \textbf{Method} & \textbf{Aesthetic} & \textbf{Imaging} & \textbf{CLIP-cap} & \textbf{VLLM-QA} & \textbf{CLIP-src} & \textbf{L1-src} \\
            \midrule
            ACE++~\citep{han2025aceplus} & 5.064 & 61.661 & 0.272 & 0.910 & 0.776 & 0.016 \\
            ACE~\citep{han2024ace} & 4.878 & 51.793 & 0.269 & 0.833 & 0.785 & 0.024 \\
            UltraEdit~\citep{zhao2024ultraedit} & 3.817 & 46.284 & 0.250 & 0.180 & \textbf{0.952} & 0.019 \\
            \textbf{BRIDGE} & \textbf{5.214} & \textbf{61.685} & \textbf{0.278} & \textbf{0.953} & 0.789 & \textbf{0.013} \\
            \bottomrule
        \end{tabular}
    }
\end{table}

\begin{table}[H]
    \centering
    \caption{ICE-Bench Task 19 (Local Subject Addition) detailed metrics. Baseline metrics are taken from the original ICE-Bench paper~\citep{pan2025icebench}.}
    \label{tab:icebench_t19_metrics}
    \resizebox{\linewidth}{!}{
        \setlength{\tabcolsep}{2.5pt}
        \begin{tabular}{lcccccc}
            \toprule
            \textbf{Method} & \textbf{Aesthetic} & \textbf{Imaging} & \textbf{CLIP-cap} & \textbf{VLLM-QA} & \textbf{CLIP-src} & \textbf{L1-src} \\
            \midrule
            ACE++~\citep{han2025aceplus} & 5.014 & \textbf{62.083} & 0.268 & \textbf{0.785} & 0.894 & 0.018 \\
            ACE~\citep{han2024ace} & 4.965 & 51.704 & 0.272 & 0.555 & 0.897 & 0.029 \\
            UltraEdit~\citep{zhao2024ultraedit} & 4.881 & 47.855 & 0.275 & 0.555 & 0.909 & 0.021 \\
            \textbf{BRIDGE} & \textbf{5.146} & 61.239 & \textbf{0.278} & 0.426 & \textbf{0.937} & \textbf{0.013} \\
            \bottomrule
        \end{tabular}
    }
\end{table}

\begin{table}[H]
    \centering
    \caption{ICE-Bench Task 20 (Local Subject Removal) detailed metrics. Baseline metrics are taken from the original ICE-Bench paper~\citep{pan2025icebench}.}
    \label{tab:icebench_t20_metrics}
    \resizebox{\linewidth}{!}{
        \setlength{\tabcolsep}{2.5pt}
        \begin{tabular}{lcccccc}
            \toprule
            \textbf{Method} & \textbf{Aesthetic} & \textbf{Imaging} & \textbf{CLIP-cap} & \textbf{VLLM-QA} & \textbf{CLIP-src} & \textbf{L1-src} \\
            \midrule
            ACE++~\citep{han2025aceplus} & \textbf{5.061} & \textbf{61.614} & 0.229 & 0.312 & \textbf{0.901} & 0.017 \\
            ACE~\citep{han2024ace} & 4.996 & 47.011 & 0.258 & \textbf{0.757} & 0.852 & 0.024 \\
            UltraEdit~\citep{zhao2024ultraedit} & 4.858 & 48.748 & 0.226 & 0.287 & 0.888 & 0.018 \\
            \textbf{BRIDGE} & 5.009 & 57.203 & \textbf{0.262} & 0.748 & 0.861 & \textbf{0.013} \\
            \bottomrule
        \end{tabular}
    }
\end{table}

\begin{table}[H]
    \centering
    \caption{ICE-Bench Task 21 (Local Text Render) detailed metrics. Baseline metrics are taken from the original ICE-Bench paper~\citep{pan2025icebench}.}
    \label{tab:icebench_t21_metrics}
    \resizebox{\linewidth}{!}{
        \setlength{\tabcolsep}{2.5pt}
        \begin{tabular}{lcccccc}
            \toprule
            \textbf{Method} & \textbf{Aesthetic} & \textbf{Imaging} & \textbf{CLIP-cap} & \textbf{VLLM-QA} & \textbf{CLIP-src} & \textbf{L1-src} \\
            \midrule
            ACE++~\citep{han2025aceplus} & 4.231 & \textbf{43.276} & 0.277 & \textbf{0.834} & 0.899 & 0.012 \\
            ACE~\citep{han2024ace} & 4.275 & 43.159 & 0.276 & 0.791 & 0.860 & 0.016 \\
            UltraEdit~\citep{zhao2024ultraedit} & 4.506 & 38.887 & 0.277 & 0.098 & 0.946 & 0.014 \\
            \textbf{BRIDGE} & \textbf{4.527} & 43.224 & \textbf{0.279} & 0.307 & \textbf{0.957} & \textbf{0.006} \\
            \bottomrule
        \end{tabular}
    }
\end{table}

\begin{table}[H]
    \centering
    \caption{ICE-Bench Task 22 (Local Text Removal) detailed metrics. Baseline metrics are taken from the original ICE-Bench paper~\citep{pan2025icebench}.}
    \label{tab:icebench_t22_metrics}
    \resizebox{\linewidth}{!}{
        \setlength{\tabcolsep}{2.5pt}
        \begin{tabular}{lcccccc}
            \toprule
            \textbf{Method} & \textbf{Aesthetic} & \textbf{Imaging} & \textbf{CLIP-cap} & \textbf{VLLM-QA} & \textbf{CLIP-src} & \textbf{L1-src} \\
            \midrule
            ACE++~\citep{han2025aceplus} & 4.694 & \textbf{59.636} & 0.260 & 0.704 & 0.905 & 0.017 \\
            ACE~\citep{han2024ace} & \textbf{4.896} & 49.766 & \textbf{0.273} & \textbf{0.801} & 0.888 & 0.033 \\
            UltraEdit~\citep{zhao2024ultraedit} & 4.665 & 47.294 & 0.264 & 0.714 & 0.910 & 0.023 \\
            \textbf{BRIDGE} & 4.820 & 58.985 & 0.271 & 0.653 & \textbf{0.937} & \textbf{0.012} \\
            \bottomrule
        \end{tabular}
    }
\end{table}

\paragraph{Direct Evaluation Reports for Q-Control, ACE++, and BRIDGE.}
The direct Q-Control, ACE++, and BRIDGE evaluations use the modified-input and
standardized-prompt protocol defined above. Table~\ref{tab:icebench_direct_final_scores}
reports per-task final scores and the five-task average, and
Table~\ref{tab:icebench_direct_avg_dims} reports the averaged mapped dimensions.
Tables~\ref{tab:icebench_direct_t17}, \ref{tab:icebench_direct_t19},
\ref{tab:icebench_direct_t20}, \ref{tab:icebench_direct_t21},
and~\ref{tab:icebench_direct_t22} provide the corresponding per-task metrics.

\begin{table}[H]
    \centering
    \caption{ICE-Bench final scores for the three evaluated methods under the standardized protocol. Task 18 (Outpainting) is not included in this evaluation.}
    \label{tab:icebench_direct_final_scores}
    \resizebox{\linewidth}{!}{
        \setlength{\tabcolsep}{3pt}
        \begin{tabular}{lcccccc}
            \toprule
            \textbf{Run} & \textbf{T17} & \textbf{T19} & \textbf{T20} & \textbf{T21} & \textbf{T22} & \textbf{Avg.} \\
            \midrule
            Q-Control & 0.601768 & 0.613342 & 0.591195 & 0.514650 & 0.590880 & 0.582367 \\
            ACE++ & 0.642074 & 0.607212 & 0.526383 & 0.557288 & 0.574833 & 0.581558 \\
            \textbf{BRIDGE} & \textbf{0.656471} & 0.581629 & \textbf{0.605180} & 0.492648 & \textbf{0.597094} & \textbf{0.586604} \\
            \bottomrule
        \end{tabular}
    }
\end{table}

\begin{table}[H]
    \centering
    \caption{Averaged mapped dimension scores for the three evaluated methods on ICE-Bench Tasks 17 and 19--22.}
    \label{tab:icebench_direct_avg_dims}
    \resizebox{0.82\linewidth}{!}{
        \setlength{\tabcolsep}{4pt}
        \begin{tabular}{lccccc}
            \toprule
            \textbf{Run} & \textbf{AES} & \textbf{IMG} & \textbf{PF} & \textbf{SRC} & \textbf{Avg. Score} \\
            \midrule
            Q-Control & 0.472404 & 0.568606 & 0.587753 & 0.937379 & 0.582367 \\
            ACE++ & 0.479863 & \textbf{0.575003} & 0.575187 & 0.925421 & 0.581558 \\
            \textbf{BRIDGE} & \textbf{0.494323} & 0.564674 & 0.582213 & \textbf{0.942415} & \textbf{0.586604} \\
            \bottomrule
        \end{tabular}
    }
\end{table}

\begin{table}[H]
    \centering
    \caption{Per-method metrics for ICE-Bench Task 17 (Inpainting).}
    \label{tab:icebench_direct_t17}
    \resizebox{\linewidth}{!}{
        \setlength{\tabcolsep}{2.5pt}
        \begin{tabular}{lcccccc}
            \toprule
            \textbf{Run} & \textbf{Aesthetic} & \textbf{Imaging} & \textbf{CLIP-cap} & \textbf{VLLM-QA} & \textbf{CLIP-src} & \textbf{L1-src} \\
            \midrule
            Q-Control & 4.353530 & 61.674982 & 0.267331 & 0.765586 & 0.838662 & 0.016663 \\
            ACE++ & 5.069124 & 61.198315 & 0.276025 & 0.897756 & 0.793418 & 0.014730 \\
            \textbf{BRIDGE} & \textbf{5.214464} & \textbf{61.685315} & \textbf{0.277598} & \textbf{0.952618} & 0.789001 & \textbf{0.012815} \\
            \bottomrule
        \end{tabular}
    }
\end{table}

\begin{table}[H]
    \centering
    \caption{Per-method metrics for ICE-Bench Task 19 (Local Subject Addition).}
    \label{tab:icebench_direct_t19}
    \resizebox{\linewidth}{!}{
        \setlength{\tabcolsep}{2.5pt}
        \begin{tabular}{lcccccc}
            \toprule
            \textbf{Run} & \textbf{Aesthetic} & \textbf{Imaging} & \textbf{CLIP-cap} & \textbf{VLLM-QA} & \textbf{CLIP-src} & \textbf{L1-src} \\
            \midrule
            Q-Control & 5.114608 & \textbf{62.391083} & 0.276779 & 0.641148 & 0.887129 & 0.016649 \\
            ACE++ & 5.075284 & 61.490830 & 0.272050 & 0.631579 & 0.897898 & 0.015308 \\
            \textbf{BRIDGE} & \textbf{5.145709} & 61.239281 & \textbf{0.278256} & 0.425837 & \textbf{0.936747} & \textbf{0.013002} \\
            \bottomrule
        \end{tabular}
    }
\end{table}

\begin{table}[H]
    \centering
    \caption{Per-method metrics for ICE-Bench Task 20 (Local Subject Removal).}
    \label{tab:icebench_direct_t20}
    \resizebox{\linewidth}{!}{
        \setlength{\tabcolsep}{2.5pt}
        \begin{tabular}{lcccccc}
            \toprule
            \textbf{Run} & \textbf{Aesthetic} & \textbf{Imaging} & \textbf{CLIP-cap} & \textbf{VLLM-QA} & \textbf{CLIP-src} & \textbf{L1-src} \\
            \midrule
            Q-Control & \textbf{5.021658} & 58.158275 & 0.260421 & 0.633663 & \textbf{0.873895} & 0.015996 \\
            ACE++ & 4.940362 & \textbf{61.914121} & 0.229052 & 0.207921 & 0.867261 & 0.016733 \\
            \textbf{BRIDGE} & 5.009282 & 57.203025 & \textbf{0.262418} & \textbf{0.747525} & 0.861470 & \textbf{0.012711} \\
            \bottomrule
        \end{tabular}
    }
\end{table}

\begin{table}[H]
    \centering
    \caption{Per-method metrics for ICE-Bench Task 21 (Local Text Render).}
    \label{tab:icebench_direct_t21}
    \resizebox{\linewidth}{!}{
        \setlength{\tabcolsep}{2.5pt}
        \begin{tabular}{lcccccc}
            \toprule
            \textbf{Run} & \textbf{Aesthetic} & \textbf{Imaging} & \textbf{CLIP-cap} & \textbf{VLLM-QA} & \textbf{CLIP-src} & \textbf{L1-src} \\
            \midrule
            Q-Control & 4.330330 & \textbf{44.016858} & \textbf{0.281760} & 0.490798 & 0.901521 & 0.010693 \\
            ACE++ & 4.266104 & 43.993627 & 0.275505 & \textbf{0.803681} & 0.893160 & 0.010754 \\
            \textbf{BRIDGE} & \textbf{4.526936} & 43.223927 & 0.278827 & 0.306748 & \textbf{0.956576} & \textbf{0.006425} \\
            \bottomrule
        \end{tabular}
    }
\end{table}

\begin{table}[H]
    \centering
    \caption{Per-method metrics for ICE-Bench Task 22 (Local Text Removal) under the modified-input protocol.}
    \label{tab:icebench_direct_t22}
    \resizebox{\linewidth}{!}{
        \setlength{\tabcolsep}{2.5pt}
        \begin{tabular}{lcccccc}
            \toprule
            \textbf{Run} & \textbf{Aesthetic} & \textbf{Imaging} & \textbf{CLIP-cap} & \textbf{VLLM-QA} & \textbf{CLIP-src} & \textbf{L1-src} \\
            \midrule
            Q-Control & 4.800064 & 58.061950 & 0.270548 & 0.632653 & \textbf{0.949094} & 0.016509 \\
            ACE++ & 4.642299 & 58.904480 & 0.262020 & 0.581633 & 0.875389 & 0.015392 \\
            \textbf{BRIDGE} & \textbf{4.819754} & \textbf{58.985408} & \textbf{0.271069} & \textbf{0.653061} & 0.937104 & \textbf{0.011801} \\
            \bottomrule
        \end{tabular}
    }
\end{table}

\clearpage

\section{Additional Qualitative Results}
\label{sec:supp_qualitative}
We provide additional qualitative results on BRIDGE-Bench across several editing types.
In particular, we highlight cases where the model generates \textbf{sharp, detailed structures} that are less constrained by the initial mask shape. When the mask does not match the object's natural geometry, inpainting baselines often produce blurry or box-like artifacts. BRIDGE can synthesize more organic shapes (e.g., hair strands, animal limbs, complex textures) while maintaining background consistency in these examples.

\subsection{Additional Results: Remove}
\begin{figure}[H]
\centering
\setlength{\tabcolsep}{1pt}
\renewcommand{\arraystretch}{0.5}
\begin{tabular}{c c c}
\includegraphics[width=0.30\linewidth]{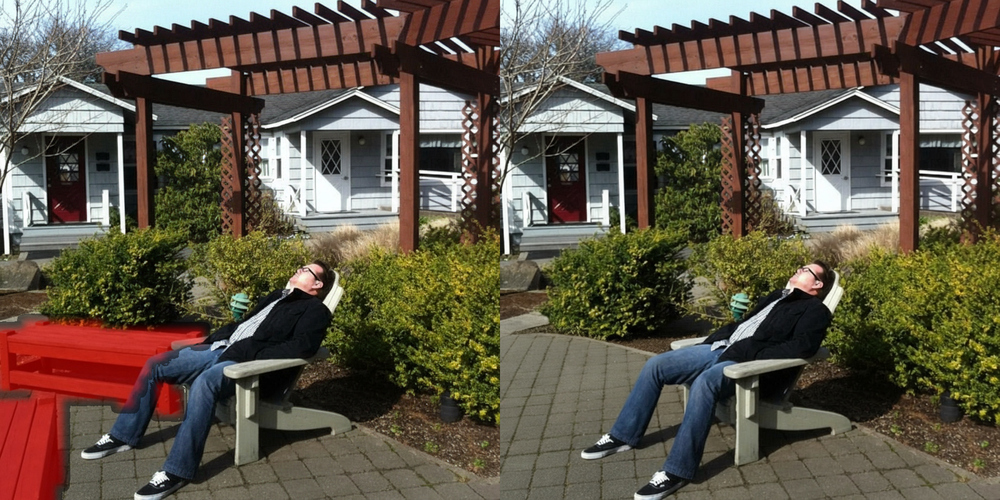} & \includegraphics[width=0.30\linewidth]{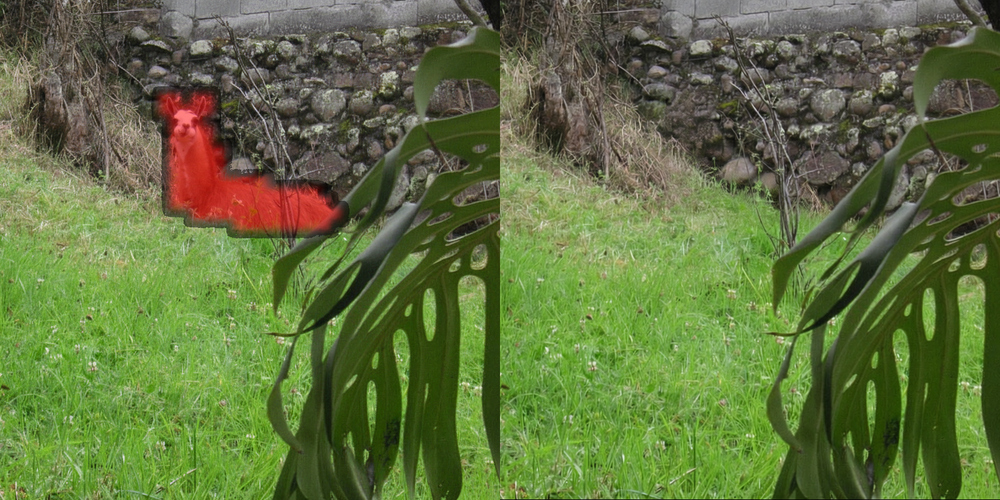} & \includegraphics[width=0.30\linewidth]{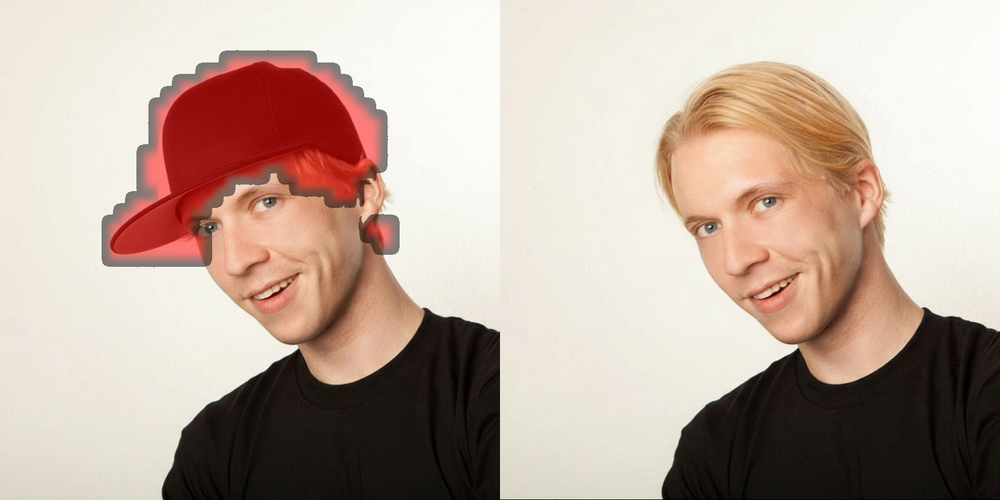}\\
\parbox{0.30\linewidth}{\centering\scriptsize \textit{extend the paved ground and dirt seamlessly where the wooden bench was....}} & \parbox{0.30\linewidth}{\centering\scriptsize \textit{Remove the brown llama and blend the space with green grass and...}} & \parbox{0.30\linewidth}{\centering\scriptsize \textit{Add blonde hair to cover the area where the cap was, matching...}}\\
\includegraphics[width=0.30\linewidth]{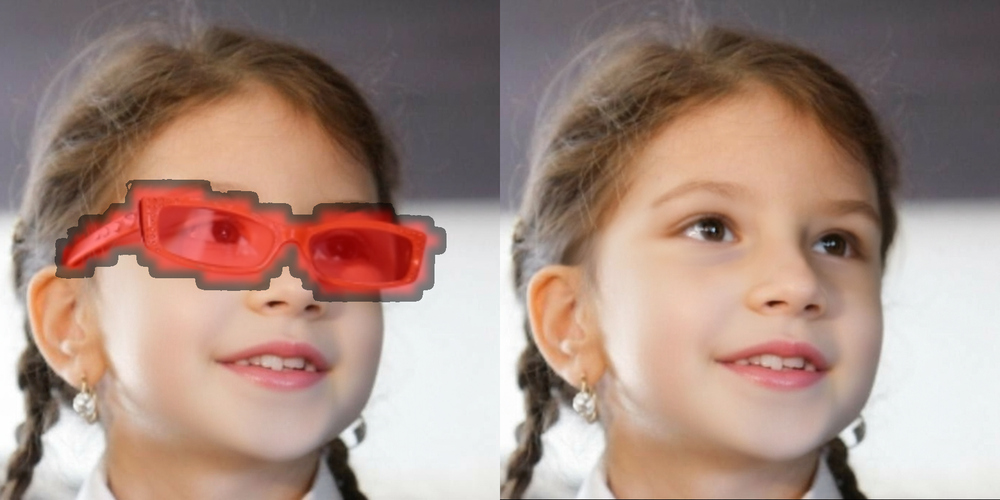}\\
\parbox{0.30\linewidth}{\centering\scriptsize \textit{Remove the glasses, smooth the skin, extend the hair, and maintain the...}}\\
\end{tabular}
\caption{Additional results for \textbf{Remove}.}
\end{figure}
\clearpage
\subsection{Additional Results: Add}
\begin{figure}[H]
\centering
\setlength{\tabcolsep}{1pt}
\renewcommand{\arraystretch}{0.5}
\begin{tabular}{c c c}
\includegraphics[width=0.30\linewidth]{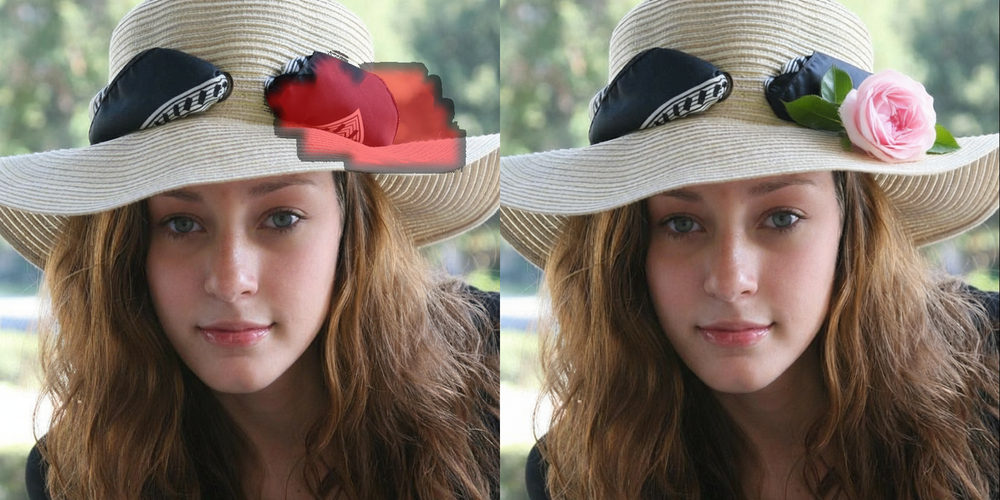} & \includegraphics[width=0.30\linewidth]{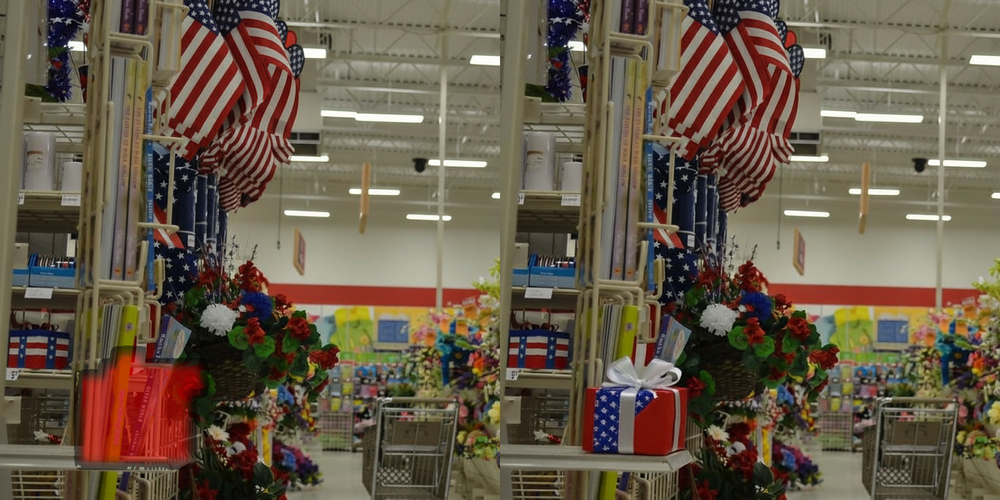} & \includegraphics[width=0.30\linewidth]{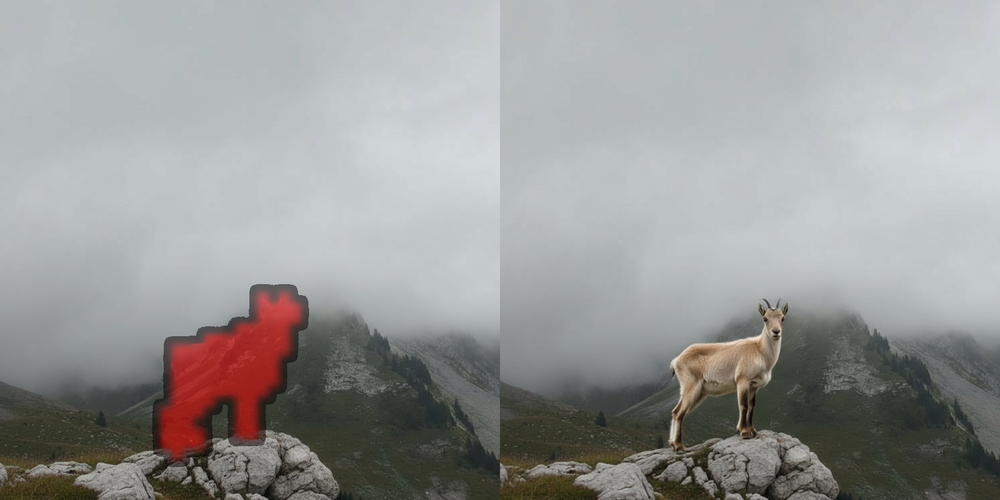}\\
\parbox{0.30\linewidth}{\centering\scriptsize \textit{Add a light pink rosebud with soft green leaves to the hat's...}} & \parbox{0.30\linewidth}{\centering\scriptsize \textit{Seamlessly integrate a small, festive red, white, and blue gift box, adorned...}} & \parbox{0.30\linewidth}{\centering\scriptsize \textit{Add a light brown, white-furred mountain goat standing alert on a rocky...}}\\
\includegraphics[width=0.30\linewidth]{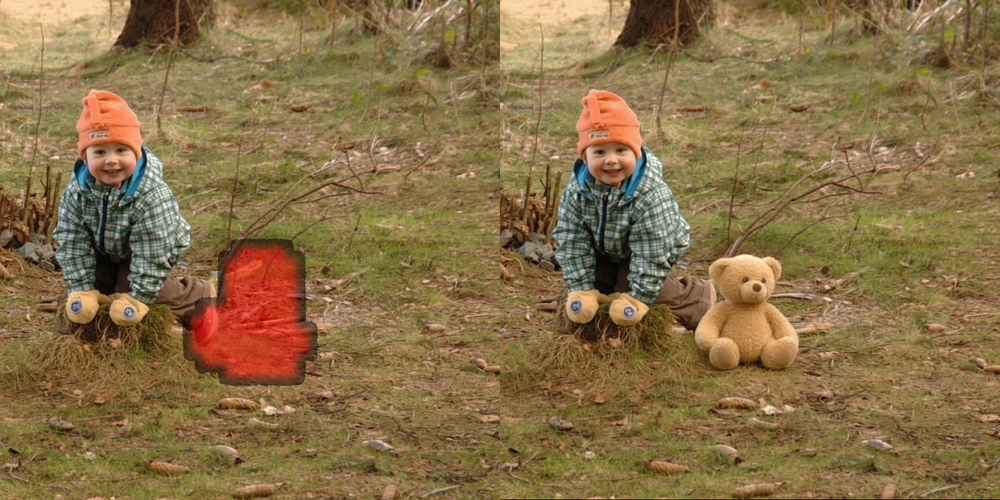} & \includegraphics[width=0.30\linewidth]{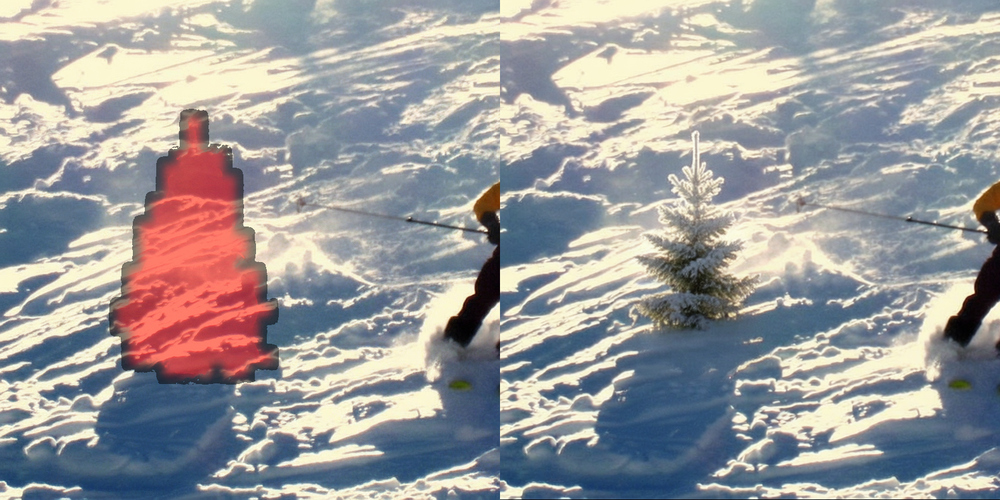} & \includegraphics[width=0.30\linewidth]{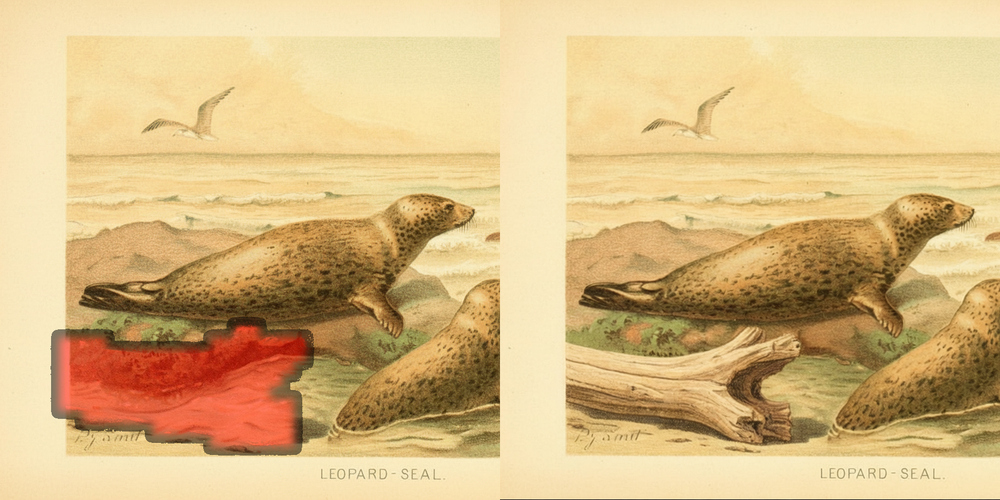}\\
\parbox{0.30\linewidth}{\centering\scriptsize \textit{Add a light brown teddy bear sitting upright on the moss, observing...}} & \parbox{0.30\linewidth}{\centering\scriptsize \textit{Add a small, frosted pine tree to the left mid-ground, matching the sunlight and snow tones.}} & \parbox{0.30\linewidth}{\centering\scriptsize \textit{Add a weathered, light brown driftwood piece in the lower left, partly...}}\\
\includegraphics[width=0.30\linewidth]{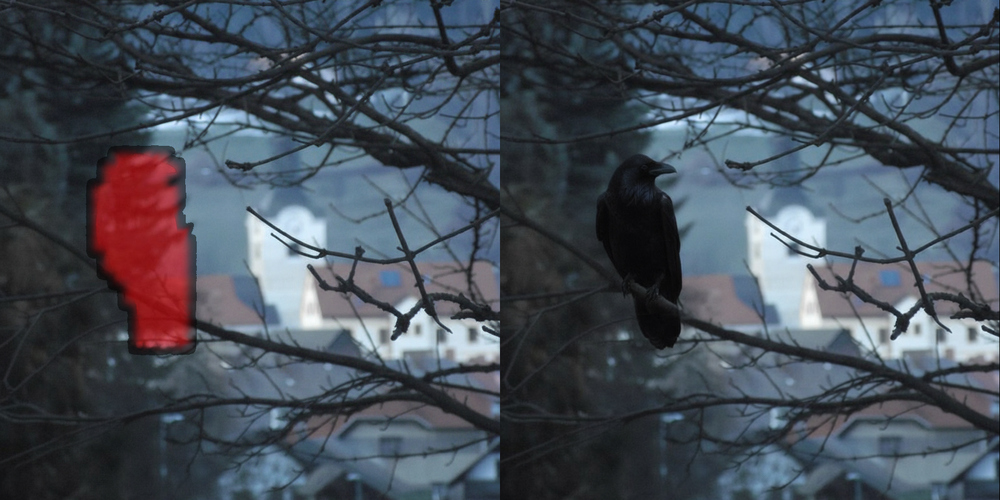} & \includegraphics[width=0.30\linewidth]{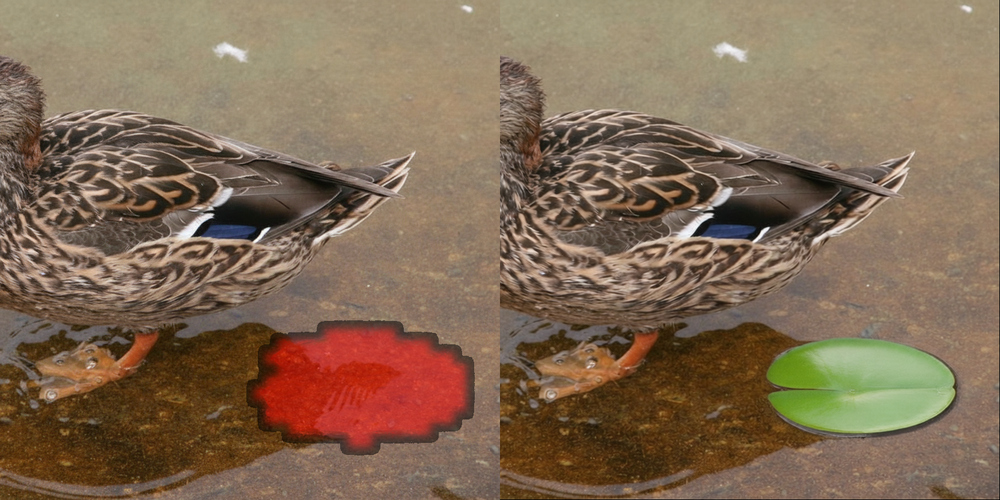}\\
\parbox{0.30\linewidth}{\centering\scriptsize \textit{Add a dark raven perched on a mid-left branch, facing the village....}} & \parbox{0.30\linewidth}{\centering\scriptsize \textit{Add a small, realistic green lily pad to the water next to...}}\\
\end{tabular}
\caption{Additional results for \textbf{Add}.}
\end{figure}
\clearpage
\subsection{Additional Results: Replace}
\begin{figure}[H]
\centering
\setlength{\tabcolsep}{1pt}
\renewcommand{\arraystretch}{0.5}
\begin{tabular}{c c c}
\includegraphics[width=0.30\linewidth]{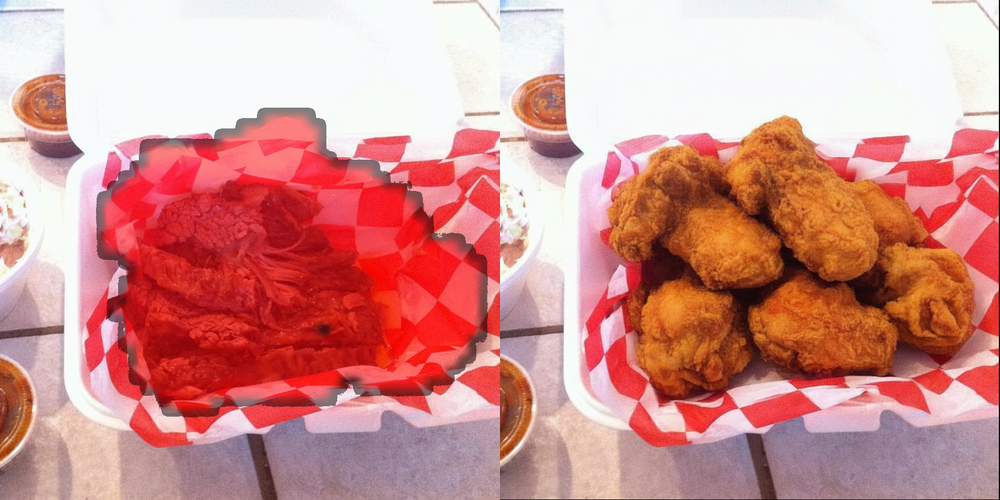} & \includegraphics[width=0.30\linewidth]{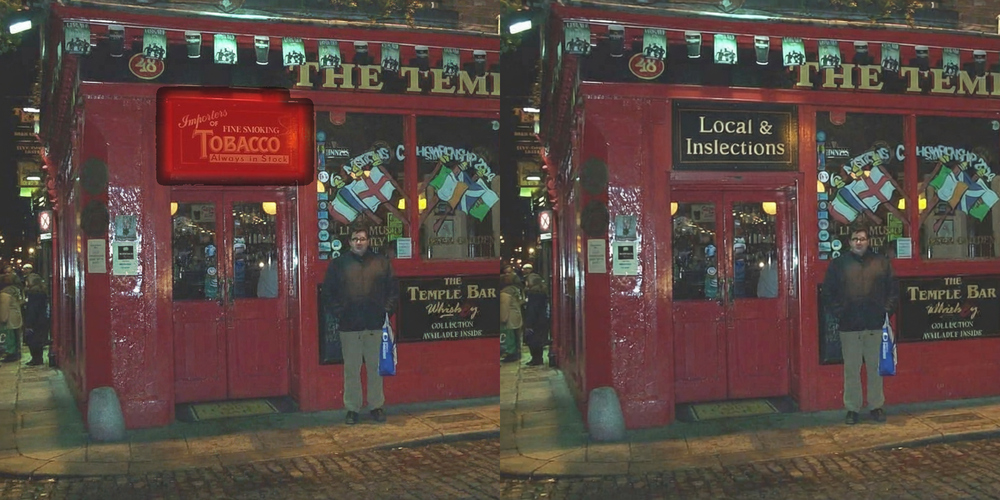} & \includegraphics[width=0.30\linewidth]{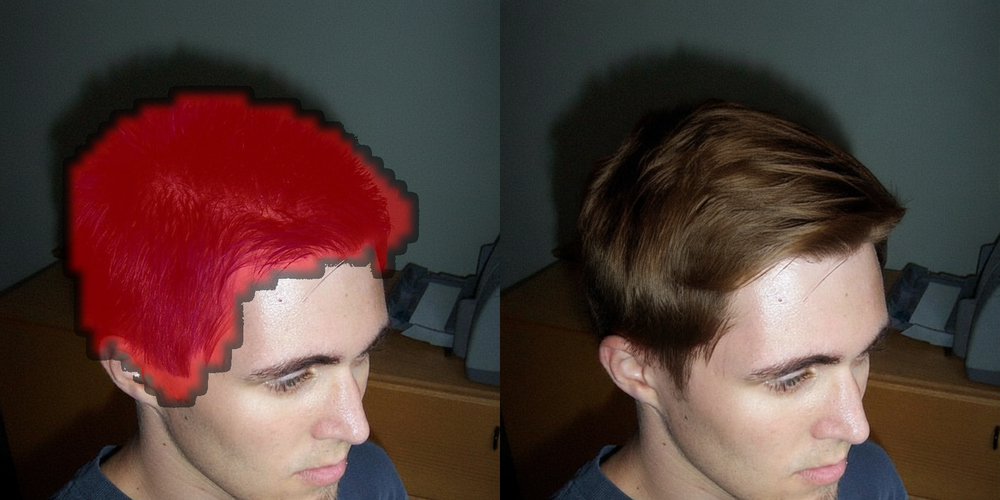}\\
\parbox{0.30\linewidth}{\centering\scriptsize \textit{Add crispy fried chicken to the container, matching the texture and lighting....}} & \parbox{0.30\linewidth}{\centering\scriptsize \textit{Sign: ``Craft Beer: Local \& Imported Selections'' with dark bg, cream/gold text,...}} & \parbox{0.30\linewidth}{\centering\scriptsize \textit{Hair color the vibrant purple to natural medium brown, matching the lighting...}}\\
\includegraphics[width=0.30\linewidth]{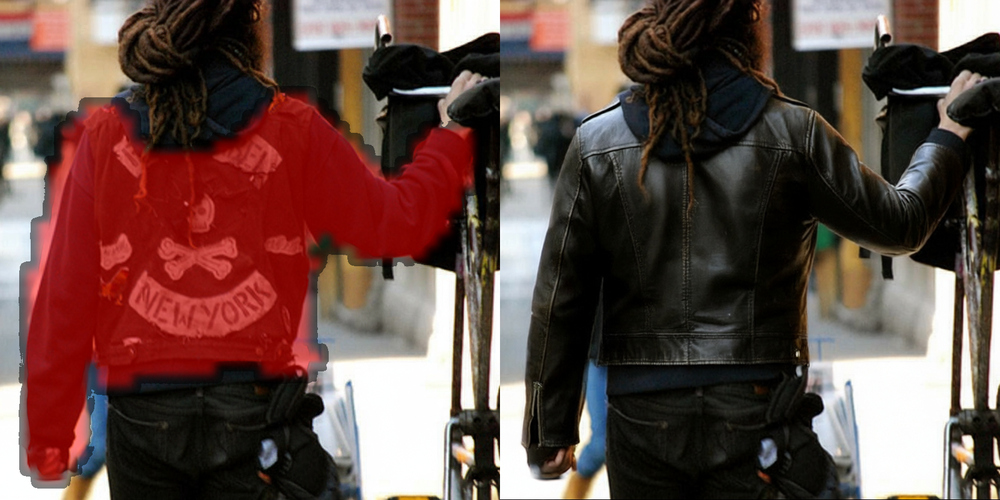} & \includegraphics[width=0.30\linewidth]{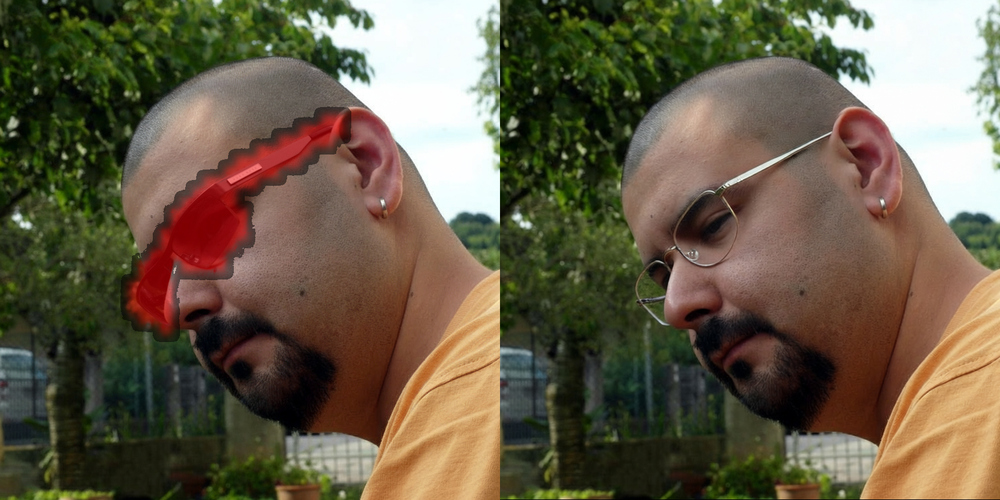} & \includegraphics[width=0.30\linewidth]{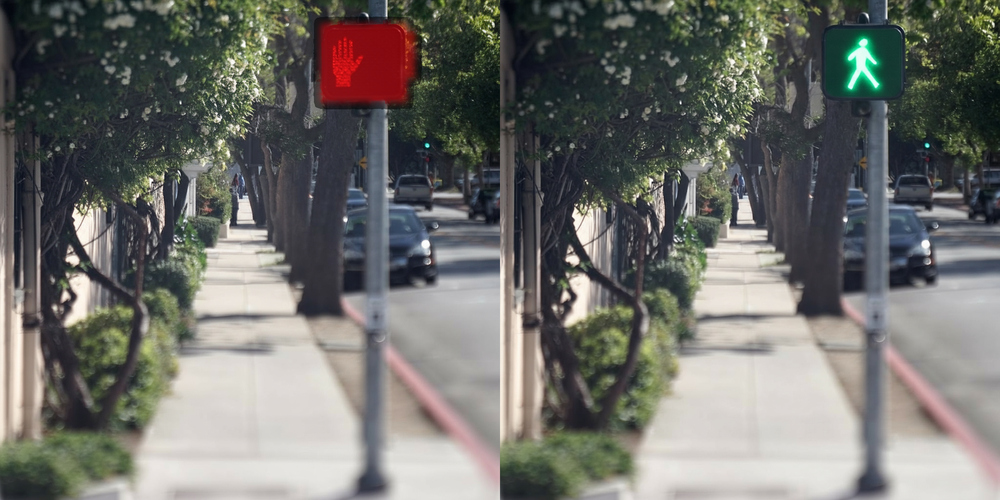}\\
\parbox{0.30\linewidth}{\centering\scriptsize \textit{Add a sleek, worn black leather jacket, reflecting street light, draping over...}} & \parbox{0.30\linewidth}{\centering\scriptsize \textit{Change the man's dark sunglasses to clear, thin metallic silver glasses that...}} & \parbox{0.30\linewidth}{\centering\scriptsize \textit{Change the red 'Don't Walk' signal to a vibrant green 'Walk' signal,...}}\\
\includegraphics[width=0.30\linewidth]{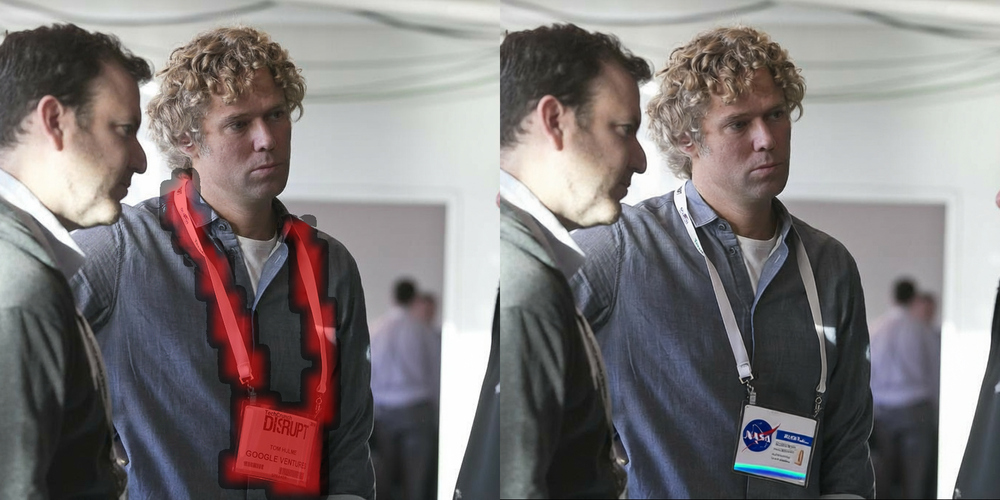} & \includegraphics[width=0.30\linewidth]{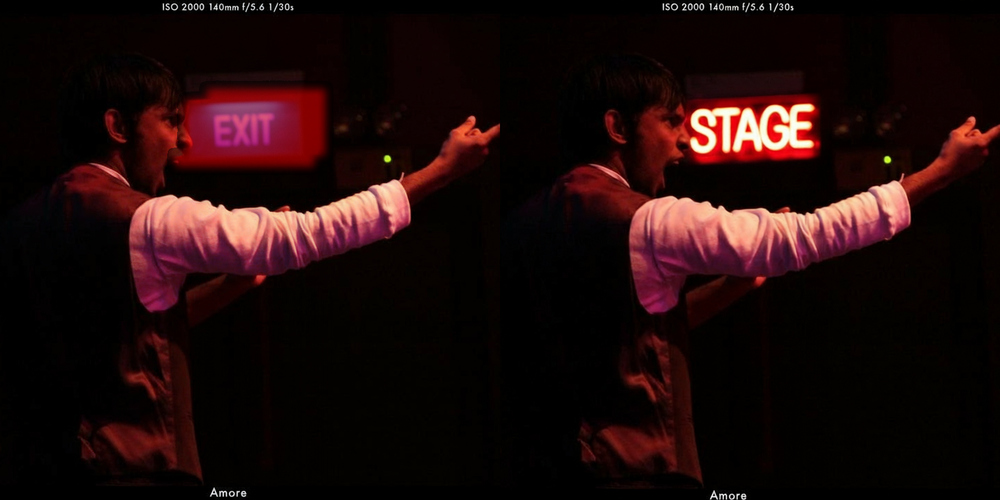} & \includegraphics[width=0.30\linewidth]{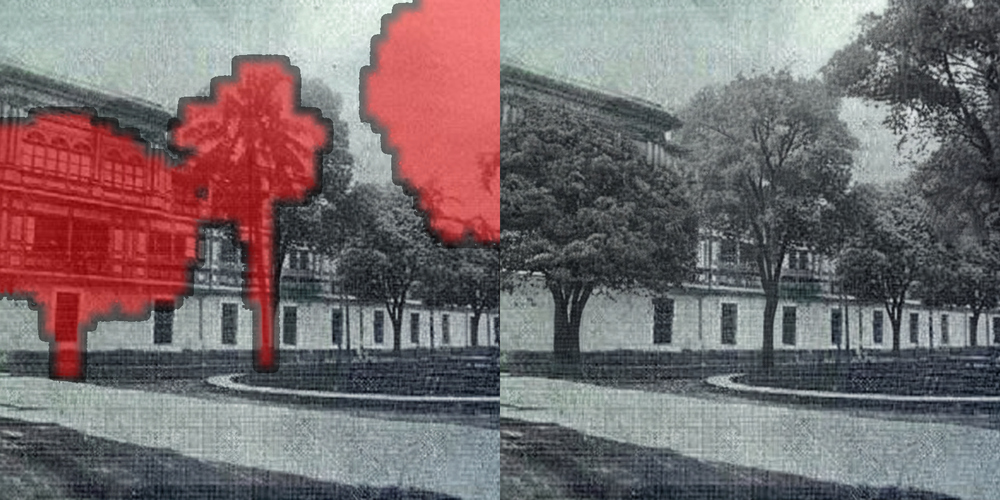}\\
\parbox{0.30\linewidth}{\centering\scriptsize \textit{Change the man's TechCrunch badge to a realistic NASA badge with blue...}} & \parbox{0.30\linewidth}{\centering\scriptsize \textit{Replace the blue glowing 'EXIT' sign in the background with a vibrant,...}} & \parbox{0.30\linewidth}{\centering\scriptsize \textit{Change all palm trees to mature deciduous trees with textured bark, maintaining...}}\\
\end{tabular}
\caption{Additional results for \textbf{Replace} (1/2).}
\end{figure}

\begin{figure}[H]
\centering
\setlength{\tabcolsep}{1pt}
\renewcommand{\arraystretch}{0.5}
\begin{tabular}{c c c}
\includegraphics[width=0.30\linewidth]{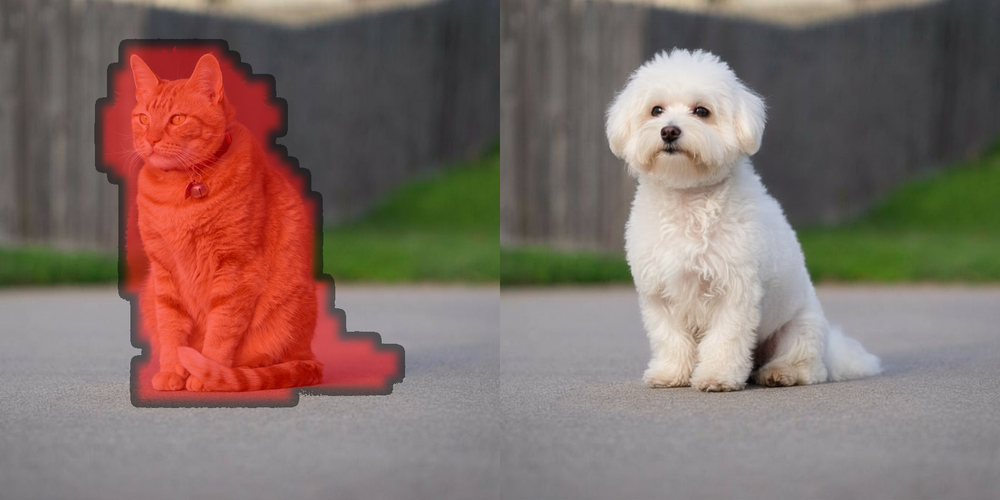} & \includegraphics[width=0.30\linewidth]{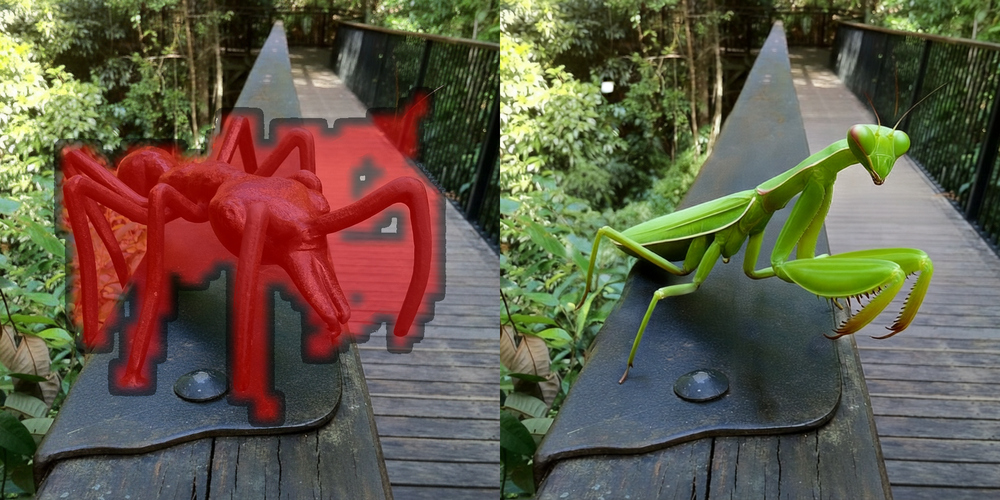} & \includegraphics[width=0.30\linewidth]{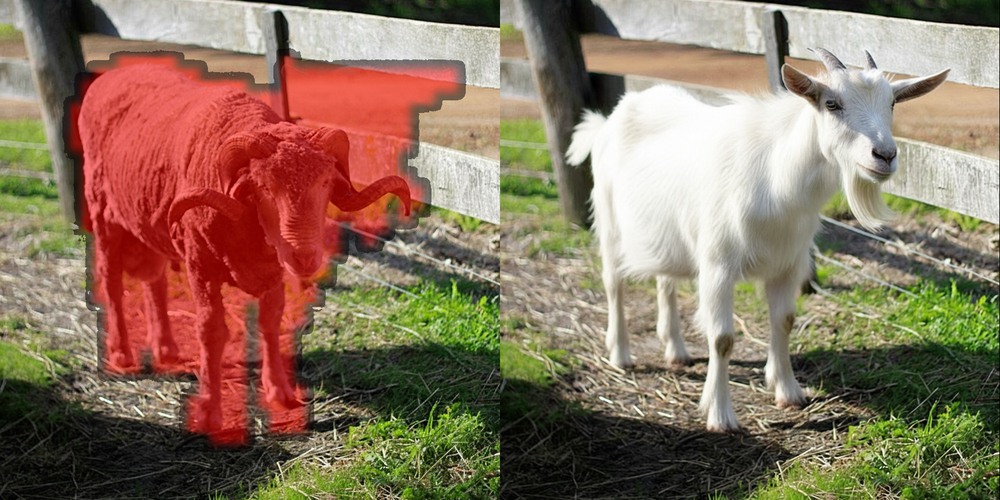}\\
\parbox{0.30\linewidth}{\centering\scriptsize \textit{Replace the orange tabby cat with a fluffy, white Bichon Frise dog,...}} & \parbox{0.30\linewidth}{\centering\scriptsize \textit{Add a large, realistic, vibrant green praying mantis perched on the railing,...}} & \parbox{0.30\linewidth}{\centering\scriptsize \textit{Change the ram to a white goat with shaggy fur, standing next...}}\\
\includegraphics[width=0.30\linewidth]{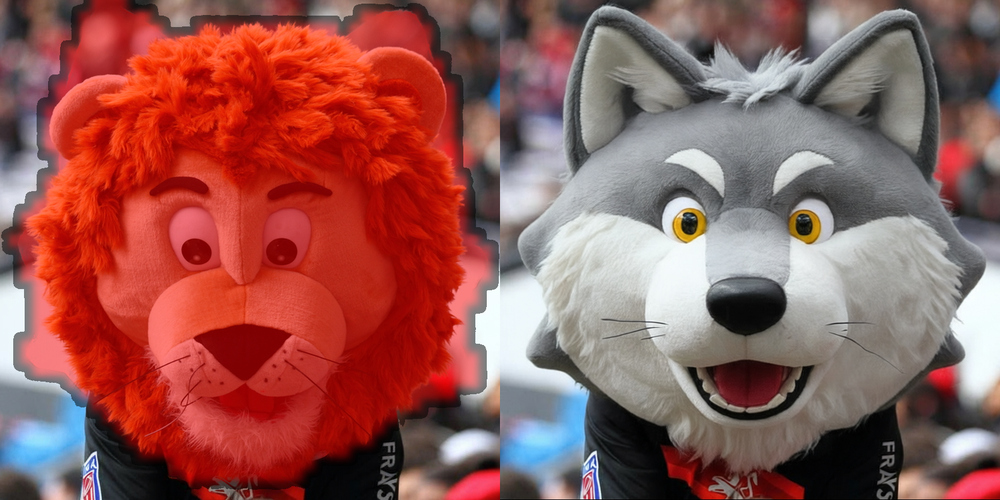} & \includegraphics[width=0.30\linewidth]{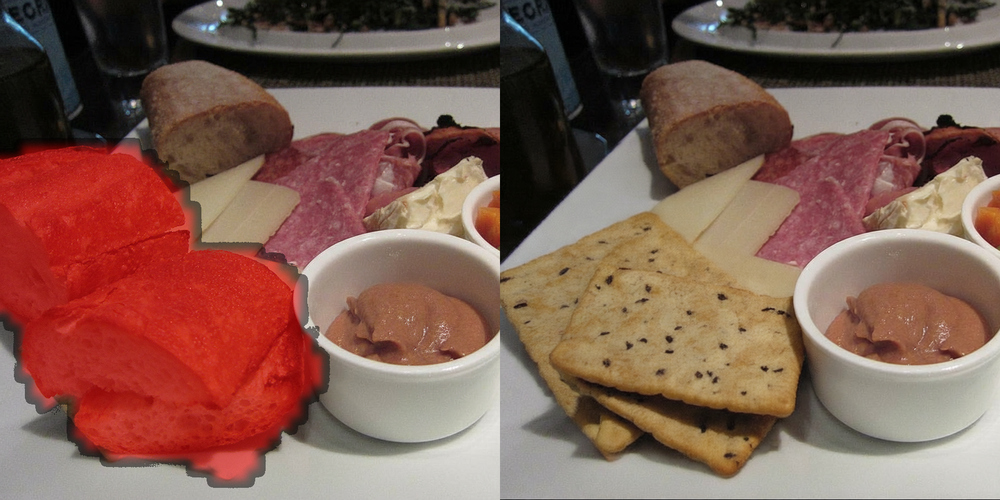} & \includegraphics[width=0.30\linewidth]{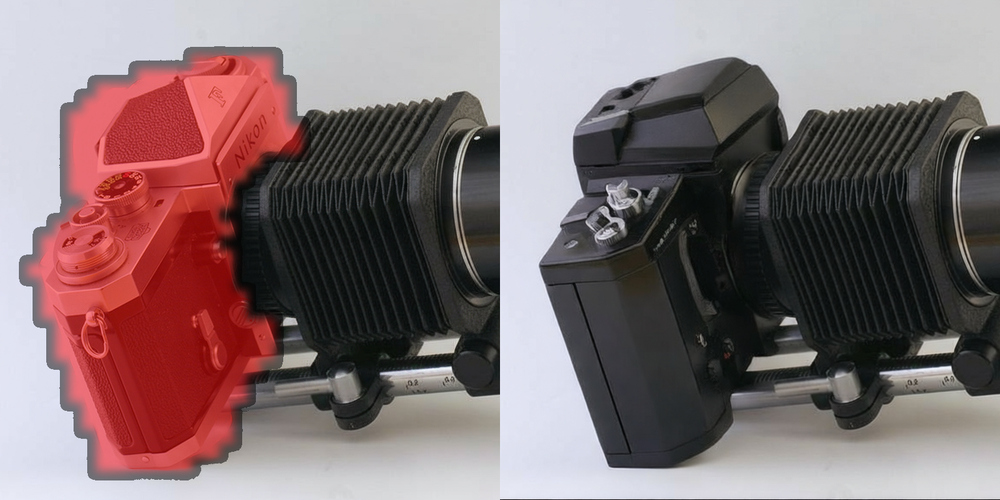}\\
\parbox{0.30\linewidth}{\centering\scriptsize \textit{Change the lion to a friendly, plush grey and white wolf with...}} & \parbox{0.30\linewidth}{\centering\scriptsize \textit{Change the rustic bread to light-brown, multi-grain crackers, keeping soft lighting and...}} & \parbox{0.30\linewidth}{\centering\scriptsize \textit{Change the vintage camera to a sleek black digital one, keeping it...}}\\
\includegraphics[width=0.30\linewidth]{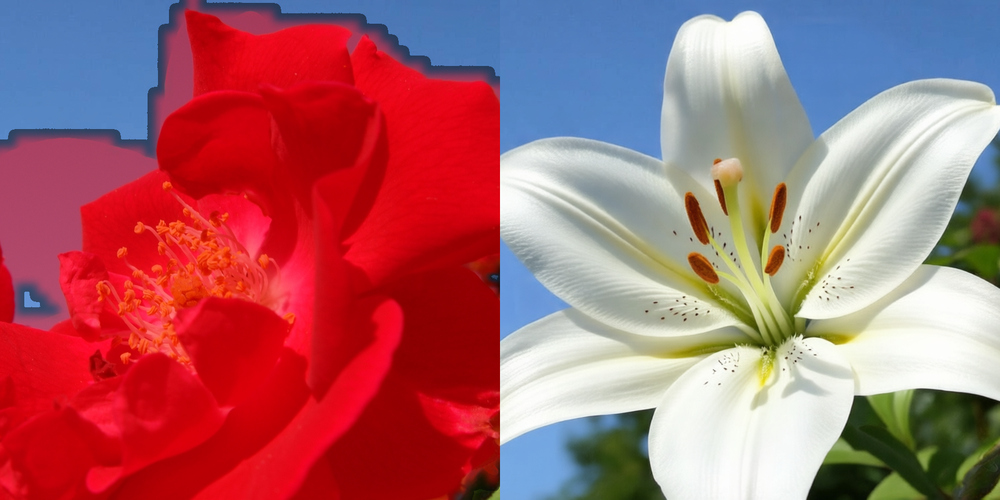} & \includegraphics[width=0.30\linewidth]{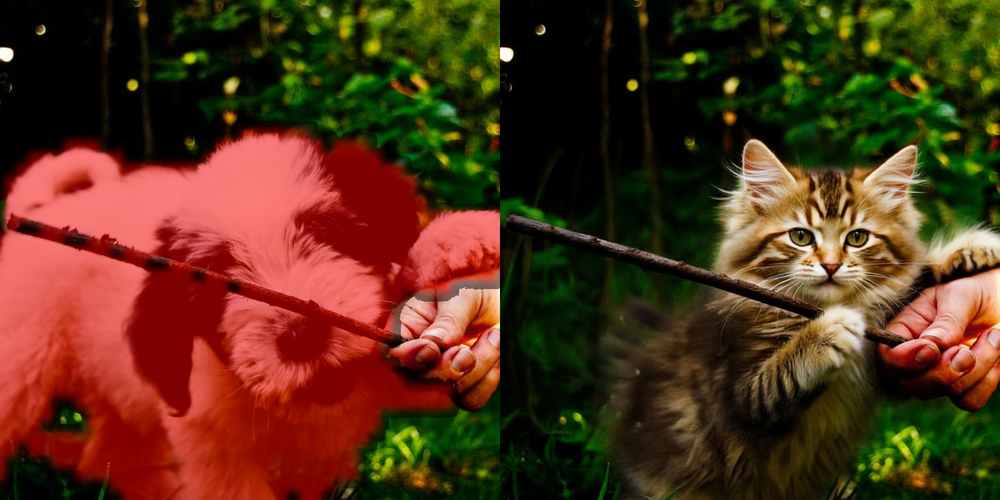} & \includegraphics[width=0.30\linewidth]{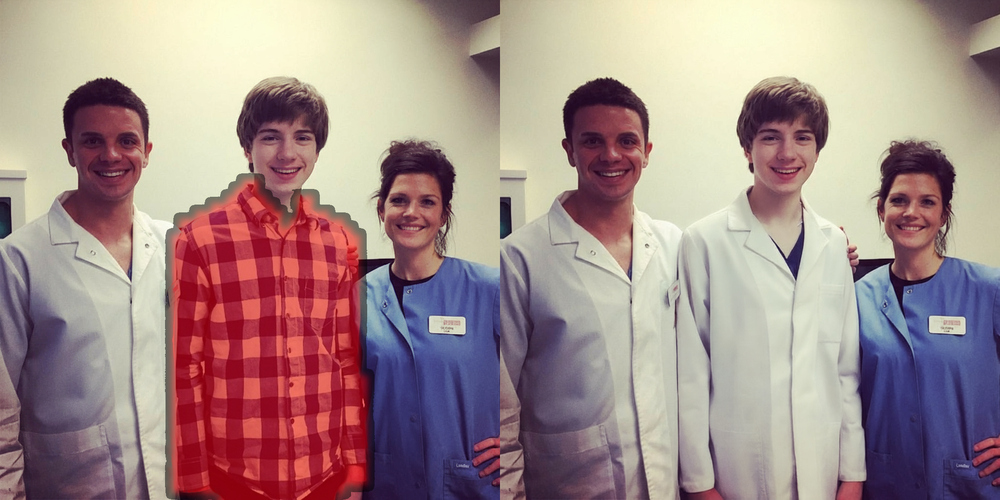}\\
\parbox{0.30\linewidth}{\centering\scriptsize \textit{Replace the red rose with a delicate white lily, keeping the sunlit...}} & \parbox{0.30\linewidth}{\centering\scriptsize \textit{Add a playful, fluffy kitten with paws on the stick and hand,...}} & \parbox{0.30\linewidth}{\centering\scriptsize \textit{Change the middle person's blue plaid shirt to a matching white lab...}}\\
\end{tabular}
\caption{Additional results for \textbf{Replace} (2/2).}
\end{figure}
\clearpage

\section{Selected Instruction Examples}
\label{sec:selected_prompts}

To provide further insight into our VLM-driven automated pipeline, we list several representative ``Result-Oriented Dense Captions'' generated by Qwen3-VL during our data curation process. To address keyword bias and token limits in imperative instructions, these dense captions (incorporating both global and local scene descriptions) ensure that the text aligns closely with the intended visual outcome.

\begin{sloppypar}
\begin{itemize}
    \item \textbf{Category Replacement:} 
    \begin{itemize}
        \item \textit{Instruction:} ``Change the horse to a majestic white unicorn
        with shimmering coat, silver mane and tail, and an elegant golden horn.''
        \item \textit{Global Caption:} ``A rider in equestrian gear gently touches a majestic white unicorn with a shimmering coat, silver mane and tail, and an elegant golden horn.''
        \item \textit{Local Caption:} ``a white majestic unicorn with silver mane and tail, and an elegant golden horn''
    \end{itemize}

    \item \textbf{Object Addition:}
    \begin{itemize}
        \item \textit{Instruction:} ``Add a perfectly halved, marinated soft-boiled egg (ajitsuke tamago) to the bowl, nestled gently into the broth among the noodles and existing toppings on the left side, ensuring its creamy, golden-orange yolk is visible and its glossy, light brown marinated white reflects the warm overhead light...''
        \item \textit{Global Caption:} ``A steaming bowl of ramen with noodles,
        seaweed, mushrooms, cilantro, and a perfectly halved marinated soft-boiled egg on the left, its golden yolk and glossy white reflecting warm light, all under shallow depth of field.''
        \item \textit{Local Caption:} ``A halved marinated soft-boiled egg with creamy golden-orange yolk and glossy light brown white, nestled in broth on the left, reflecting warm overhead light.''
    \end{itemize}

    \item \textbf{Object Removal:}
    \begin{itemize}
        \item \textit{Instruction:} ``Remove the child and pony, replacing them with lush green grass matching the existing lawn.''
        \item \textit{Global Caption:} ``A grassy lawn with palm fronds on the left and green barriers in the background, uniformly covered in lush green grass.''
        \item \textit{Local Caption:} ``Lush green grass seamlessly matching the existing lawn, replacing any non-grass elements.''
    \end{itemize}

    \item \textbf{Complex Scene Integration:}
    \begin{itemize}
        \item \textit{Instruction:} ``Integrate a sleek, dark metallic condenser microphone on a subtle stand to the right of the woman, positioned slightly in front of her mouth to suggest an interview or performance, ensuring its surface reflects the warm, soft lighting present in the scene and casts a soft shadow consistent with the existing environment.''
        \item \textit{Global Caption:} ``A woman with curly hair smiles into a sleek dark metallic condenser microphone on a subtle stand, positioned slightly in front of her mouth, under warm soft lighting with reflective surfaces and soft shadows.''
        \item \textit{Local Caption:} ``A sleek dark metallic condenser microphone on a subtle stand, reflecting warm soft light and casting a soft shadow, positioned slightly in front of the woman's mouth.''
    \end{itemize}

    \item \textbf{Structural Accessory Modification:}
    \begin{itemize}
        \item \textit{Instruction:} ``Change the product box to a modern, glossy, luxurious package with a sophisticated logo, keeping the hand holding it in place.''
        \item \textit{Global Caption:} ``A hand holds a modern, glossy, luxurious
        blue product box with a sophisticated silver logo, set against a wooden
        table background.''
        \item \textit{Local Caption:} ``a sleek, glossy blue box with a refined silver circular logo and elegant text, exuding luxury and sophistication''
    \end{itemize}
\end{itemize}
\end{sloppypar}

\clearpage

\section{Support and Trade-off Analysis}
\label{sec:failure_cases}
\enlargethispage{2\baselineskip}

\paragraph{Inference Cost.}
We measured complete-pipeline inference after model loading on an NVIDIA RTX PRO 6000 Blackwell Max-Q using BF16, no offload, 1024$\times$1024 resolution, 50 steps, CFG 4, seed 0, an exact 25\% edit region, and one full warmup per process. The matched Qwen and BRIDGE runs use the same fused rank-512 LoRA. Table~\ref{tab:inference_cost} reports latency and peak process VRAM.

\begin{table}[H]
    \centering
    \caption{End-to-end inference measurement at 1024$\times$1024 with an exact 25\% edit region. Latency is mean $\pm$ standard deviation and excludes model loading. Qwen and BRIDGE use the same fused rank-512 LoRA; Q-Control uses its released base checkpoint.}
    \label{tab:inference_cost}
    \small
    \resizebox{0.94\linewidth}{!}{%
    \begin{tabular}{lrrrrr}
        \toprule
        \textbf{Method} & $\boldsymbol{n}$ & \textbf{Latency (s)} & \textbf{vs. Qwen} & \textbf{Peak VRAM (GiB)} & \textbf{vs. Qwen} \\
        \midrule
        Qwen baseline & 5 & $91.985 \pm 0.205$ & --- & 57.354 & --- \\
        \textbf{BRIDGE} & 3 & $111.649 \pm 0.193$ & +21.38\% & 57.379 & +0.04\% \\
        Q-Control & 5 & $112.311 \pm 1.908$ & +22.10\% & 59.482 & +3.71\% \\
        \bottomrule
    \end{tabular}}
\end{table}

For fixed conditioning lengths, Subject Path token count grows approximately with edit area; at full-image editing, it approaches one additional image-token grid. The percentage overhead therefore also depends on the full text/image-conditioning sequence length. The measured BRIDGE latency increase has a 21.07\%--21.64\% bootstrap 95\% interval, while its 26 MiB peak-process increase is consistent with the 13.31M BF16 GateBlock parameters. Dynamic token cropping and packed-latent execution remain directions for reducing redundant Subject Path computation.

\FloatBarrier
\section{FLUX.2 Cross-Backbone Extension Details}
\label{sec:flux2_extension_appendix}

\paragraph{Training records and branch targets.}
The FLUX.2 extension uses 27,834 training records.
Table~\ref{tab:flux2_data_roles_full} distinguishes the full-canvas target,
source condition, subject-reference condition, and Sub-branch crop target. The
subject-reference images were generated with Qwen-Image-Edit-2511~\citep{qwenimage}
and selected for subject identity and visual quality. They condition the model
and do not serve as the Sub-branch target.

\begin{table}[H]
    \centering
    \caption{Image roles in the FLUX.2 subject-driven training records.}
    \label{tab:flux2_data_roles_full}
    \small
    \setlength{\tabcolsep}{4pt}
    \begin{tabular}{p{0.32\linewidth}p{0.59\linewidth}}
        \toprule
        \textbf{Record field} & \textbf{Training role} \\
        \midrule
        \texttt{image} & Full edited target for the Main branch. \\
        \texttt{edit\_image} & Full-canvas source/background image condition. \\
        \texttt{generated\_subject\_image} & Generated and selected subject-reference image condition. \\
        \texttt{sub} & Aligned crop of the edited subject used as the Sub-branch target. \\
        \texttt{back\_mask} & Spatial support used to derive the bounding box and positional correspondence. \\
        \bottomrule
    \end{tabular}
\end{table}

For the evaluated Dense/BBox variant, the Sub branch retains every token in the
bounding rectangle derived from \texttt{back\_mask}. Main and Sub targets use
independently sampled Gaussian noise. The visual sequence is ordered as noisy
Main tokens, noisy Sub tokens, source/background condition tokens, and, when
retained by subject-image dropout, subject-reference condition tokens. The
routing module observes the joint visual hidden states. The backbone-specific
STE parameter counts are reported in Table~\ref{tab:ste_parameter_accounting}.
The Main and Sub flow-matching losses are combined with equal default weight. The evaluated
checkpoint was trained for 1,400 optimizer steps on eight GPUs with a per-GPU
micro-batch size of two and gradient accumulation of eight, giving an effective
global batch size of 128.

\subsection{Training Dynamics of Dense/BBox and Sparse/Mask Runs}
\label{sec:flux2_training_dynamics}

\begin{figure}[H]
    \centering
    \includegraphics[width=0.32\linewidth]{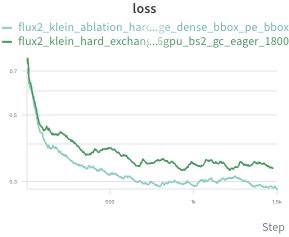}\hfill
    \includegraphics[width=0.32\linewidth]{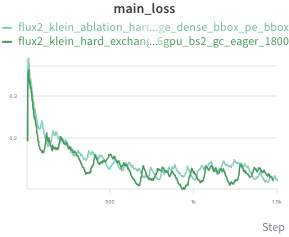}\hfill
    \includegraphics[width=0.32\linewidth]{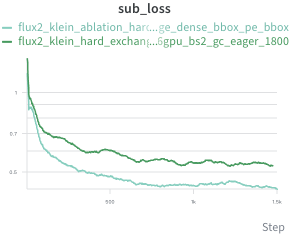}

    \vspace{4pt}
    \includegraphics[width=0.32\linewidth]{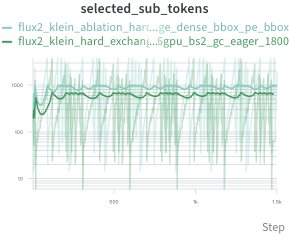}\hfill
    \includegraphics[width=0.32\linewidth]{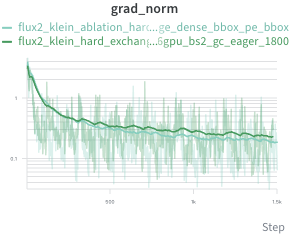}\hfill
    \includegraphics[width=0.32\linewidth]{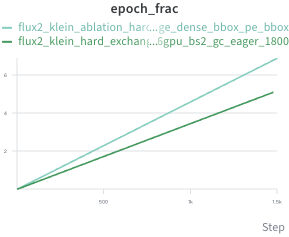}
    \caption{\textbf{Training dynamics for the two FLUX.2 checkpoints.}
    Light cyan denotes Dense/BBox training and dark green denotes Sparse/Mask
    training. From top left to bottom right: total loss, Main loss, Sub loss,
    selected Sub-token count, gradient norm, and fractional data passes. The
    different sample-exposure rates reflect the eight-GPU and six-GPU training
    configurations.}
    \label{fig:flux_training_dynamics}
\end{figure}

Figure~\ref{fig:flux_training_dynamics} reports the complete logged training
curves supplied for the Dense/BBox and Sparse/Mask checkpoints. The light-cyan
curve is the eight-GPU Dense/BBox run
(\texttt{use\_sparse\_sub\_branch=false}); the dark-green curve is the six-GPU
Sparse/Mask run (\texttt{use\_sparse\_sub\_branch=true}). Both use equal Main
and Sub loss weights. The plots record total, Main, and Sub losses, selected Sub
tokens, gradient norm, and fractional data passes. Both loss trajectories
decrease and remain finite over the recorded interval. Because the runs use
different GPU counts and effective global batches, equal optimizer steps do not
correspond to equal sample exposure; the curves document optimization behavior
and are not an equal-budget quality comparison.

\subsection{Why BBox Training Uses Optional Mask-Cropped Inference}
\label{sec:flux2_bbox_sparse_rationale}

The dense and sparse variants address different failure modes. With a dense
BBox Sub branch, every cell in the bounding rectangle receives a Sub token.
This preserves a complete local latent grid and produces stronger Sub-image
quality, but a non-rectangular user mask leaves rectangular support in regions
where the user did not request new content. The model can consequently generate
throughout the bounding rectangle.

\begin{figure}[H]
    \centering
    \includegraphics[width=\linewidth,height=0.79\textheight,keepaspectratio]{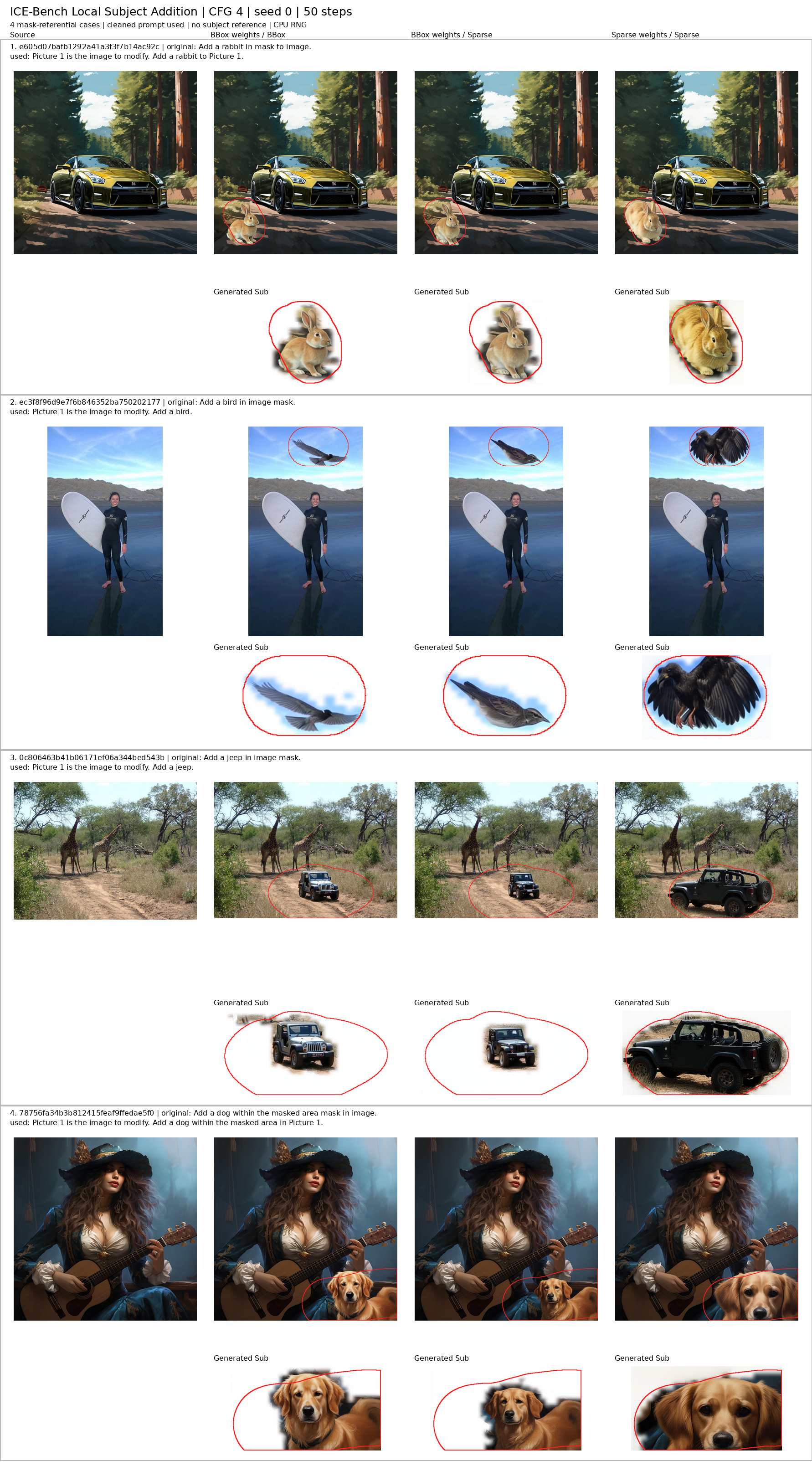}
    \caption{\textbf{Qualitative motivation for BBox training.} Four ICE
    Addition cases at CFG $=4$ compare the Dense/BBox and separately trained
    Sparse/Mask variants. In the first example, the mask retains only a small
    number of Sub tokens; the decoded Sparse/Mask Sub image exhibits an
    incorrect color distribution and closely fills the supplied mask contour,
    revealing severe mask-shape overfitting. The remaining examples show the
    same quality and support-control trade-off. The final design therefore
    retains BBox training and applies mask cropping only at inference with the
    same BBox-trained checkpoint.}
    \label{fig:flux_sparse_bbox_five_cases}
\end{figure}

We initially addressed this mismatch by training a Sparse/Mask model whose Sub
sequence retained only mask-selected cells. Development comparisons revealed
three recurring problems: the decoded Sub image was less coherent than that of
the BBox-trained model, small supports exhibited visible color drift, and the
model learned to fill the supplied mask shape. The last behavior reintroduced
the mask-shape dependence that BRIDGE is designed to reduce. These observations
motivated retaining dense BBox supervision during training, as illustrated in
Fig.~\ref{fig:flux_sparse_bbox_five_cases}.

\clearpage

The final support-control design applies sparsification at inference to the
same BBox-trained checkpoint. It removes Sub tokens outside the user mask and
restricts the corresponding PE candidate pairs to the retained cells. In the
no-Sub diagnostic, changing the prompt did not visibly alter the source
background when the Sub branch was absent. Removing outside-mask Sub support
therefore makes those locations approximate the no-Sub condition while the
retained local tokens continue to generate the requested subject. This combines
the subject quality of BBox training with tighter user-mask support and avoids
training the model to reproduce the mask silhouette.

\raggedbottom
\subsection{Coercive-Mask Stress Test and STE Routing Dynamics}
\label{sec:flux2_token_control_evidence}

\begin{figure}[H]
    \centering
    \includegraphics[width=\linewidth,height=0.75\textheight,keepaspectratio]{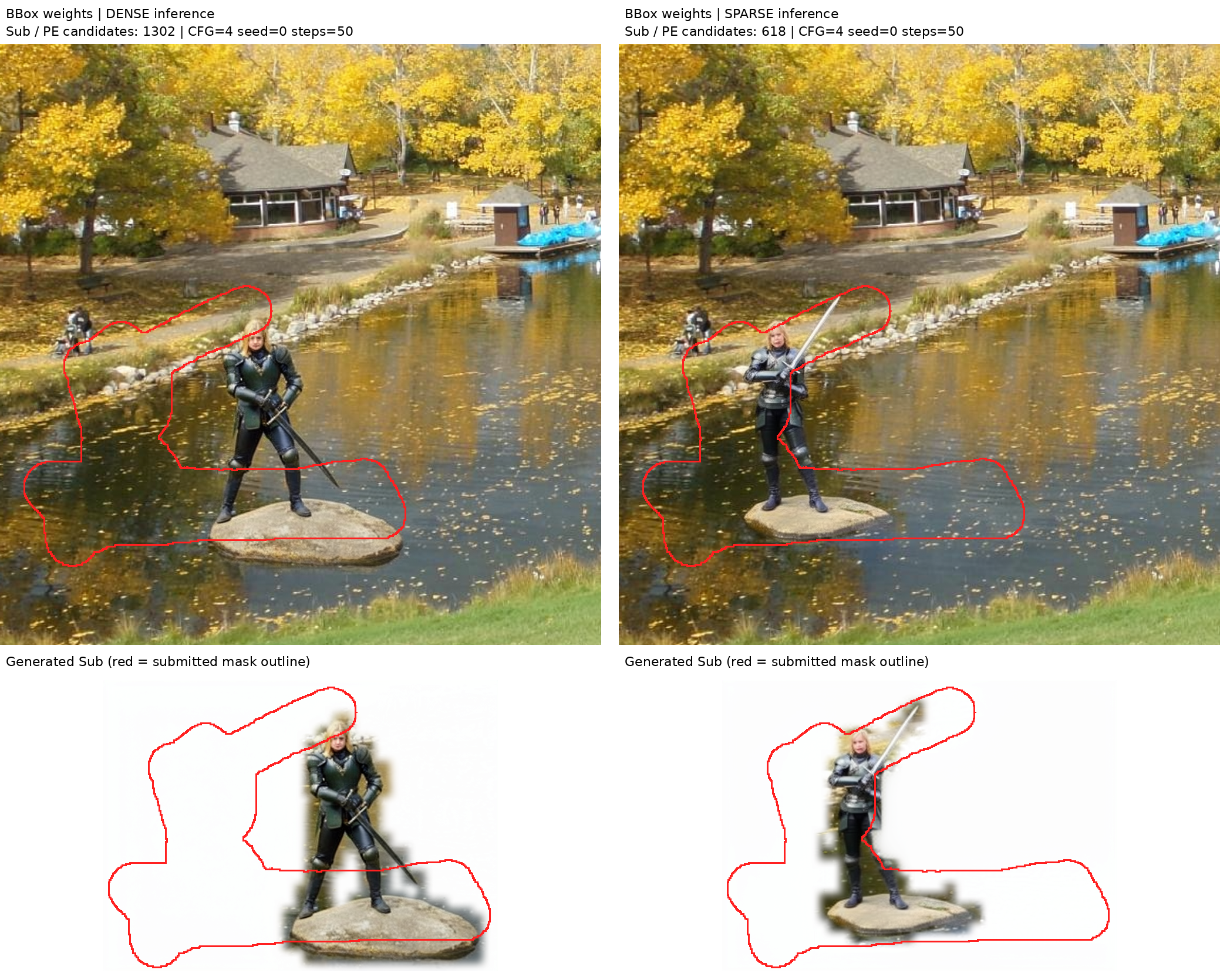}
    \caption{\textbf{Same-checkpoint coercive-mask stress test.} The left column
    uses Dense/BBox inference with 1,302 Sub tokens; the right column applies
    mask-cropped inference with 618 retained tokens. Both use the BBox-trained
    checkpoint and identical generation settings without latent blending. The
    red curve is the submitted mask boundary and is shown only for diagnosis.
    The top row is the decoded Main output and the bottom row is the decoded Sub
    output.}
    \label{fig:flux_coercive_mask_comparison}
\end{figure}

We further evaluate the support-control mechanism on a web-sourced qualitative
stress case with an intentionally irregular, coercive mask. Both inference arms
in Fig.~\ref{fig:flux_coercive_mask_comparison} use the same BBox-trained
checkpoint, source image, prompt, subject references, submitted mask, initial
noise, seed 0, CFG 4, and 50 denoising steps. Latent blending is disabled in
both arms. Dense/BBox inference retains all 1,302 Sub tokens inside the enclosing
rectangle, whereas mask-cropped inference retains the 618 mask-selected tokens
and restricts the PE candidate pairs to the same support. The decoded Sub views
make the support mismatch visible: the dense route allocates generation capacity
throughout the rectangle, while the cropped route confines active Sub content to
the user-selected cells. Earlier Qwen experiments used inference-time $\alpha$
blending for background preservation; this comparison disables blending and
isolates Sub-token support control.

Table~\ref{tab:token_control_file_map} provides the exact relative paths in the
anonymous supplementary archive. Reviewers can begin with the output comparison,
then inspect the complete mask-cropped animation, and use the remaining records
for the matched Dense/BBox control or layer-specific views.

\begin{table}[H]
    \centering
    \caption{Reviewer file map for the anonymous token-control supplementary
    archive. Paths are relative to the archive root. Layer indices run from
    \texttt{00} to \texttt{31}; denoising-step indices run from \texttt{00} to
    \texttt{49}.}
    \label{tab:token_control_file_map}
    \scriptsize
    \setlength{\tabcolsep}{4pt}
    \begin{tabular}{p{0.54\linewidth}p{0.38\linewidth}}
        \toprule
        \textbf{Relative path} & \textbf{Content} \\
        \midrule
        \path{README.md} & Quick-start reading order, protocol, file map, and anonymization record. \\
        \path{comparison.png} & Same-checkpoint Dense/BBox versus mask-cropped Main and decoded-Sub comparison. \\
        \path{mask_cropped/ste_masks/run/ste_layer_exchange_timestep_grid.gif} & Primary 32-layer, 50-step STE animation for mask-cropped Sub inference. \\
        \path{dense/ste_masks/run/ste_layer_exchange_timestep_grid.gif} & Matched 32-layer, 50-step Dense/BBox routing control. \\
        \path{mask_cropped/ste_masks/run/ste_layer_exchange_by_timestep/layer_XX_exchange.gif} & Animation for one selected transformer layer. \\
        \path{mask_cropped/ste_masks/run/ste_layer_exchange_by_timestep/layer_XX/step_YY.png} & Raw hard-forward routing mask for one layer and denoising step. \\
        \path{experiment.json} & Effective CFG, step count, seed, checkpoint role, and no-blending setting. \\
        \path{verification.json} & Matched-condition checks and the 1,302/618 Sub-token counts. \\
        \bottomrule
    \end{tabular}
\end{table}

\begin{figure}[H]
    \centering
    \includegraphics[width=\linewidth,height=0.78\textheight,keepaspectratio]{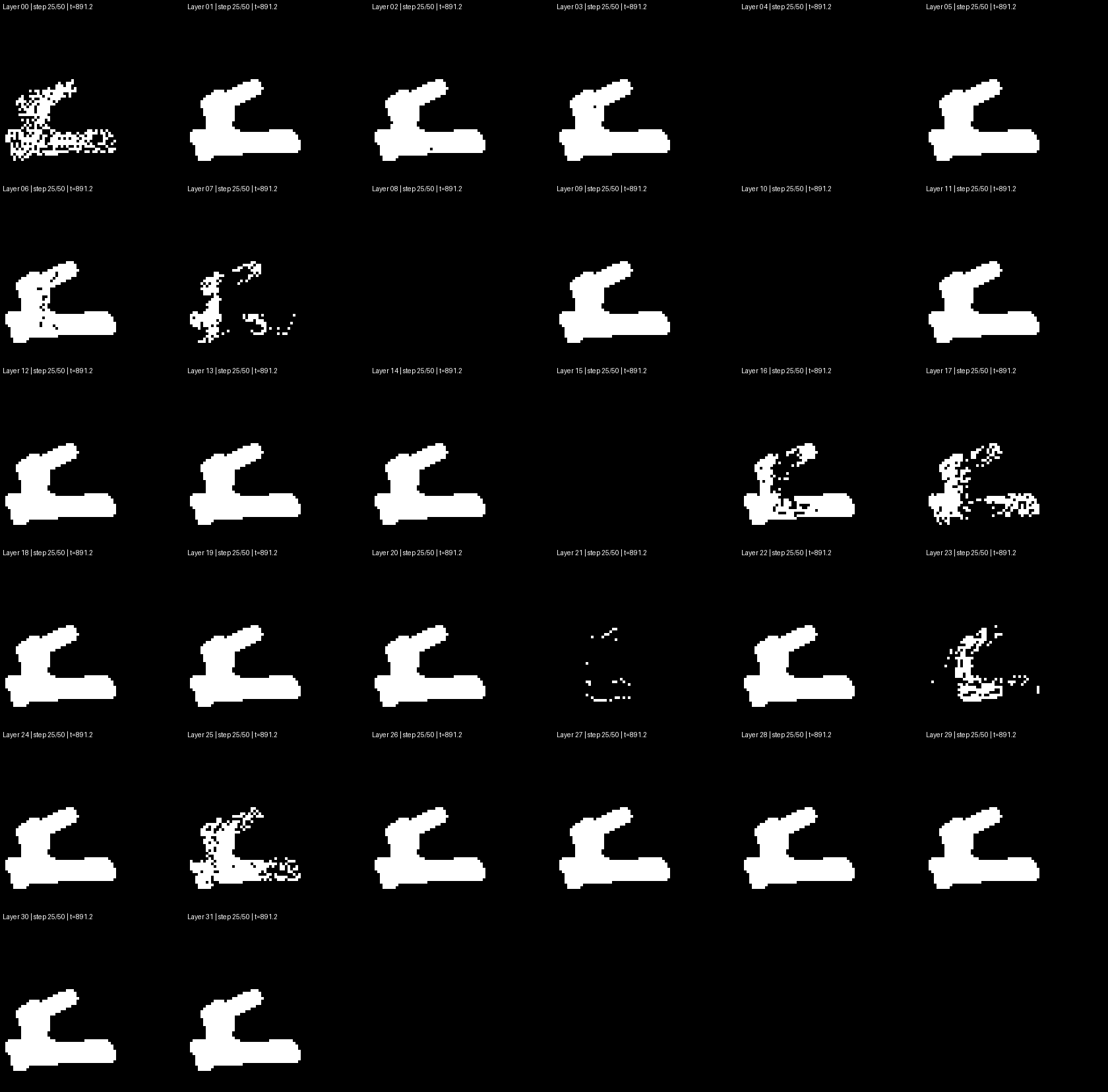}
    \caption{\textbf{STE routing on mask-cropped Sub tokens.} Hard-forward PE
    exchange decisions for all 32 layers at denoising step 25 of 50. White denotes
    exchange to the paired Main-image PE and black denotes retention of the Sub
    PE. The complete 50-frame temporal evolution is provided in the anonymous
    supplementary archive at the exact path listed in
    Table~\ref{tab:token_control_file_map}.}
    \label{fig:flux_ste_mask_step25}
\end{figure}

Figure~\ref{fig:flux_ste_mask_step25} visualizes the hard-forward STE decisions
for all 32 transformer layers at denoising step 25 in the mask-cropped arm. White
cells use the paired Main-image positional embedding, while black cells retain
the original Sub positional embedding. Mask cropping first determines which Sub
tokens exist; the learned gate then selects the positional source independently
for every retained token and layer. The anonymous supplementary archive includes
the full 50-frame layer-by-timestep GIFs for both Dense/BBox and mask-cropped
inference, together with the individual frames, aggregate exchange maps, and
sanitized configuration records; Table~\ref{tab:token_control_file_map} gives
their exact relative paths.

\flushbottom

The ICE-Bench scores in Tables~\ref{tab:flux2_grouped_comparison}
and~\ref{tab:flux2_per_task} use the full Dense/BBox inference path
(\texttt{use\_sparse\_sub\_branch=false} with BBox PE exchange). The optional
mask-cropped mode uses the same checkpoint and is reported separately as a
support-control analysis.

\subsection{ICE-Bench Protocol and Score Provenance}
\label{sec:flux2_grouped_comparison_provenance}

\paragraph{Prompt-normalized ICE-Bench protocol.}
The evaluation covers Tasks 17--22 and 29--31, the nine ICE-Bench tasks that
provide \texttt{SourceMask}~\citep{pan2025icebench}. GPT-6 Luna receives the
original instruction, source image, mask, and any available subject-reference
image, and rewrites the instruction into a self-contained prompt. ``Picture 1''
denotes the source image and ``Picture 2'' denotes the reference when present.
The rewriter does not receive the target image or \texttt{TargetCaption}. The
source image remains unmodified for every task, including Task 17, and
\texttt{SourceMask} is supplied as a separate spatial input. The Dense/BBox checkpoint uses CFG 4,
50 steps, seed 0, global Sub spatial IDs, and no latent blending. Official scoring uses
the original \texttt{Instruction}, \texttt{TargetCaption}, task-to-metric
mapping, and aggregation code.

\paragraph{Mask-group model comparisons.}
The columns in Table~\ref{tab:flux2_grouped_comparison} summarize all
ICE-Bench tasks with a \texttt{SourceMask}: Tasks 17--22, 29--30, and 31,
following the corresponding groups in Emu3.5 Table~13~\citep{cui2025emu35}.
Its upper block contains open-source models, while the lower block reports the
proprietary Gemini references. FLUX.2 BRIDGE was not trained for Tasks 17, 18,
or 31. Tasks 19--22 average 0.5941; later papers report only their combined
Tasks 17--22 aggregate. The Task 29--30 value 0.6254 is the simple
mean of Subject-guided Inpainting and Virtual Try On, and Task 31 contributes
the Face Swap score 0.5700 directly.

The ICE-Bench original-paper block is computed from the approximate per-task
scores in Table~\ref{tab:flux2_per_task}: Tasks 17--22 and Tasks 29--30 are
simple group means, while Task 31 is copied directly. In the original-model
block of Table~\ref{tab:flux2_grouped_comparison}, Tasks 17 and 18 are reported
separately and the \texttt{19--22} column averages the four trained local-edit tasks.
The later open-source
and closed-model values are transcribed from Emu3.5 Table~13. FLUX.2 BRIDGE
was evaluated on these nine \texttt{SourceMask} tasks, so the comparison
reports the three corresponding mask groups.

\paragraph{Per-task results.}
Table~\ref{tab:flux2_per_task} separates the current FLUX.2 run from external
reference values. The FLUX.2 values are computed from the per-example
ICE-Bench records. The baseline values are
reconstructed from the rounded component metrics in the ICE-Bench
supplementary material using its official scoring formula, so their displayed
four-decimal task scores are approximate. Inpainting, Outpainting, and Face
Swap are completed diagnostic runs because the model was not trained for these
tasks; all three are excluded from the supported six-task mean.

\begin{table}[H]
    \centering
    \caption{Per-task ICE-Bench scores for tasks with a \texttt{SourceMask}.
    Tasks 17, 18, and 31 are diagnostics because FLUX.2 BRIDGE was not trained
    for Inpainting, Outpainting, or Face Swap. ACE++, ACE, OmniGen, and UltraEdit are external reference
    values reconstructed from rounded metrics reported by
    ICE-Bench~\citep{pan2025icebench}; they are not new local reruns.}
    \label{tab:flux2_per_task}
    \small
    \resizebox{\linewidth}{!}{%
    \setlength{\tabcolsep}{3.5pt}
    \begin{tabular}{lrrrrr}
        \toprule
        \textbf{ICE-Bench task} & \textbf{Ours (FLUX Subgroup)} & \textbf{ACE++} & \textbf{ACE} & \textbf{OmniGen} & \textbf{UltraEdit} \\
        \midrule
        17: Inpainting (diagnostic) & \textbf{0.6506} & 0.6430 & 0.5954 & 0.5454 & 0.4520 \\
        18: Outpainting (diagnostic) & 0.5801 & \textbf{0.6230} & 0.5528 & 0.5733 & 0.4784 \\
        19: Local Subject Addition & \textbf{0.6373} & 0.6286 & 0.5623 & 0.5482 & 0.5501 \\
        20: Local Subject Removal & \textbf{0.5906} & 0.5464 & 0.5733 & 0.5631 & 0.4963 \\
        21: Local Text Render & 0.5564 & \textbf{0.5593} & 0.5514 & 0.4708 & 0.4462 \\
        22: Local Text Removal & 0.5922 & \textbf{0.5977} & 0.5910 & 0.5635 & 0.5625 \\
        29: Subject-guided Inpainting & \textbf{0.5733} & 0.5406 & 0.5081 & 0.4936 & --- \\
        30: Virtual Try On & \textbf{0.6775} & 0.6725 & 0.6362 & 0.6581 & --- \\
        31: Face Swap (diagnostic) & \textbf{0.5700} & 0.5396 & 0.4907 & 0.5324 & --- \\
        \midrule
        Nine-task mean (incl. diagnostics) & \textbf{0.6031} & 0.5945 & 0.5624 & 0.5498 & --- \\
        Supported six-task mean & \textbf{0.6045} & 0.5908 & 0.5704 & 0.5496 & --- \\
        \bottomrule
    \end{tabular}}
\end{table}

\subsection{Per-Task ICE-Bench Component Comparisons}
\label{sec:flux2_complete_metric_statistics}

Tables~\ref{tab:flux2_task17_components}--\ref{tab:flux2_task31_components}
compare the FLUX Subgroup run with the models reported for each corresponding
task in the official ICE-Bench supplementary tables~\citep{pan2025icebench}.
Task 17--22 include ACE, OmniGen, ACE++, and UltraEdit. The official Task
29--31 tables report ACE, OmniGen, and ACE++ only, so UltraEdit is not added to
those comparisons. Published component metrics are transcribed at their
reported three-decimal precision. Their TaskScore values are recomputed from
those rounded components with the official \texttt{cal\_scores.py} formula and
are therefore approximate. The Ours TaskScore values use the unrounded
aggregate metrics from the current run. Bold marks the best displayed result
in each column; lower L1-src is better and all other metrics are higher-is-better.

\begin{table}[H]
    \centering
    \caption{ICE-Bench component metrics for Task 17 (Inpainting); diagnostic because this task was not used for task-specific training.}
    \label{tab:flux2_task17_components}
    \scriptsize
    \setlength{\tabcolsep}{2.8pt}
    \resizebox{\linewidth}{!}{%
    \begin{tabular}{lrrrrrrr}
        \toprule
        \textbf{Method} & \textbf{Aesthetic} & \textbf{Imaging} & \textbf{CLIP-cap} & \textbf{VLLM-QA} & \textbf{CLIP-src} & \textbf{L1-src} & \textbf{TaskScore} \\
        \midrule
        ACE & 4.878 & 51.793 & 0.269 & 0.833 & 0.785 & 0.024 & 0.5954 \\
        OmniGen & 4.545 & 59.264 & 0.238 & 0.524 & 0.734 & 0.108 & 0.5454 \\
        ACE++ & 5.064 & \textbf{61.661} & 0.272 & 0.910 & 0.776 & \textbf{0.016} & 0.6430 \\
        UltraEdit & 3.817 & 46.284 & 0.250 & 0.180 & \textbf{0.952} & 0.019 & 0.4520 \\
        \textbf{Ours (FLUX Subgroup)} & \textbf{5.316} & 60.457 & \textbf{0.277} & \textbf{0.923} & 0.785 & 0.019 & \textbf{0.6506} \\
        \bottomrule
    \end{tabular}}
\end{table}

\begin{table}[H]
    \centering
    \caption{ICE-Bench component metrics for Task 18 (Outpainting); diagnostic because this task was not used for task-specific training.}
    \label{tab:flux2_task18_components}
    \scriptsize
    \setlength{\tabcolsep}{2.8pt}
    \resizebox{\linewidth}{!}{%
    \begin{tabular}{lrrrrrrr}
        \toprule
        \textbf{Method} & \textbf{Aesthetic} & \textbf{Imaging} & \textbf{CLIP-cap} & \textbf{VLLM-QA} & \textbf{CLIP-src} & \textbf{L1-src} & \textbf{TaskScore} \\
        \midrule
        ACE & 5.514 & 50.403 & 0.287 & 0.376 & 0.891 & 0.017 & 0.5528 \\
        OmniGen & 5.442 & \textbf{65.758} & 0.265 & 0.326 & 0.802 & 0.114 & 0.5733 \\
        ACE++ & \textbf{5.644} & 64.156 & \textbf{0.289} & \textbf{0.531} & 0.908 & \textbf{0.010} & \textbf{0.6230} \\
        UltraEdit & 4.498 & 43.968 & 0.274 & 0.220 & \textbf{0.945} & 0.018 & 0.4784 \\
        \textbf{Ours (FLUX Subgroup)} & 5.200 & 61.141 & 0.283 & 0.408 & 0.942 & 0.049 & 0.5801 \\
        \bottomrule
    \end{tabular}}
\end{table}

\begin{table}[H]
    \centering
    \caption{ICE-Bench component metrics for Task 19 (Local Subject Addition).}
    \label{tab:flux2_task19_components}
    \scriptsize
    \setlength{\tabcolsep}{2.8pt}
    \resizebox{\linewidth}{!}{%
    \begin{tabular}{lrrrrrrr}
        \toprule
        \textbf{Method} & \textbf{Aesthetic} & \textbf{Imaging} & \textbf{CLIP-cap} & \textbf{VLLM-QA} & \textbf{CLIP-src} & \textbf{L1-src} & \textbf{TaskScore} \\
        \midrule
        ACE & 4.965 & 51.704 & 0.272 & 0.555 & 0.897 & 0.029 & 0.5623 \\
        OmniGen & 4.584 & 58.911 & 0.249 & 0.479 & 0.814 & 0.066 & 0.5482 \\
        ACE++ & 5.014 & \textbf{62.083} & 0.268 & 0.785 & 0.894 & 0.018 & 0.6286 \\
        UltraEdit & 4.881 & 47.855 & 0.275 & 0.555 & 0.909 & 0.021 & 0.5501 \\
        \textbf{Ours (FLUX Subgroup)} & \textbf{5.170} & 59.979 & \textbf{0.279} & \textbf{0.823} & \textbf{0.919} & \textbf{0.015} & \textbf{0.6373} \\
        \bottomrule
    \end{tabular}}
\end{table}

\begin{table}[H]
    \centering
    \caption{ICE-Bench component metrics for Task 20 (Local Subject Removal).}
    \label{tab:flux2_task20_components}
    \scriptsize
    \setlength{\tabcolsep}{2.8pt}
    \resizebox{\linewidth}{!}{%
    \begin{tabular}{lrrrrrrr}
        \toprule
        \textbf{Method} & \textbf{Aesthetic} & \textbf{Imaging} & \textbf{CLIP-cap} & \textbf{VLLM-QA} & \textbf{CLIP-src} & \textbf{L1-src} & \textbf{TaskScore} \\
        \midrule
        ACE & 4.996 & 47.011 & \textbf{0.258} & \textbf{0.757} & 0.852 & 0.024 & 0.5733 \\
        OmniGen & 4.792 & 54.320 & 0.238 & 0.658 & 0.787 & 0.061 & 0.5631 \\
        ACE++ & 5.061 & \textbf{61.614} & 0.229 & 0.312 & 0.901 & 0.017 & 0.5464 \\
        UltraEdit & 4.858 & 48.748 & 0.226 & 0.287 & 0.888 & 0.018 & 0.4963 \\
        \textbf{Ours (FLUX Subgroup)} & \textbf{5.158} & 57.486 & 0.255 & 0.614 & \textbf{0.913} & \textbf{0.014} & \textbf{0.5906} \\
        \bottomrule
    \end{tabular}}
\end{table}

\begin{table}[H]
    \centering
    \caption{ICE-Bench component metrics for Task 21 (Local Text Render).}
    \label{tab:flux2_task21_components}
    \scriptsize
    \setlength{\tabcolsep}{2.8pt}
    \resizebox{\linewidth}{!}{%
    \begin{tabular}{lrrrrrrr}
        \toprule
        \textbf{Method} & \textbf{Aesthetic} & \textbf{Imaging} & \textbf{CLIP-cap} & \textbf{VLLM-QA} & \textbf{CLIP-src} & \textbf{L1-src} & \textbf{TaskScore} \\
        \midrule
        ACE & 4.275 & 43.159 & 0.276 & 0.791 & 0.860 & 0.016 & 0.5514 \\
        OmniGen & 4.015 & 42.527 & 0.261 & 0.380 & 0.815 & 0.066 & 0.4708 \\
        ACE++ & 4.231 & \textbf{43.276} & 0.277 & 0.834 & 0.899 & 0.012 & \textbf{0.5593} \\
        UltraEdit & \textbf{4.506} & 38.887 & 0.277 & 0.098 & \textbf{0.946} & 0.014 & 0.4462 \\
        \textbf{Ours (FLUX Subgroup)} & 4.230 & 41.024 & \textbf{0.278} & \textbf{0.865} & 0.873 & \textbf{0.010} & 0.5564 \\
        \bottomrule
    \end{tabular}}
\end{table}

\begin{table}[H]
    \centering
    \caption{ICE-Bench component metrics for Task 22 (Local Text Removal).}
    \label{tab:flux2_task22_components}
    \scriptsize
    \setlength{\tabcolsep}{2.8pt}
    \resizebox{\linewidth}{!}{%
    \begin{tabular}{lrrrrrrr}
        \toprule
        \textbf{Method} & \textbf{Aesthetic} & \textbf{Imaging} & \textbf{CLIP-cap} & \textbf{VLLM-QA} & \textbf{CLIP-src} & \textbf{L1-src} & \textbf{TaskScore} \\
        \midrule
        ACE & \textbf{4.896} & 49.766 & \textbf{0.273} & \textbf{0.801} & 0.888 & 0.033 & 0.5910 \\
        OmniGen & 4.487 & 56.398 & 0.246 & 0.674 & 0.793 & 0.097 & 0.5635 \\
        ACE++ & 4.694 & \textbf{59.636} & 0.260 & 0.704 & 0.905 & 0.017 & \textbf{0.5977} \\
        UltraEdit & 4.665 & 47.294 & 0.264 & 0.714 & 0.910 & 0.023 & 0.5625 \\
        \textbf{Ours (FLUX Subgroup)} & 4.864 & 57.414 & 0.273 & 0.638 & \textbf{0.945} & \textbf{0.015} & 0.5922 \\
        \bottomrule
    \end{tabular}}
\end{table}

\begin{table}[H]
    \centering
    \caption{ICE-Bench component metrics for Task 29 (Subject-guided Inpainting).}
    \label{tab:flux2_task29_components}
    \scriptsize
    \setlength{\tabcolsep}{2.8pt}
    \resizebox{\linewidth}{!}{%
    \begin{tabular}{lrrrrrrrr}
        \toprule
        \textbf{Method} & \textbf{Aesthetic} & \textbf{Imaging} & \textbf{CLIP-cap} & \textbf{VLLM-QA} & \textbf{DINO-ref} & \textbf{CLIP-src} & \textbf{L1-src} & \textbf{TaskScore} \\
        \midrule
        ACE & 4.812 & 52.544 & 0.197 & 0.171 & 0.562 & 0.766 & 0.015 & 0.5081 \\
        OmniGen & 4.459 & 59.995 & 0.186 & 0.093 & 0.555 & 0.642 & 0.149 & 0.4936 \\
        ACE++ & 4.835 & \textbf{63.419} & 0.186 & 0.257 & 0.563 & 0.753 & 0.040 & 0.5406 \\
        \textbf{Ours (FLUX Subgroup)} & \textbf{5.031} & 60.533 & \textbf{0.228} & \textbf{0.436} & \textbf{0.580} & \textbf{0.833} & \textbf{0.014} & \textbf{0.5733} \\
        \bottomrule
    \end{tabular}}
\end{table}

\begin{table}[H]
    \centering
    \caption{ICE-Bench component metrics for Task 30 (Virtual Try On).}
    \label{tab:flux2_task30_components}
    \scriptsize
    \setlength{\tabcolsep}{2.8pt}
    \resizebox{\linewidth}{!}{%
    \begin{tabular}{lrrrrrrrr}
        \toprule
        \textbf{Method} & \textbf{Aesthetic} & \textbf{Imaging} & \textbf{CLIP-cap} & \textbf{VLLM-QA} & \textbf{DINO-ref} & \textbf{CLIP-src} & \textbf{L1-src} & \textbf{TaskScore} \\
        \midrule
        ACE & 4.837 & 64.723 & 0.231 & 0.629 & 0.751 & \textbf{0.889} & \textbf{0.006} & 0.6362 \\
        OmniGen & 4.696 & 73.313 & 0.235 & 0.722 & 0.744 & 0.847 & 0.058 & 0.6581 \\
        ACE++ & 4.577 & \textbf{73.525} & 0.243 & 0.804 & \textbf{0.763} & 0.882 & 0.029 & 0.6725 \\
        \textbf{Ours (FLUX Subgroup)} & \textbf{4.930} & 70.693 & \textbf{0.248} & \textbf{0.825} & 0.760 & 0.878 & 0.010 & \textbf{0.6775} \\
        \bottomrule
    \end{tabular}}
\end{table}

\begin{table}[H]
    \centering
    \caption{ICE-Bench component metrics for Task 31 (Face Swap); diagnostic because this task was not used for task-specific training.}
    \label{tab:flux2_task31_components}
    \scriptsize
    \setlength{\tabcolsep}{2.8pt}
    \resizebox{\linewidth}{!}{%
    \begin{tabular}{lrrrrrrrr}
        \toprule
        \textbf{Method} & \textbf{Aesthetic} & \textbf{Imaging} & \textbf{CLIP-cap} & \textbf{VLLM-QA} & \textbf{Face-ref} & \textbf{CLIP-src} & \textbf{L1-src} & \textbf{TaskScore} \\
        \midrule
        ACE & 4.983 & 56.985 & 0.232 & 0.400 & 0.250 & \textbf{0.763} & 0.018 & 0.4907 \\
        OmniGen & 4.309 & 64.021 & 0.217 & 0.484 & \textbf{0.477} & 0.661 & 0.112 & 0.5324 \\
        ACE++ & \textbf{5.034} & \textbf{64.963} & 0.231 & 0.442 & 0.378 & 0.760 & 0.054 & 0.5396 \\
        \textbf{Ours (FLUX Subgroup)} & 4.854 & 64.283 & \textbf{0.255} & \textbf{0.716} & 0.391 & 0.757 & \textbf{0.014} & \textbf{0.5700} \\
        \bottomrule
    \end{tabular}}
\end{table}

\paragraph{ACE++ reproduction.}
We additionally reran the released ACE++ checkpoints on the same
\texttt{SourceMask} task set. The run uses the original ICE-Bench source
images, the task-appropriate released local-editing, subject, and portrait
LoRAs, 50 denoising steps, guidance scale 50, repainting scale 1.0, and seed
42. We then apply the same ICE-Bench evaluator and official task-score
aggregation. Table~\ref{tab:aceplus_reproduction} compares this rerun with
scores reconstructed from the rounded component metrics published in the
ICE-Bench supplement~\citep{pan2025icebench}. The rerun is close in the
nine-task macro average (0.5941 versus 0.5945), while individual tasks differ;
Outpainting is 0.0125 lower and Face Swap is 0.0279 higher.

\begin{table}[H]
    \centering
    \caption{ACE++ reproduction on the nine ICE-Bench \texttt{SourceMask}
    tasks. Published values are approximate composite scores reconstructed
    from rounded component metrics. $\Delta$ is reproduction minus published.}
    \label{tab:aceplus_reproduction}
    \small
    \setlength{\tabcolsep}{5pt}
    \begin{tabular}{lrrr}
        \toprule
        \textbf{Task} & \textbf{Published} & \textbf{Reproduced} & $\boldsymbol{\Delta}$ \\
        \midrule
        17: Inpainting & 0.6430 & 0.6389 & $-0.0041$ \\
        18: Outpainting & 0.6230 & 0.6106 & $-0.0125$ \\
        19: Local Subject Addition & 0.6286 & 0.6224 & $-0.0062$ \\
        20: Local Subject Removal & 0.5464 & 0.5398 & $-0.0065$ \\
        21: Local Text Render & 0.5593 & 0.5667 & $+0.0074$ \\
        22: Local Text Removal & 0.5977 & 0.5870 & $-0.0108$ \\
        29: Subject-guided Inpainting & 0.5406 & 0.5592 & $+0.0187$ \\
        30: Virtual Try On & 0.6725 & 0.6544 & $-0.0181$ \\
        31: Face Swap & 0.5396 & 0.5675 & $+0.0279$ \\
        \midrule
        Nine-task macro & 0.5945 & 0.5941 & $-0.0005$ \\
        \bottomrule
    \end{tabular}
\end{table}

Relative to the ACE++ reference, FLUX.2 BRIDGE is higher on supported Tasks
19, 20, 29, and 30, and on diagnostic Tasks 17 and 31. The two supported-task deficits are 0.0029 on Local Text Render
and 0.0055 on Local Text Removal, while the largest deficit, 0.0429, occurs on
the unsupported Outpainting task. This task-level pattern explains why the
Outpainting-inclusive Task 17--22 group in
Table~\ref{tab:flux2_grouped_comparison} is treated as a diagnostic aggregate.

\subsection{Local Subject Addition Mask Intervention}
\label{sec:flux_mask_intervention}

\begin{figure}[H]
    \centering
    \includegraphics[width=\linewidth,height=0.80\textheight,keepaspectratio]{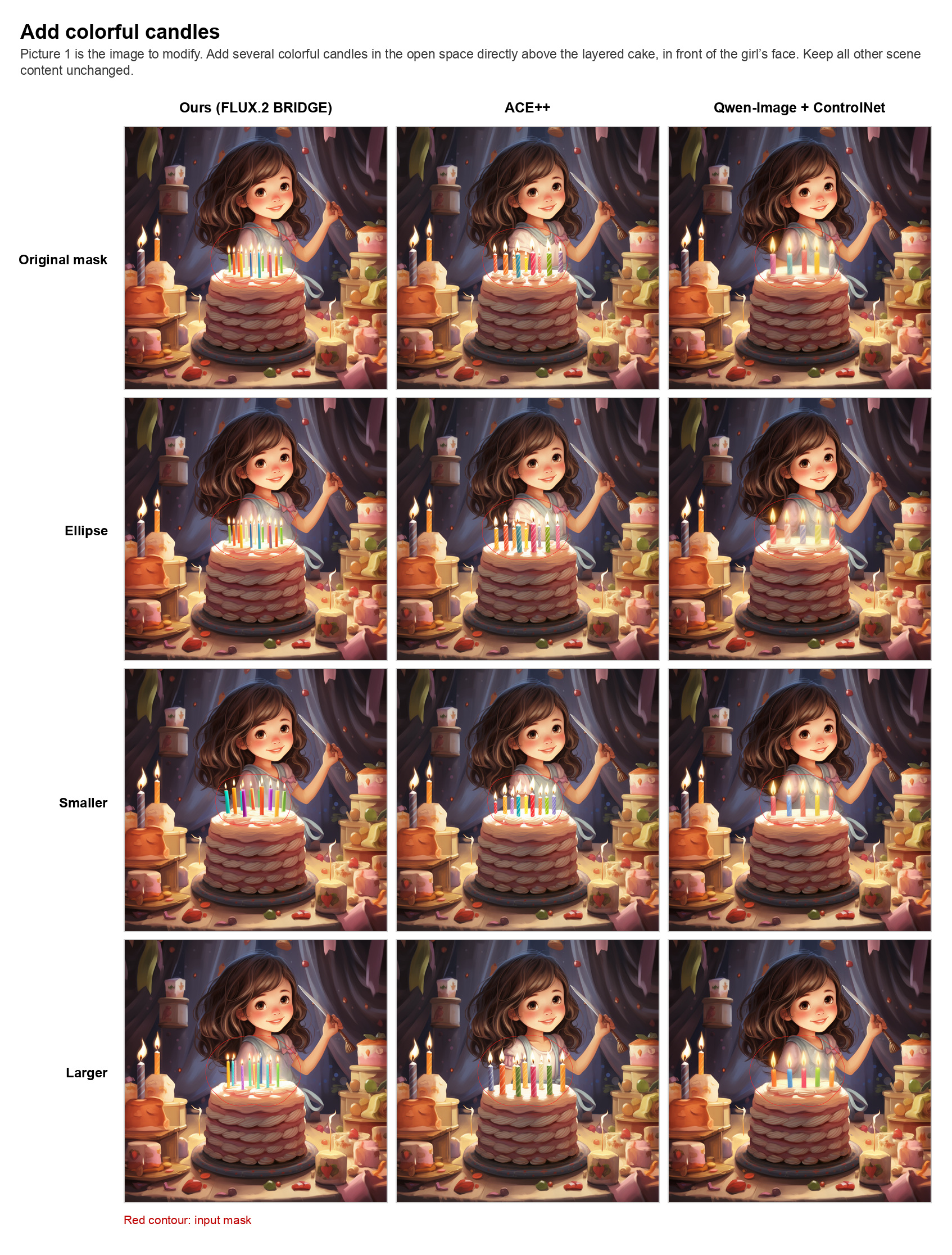}
    \caption{\textbf{Local Subject Addition mask intervention (1/4): colorful candles.}
    Each column uses the same prompt across the four mask conditions.}
    \label{fig:flux_mask_candles}
\end{figure}

\begin{figure}[H]
    \centering
    \includegraphics[width=\linewidth,height=0.80\textheight,keepaspectratio]{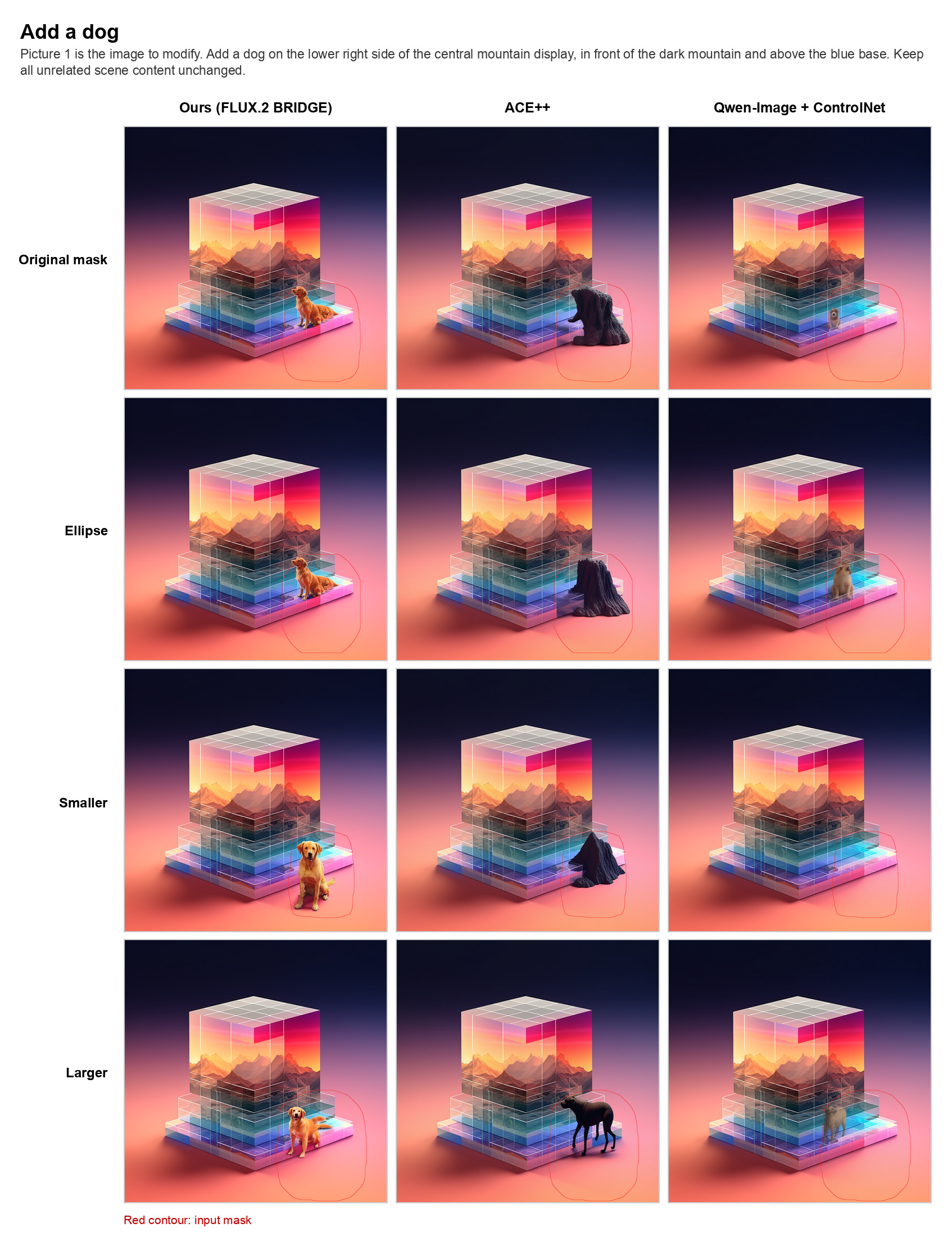}
    \caption{\textbf{Local Subject Addition mask intervention (2/4): dog.}
    The red contour shows the mask supplied to each model.}
    \label{fig:flux_mask_dog}
\end{figure}

\begin{figure}[H]
    \centering
    \includegraphics[width=\linewidth,height=0.80\textheight,keepaspectratio]{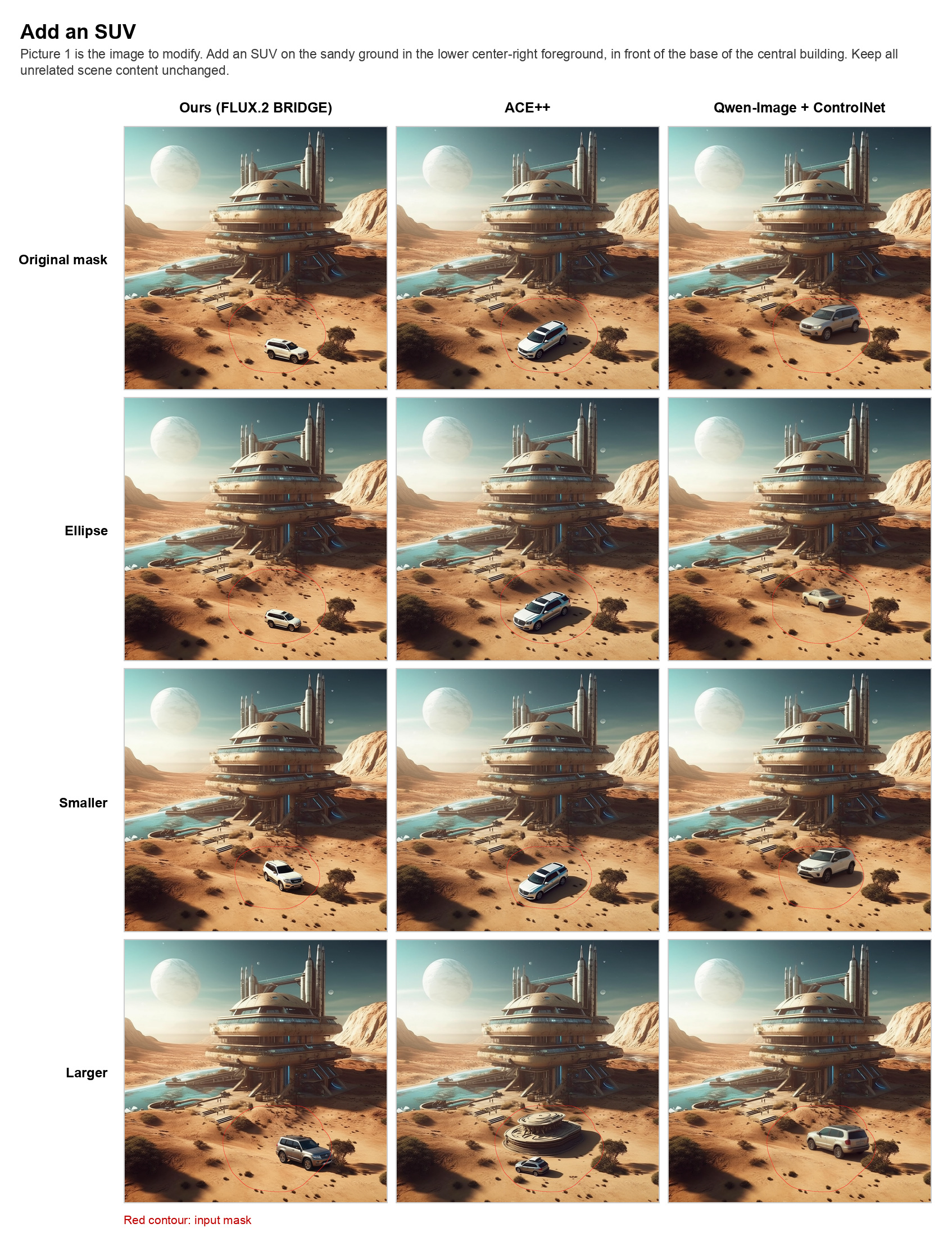}
    \caption{\textbf{Local Subject Addition mask intervention (3/4): SUV.}
    Mask area changes are shown separately from the equal-area ellipse.}
    \label{fig:flux_mask_suv}
\end{figure}

\begin{figure}[H]
    \centering
    \includegraphics[width=\linewidth,height=0.73\textheight,keepaspectratio]{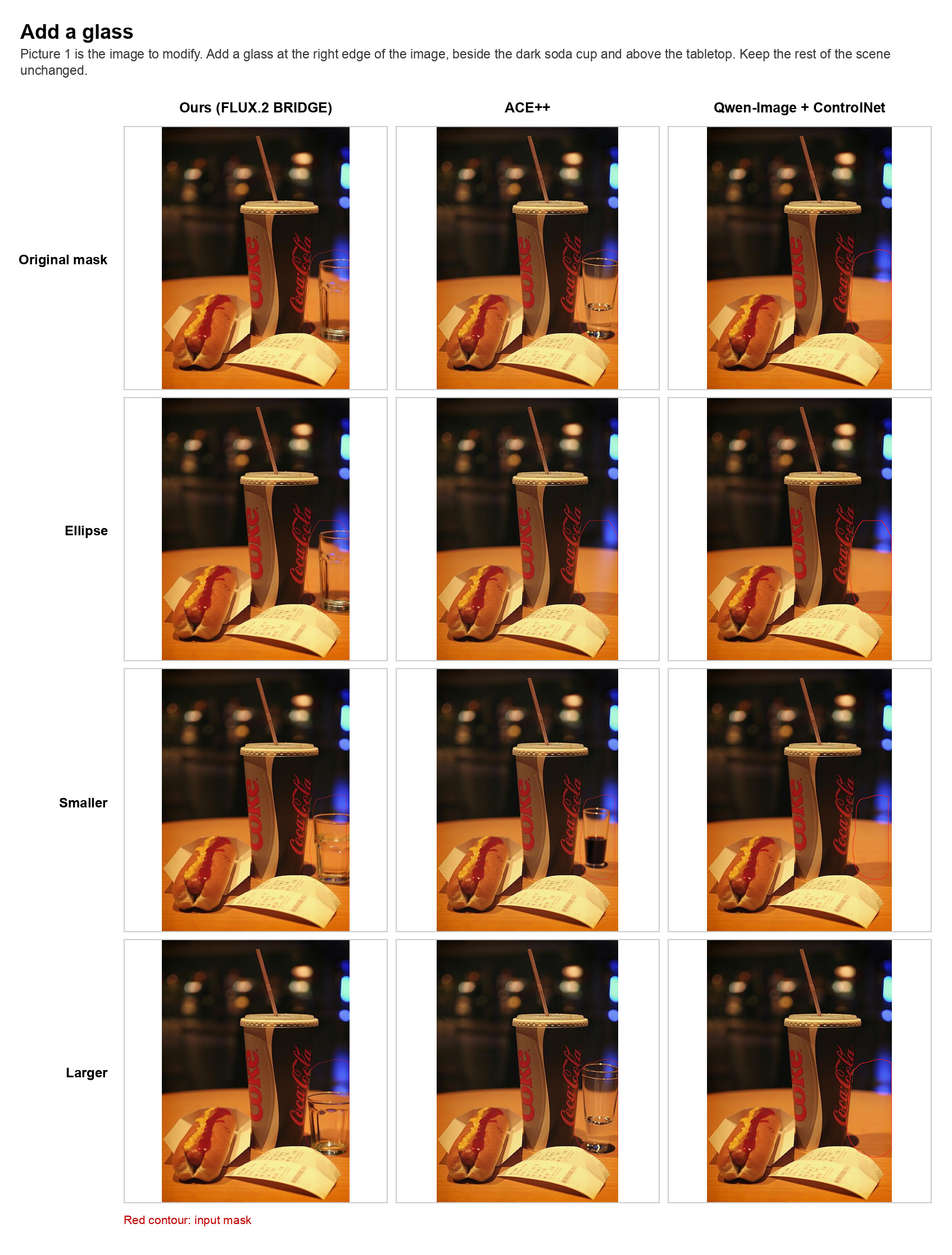}
    \caption{\textbf{Local Subject Addition mask intervention (4/4): glass.}
    BRIDGE consistently inserts a plausible glass across the displayed masks.
    ACE++ is inconsistent, producing missing or malformed glasses, while
    Q-Control fails to insert the requested object.}
    \label{fig:flux_mask_glass}
\end{figure}

We vary the mask while keeping the source image, Luna-rewritten prompt, seed,
and sampling configuration fixed within each model. The four conditions are
the original mask, an equal-area ellipse, a smaller mask, and a larger mask.
Figures~\ref{fig:flux_mask_candles}--\ref{fig:flux_mask_glass} compare BRIDGE,
ACE++, and Q-Control (Qwen-Image-Edit with Blockwise ControlNet
Inpaint)~\citep{diffsynth2025qwencontrol}; the red contour marks the input mask.
These examples diagnose response to mask support and are reported separately
from the ICE-Bench task score.

\clearpage

\clearpage

\clearpage

\clearpage

\subsection{ICE-Bench Qualitative Showcase}
\label{sec:flux2_qualitative_showcase}

The following figures show five selected outputs for each ICE-Bench task from
the same FLUX.2 Dense/BBox run used for the quantitative table. Each row in the
source showcase contains the source image, an optional subject reference, and
the generated image with the 1-pixel red mask boundary. Prompts were rewritten
with GPT-6 Luna and evaluated with CFG $=4$; the diagnostic tasks are included
for completeness and are not evidence of task-specific training.

\begin{figure}[H]
    \centering
    \makebox[\linewidth][c]{%
    \includegraphics[height=0.90\textheight,keepaspectratio]{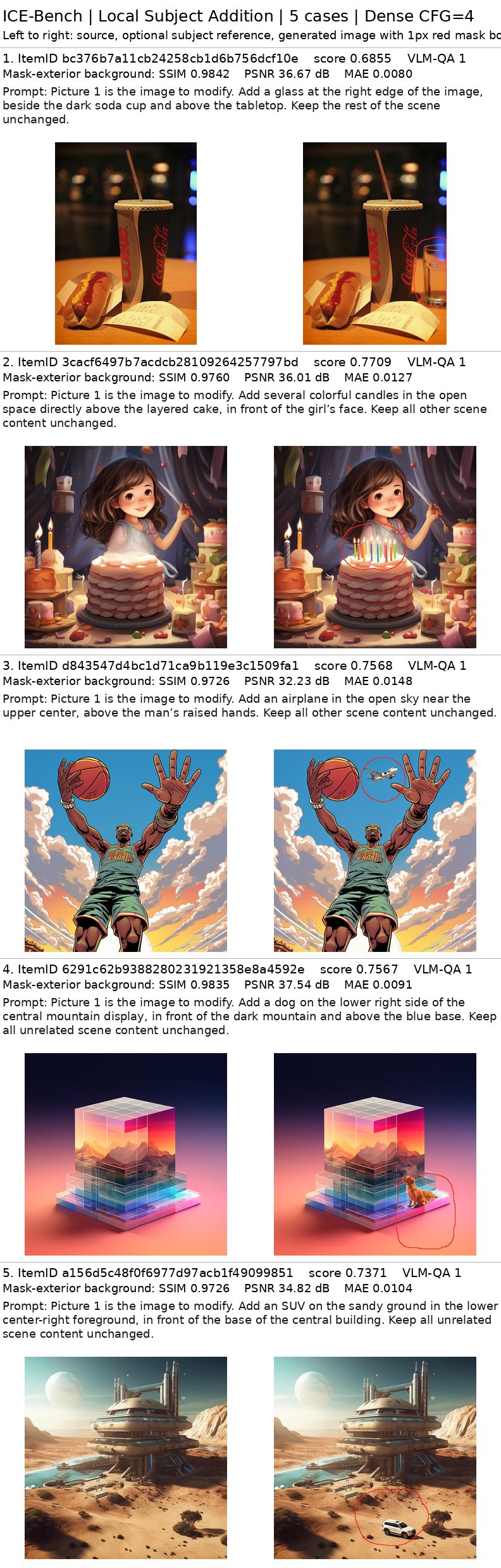}%
    \hspace{0.03\linewidth}%
    \includegraphics[height=0.90\textheight,keepaspectratio]{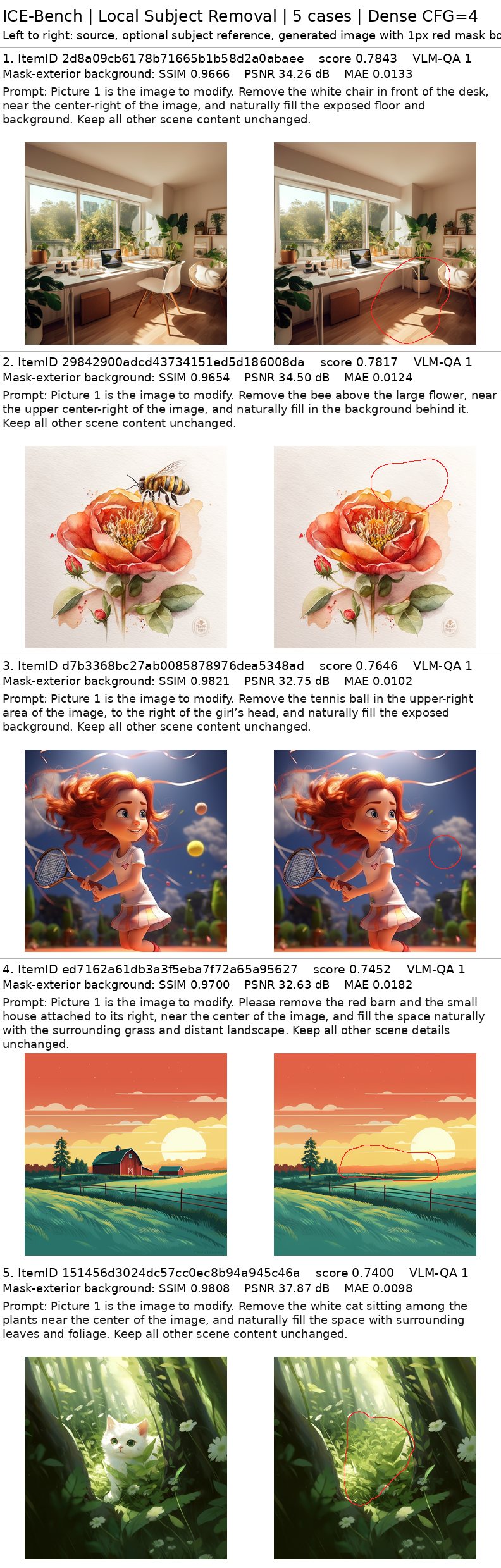}%
    }
    \caption{\textbf{ICE-Bench qualitative showcase (1/6).} Two selected five-case panels from the same Dense/BBox run. The embedded panels retain the rewritten prompt, optional reference, generated image, and red mask boundary.}
\end{figure}

\clearpage
\begin{figure}[H]
    \centering
    \makebox[\linewidth][c]{%
    \includegraphics[height=0.90\textheight,keepaspectratio]{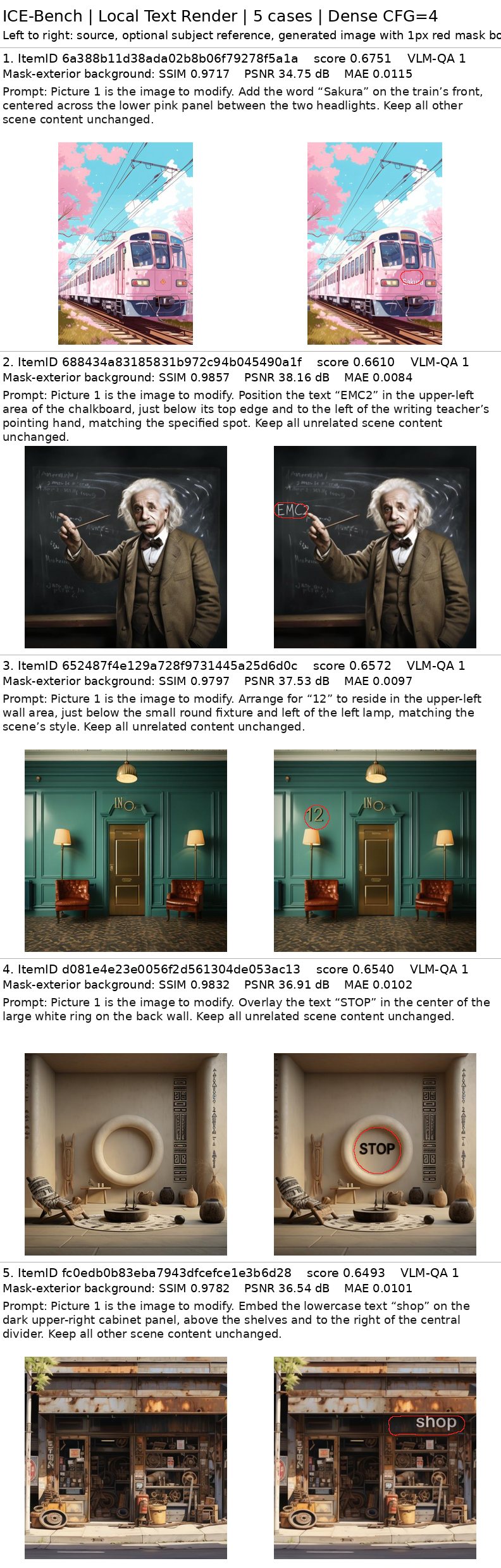}%
    \hspace{0.03\linewidth}%
    \includegraphics[height=0.90\textheight,keepaspectratio]{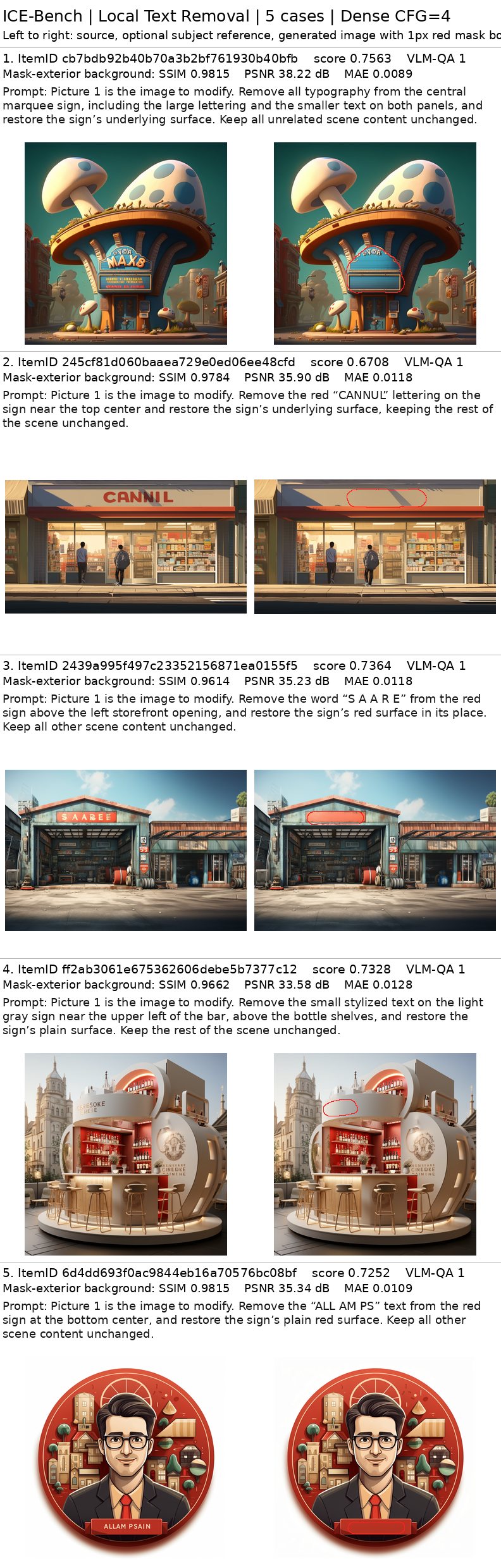}%
    }
    \caption{\textbf{ICE-Bench qualitative showcase (2/6).} Two selected five-case panels from the same Dense/BBox run.}
\end{figure}

\clearpage
\begin{figure}[H]
    \centering
    \makebox[\linewidth][c]{%
    \includegraphics[height=0.90\textheight,keepaspectratio]{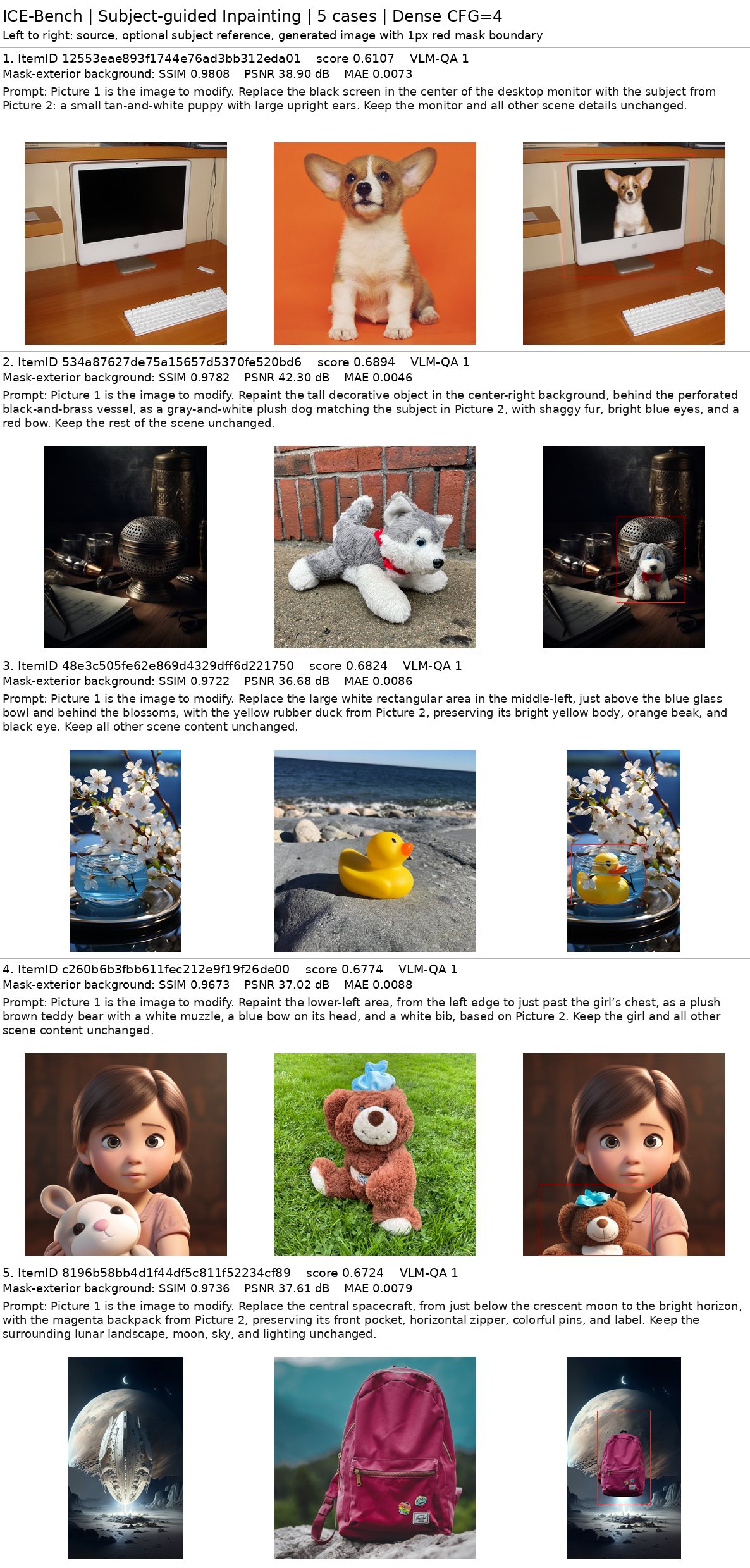}%
    }
    \caption{\textbf{ICE-Bench qualitative showcase (3/6).} One selected five-case panel from the same Dense/BBox run.}
\end{figure}
\clearpage
\begin{figure}[H]
    \centering
    \makebox[\linewidth][c]{%
    \includegraphics[height=0.90\textheight,keepaspectratio]{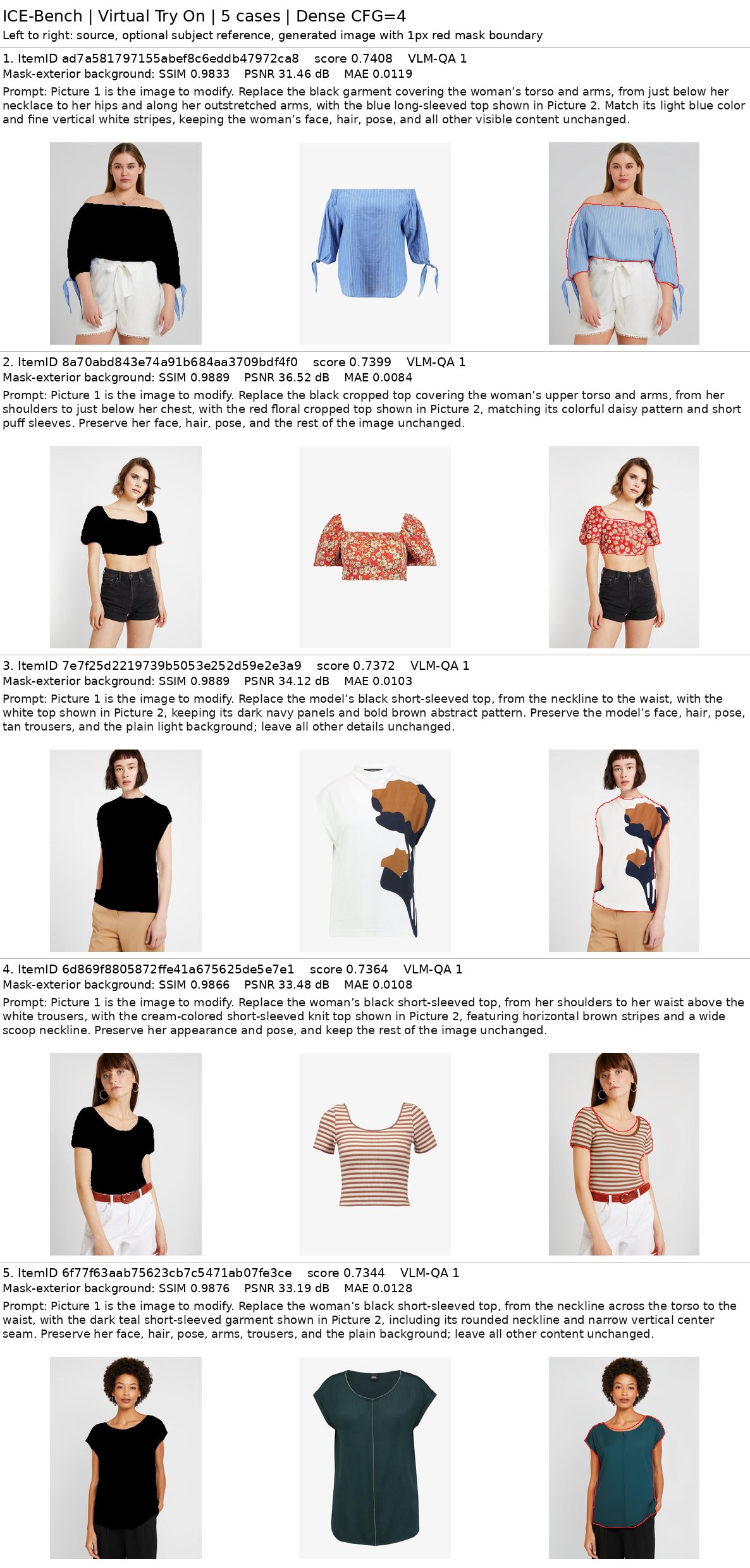}%
    }
    \caption{\textbf{ICE-Bench qualitative showcase (4/6).} One selected five-case panel from the same Dense/BBox run.}
\end{figure}
\clearpage
\begin{figure}[H]
    \centering
    \makebox[\linewidth][c]{%
    \includegraphics[height=0.90\textheight,keepaspectratio]{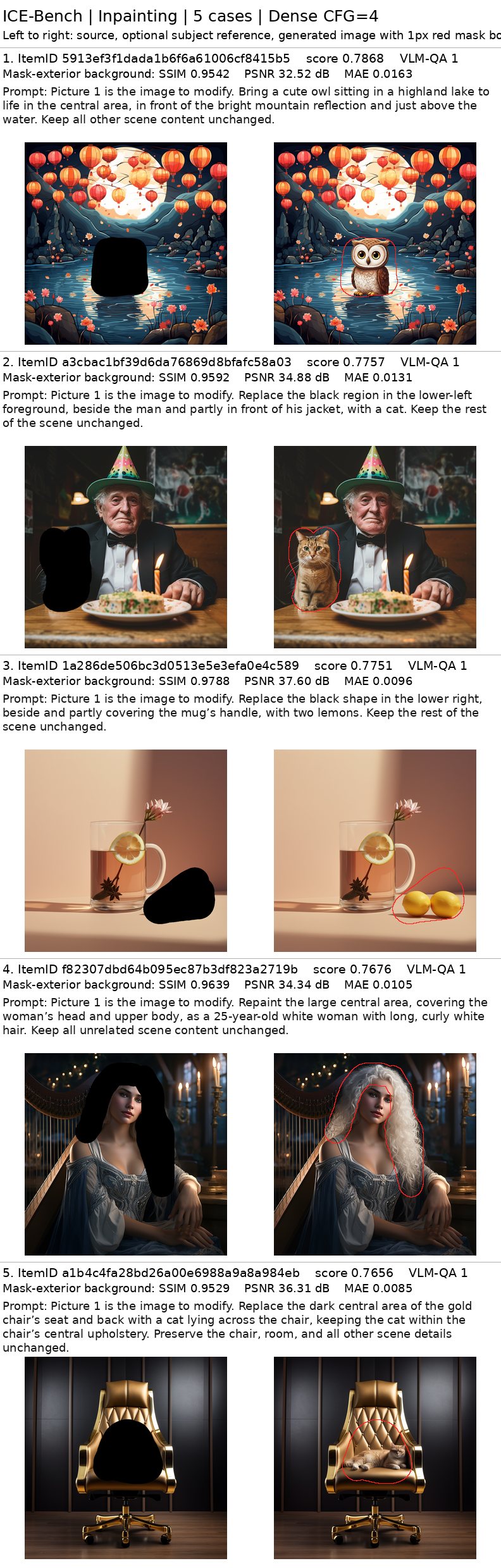}%
    \hspace{0.03\linewidth}%
    \includegraphics[height=0.90\textheight,keepaspectratio]{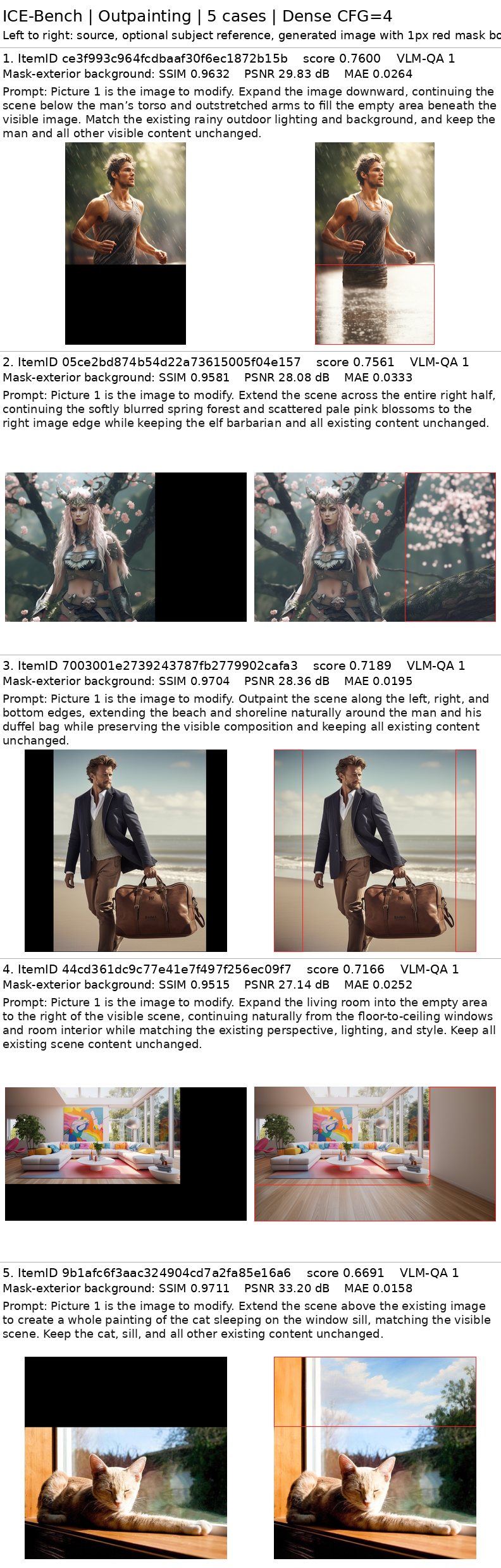}%
    }
    \caption{\textbf{ICE-Bench qualitative showcase (5/6).} Two diagnostic five-case panels from the same Dense/BBox run. These tasks were not used for task-specific training.}
\end{figure}

\clearpage
\begin{figure}[H]
    \centering
    \includegraphics[height=0.90\textheight,keepaspectratio]{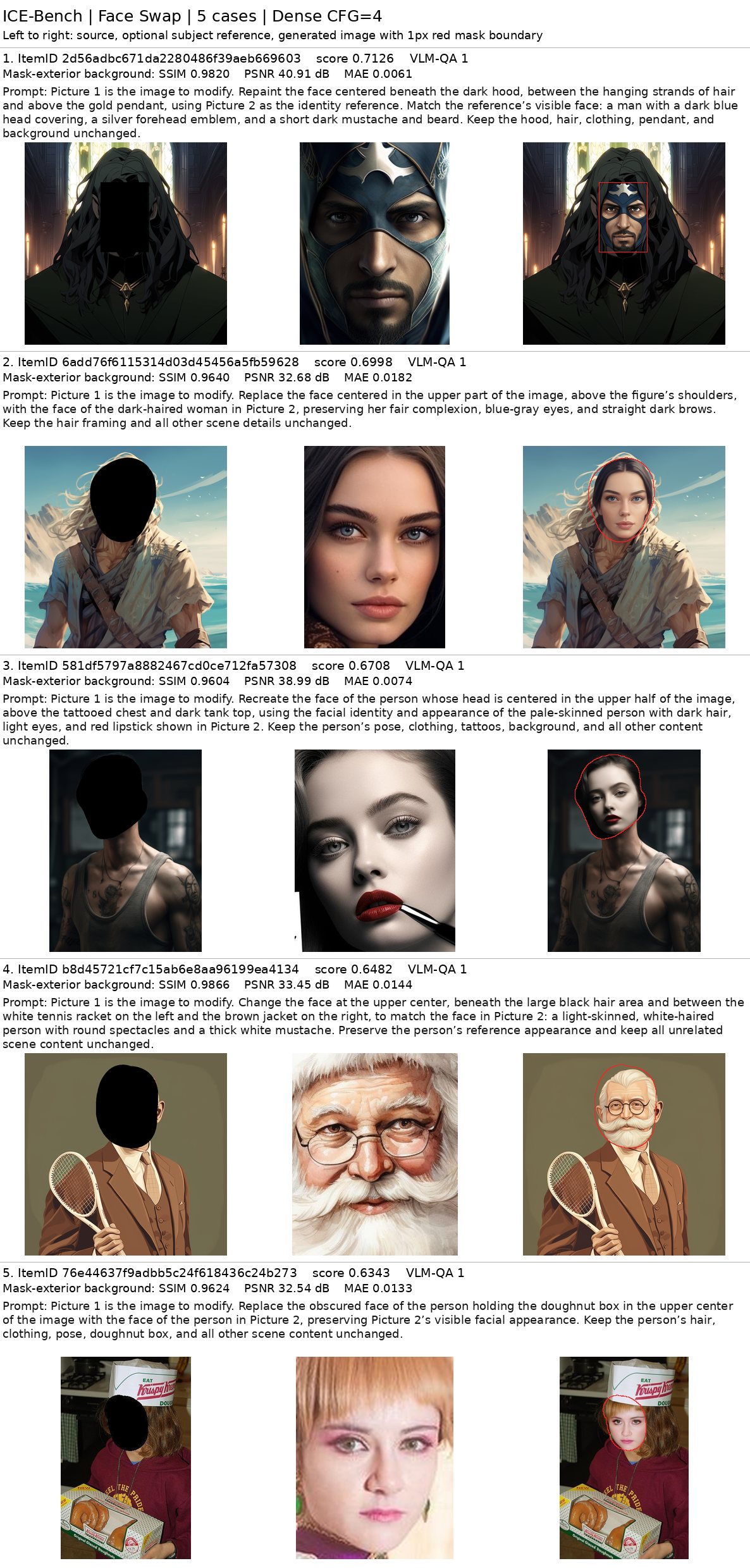}
    \caption{\textbf{ICE-Bench qualitative showcase (6/6).} One diagnostic five-case panel from the same Dense/BBox run. Face Swap was not used for task-specific training.}
\end{figure}

\end{document}